\documentclass{article}
\usepackage{authblk}
\usepackage{amsmath,amsfonts,amsthm,mathrsfs,xcolor,bm,bbm, mathtools, mathdots}
\usepackage{amssymb}
\usepackage{verbatim}
\usepackage{graphicx}
\usepackage{setspace}
\usepackage{natbib}
\usepackage{fancyhdr}
\usepackage[colorlinks=true,urlcolor=blue,citecolor=blue,linkcolor=blue,bookmarks=true]{hyperref}
\usepackage{sectsty}
\usepackage{tcolorbox}
\usepackage{float}
\usepackage[utf8]{inputenc}
\usepackage[left=1in, right=1in, top=1in, bottom=1in]{geometry}
\usepackage{enumerate} 

    \makeatletter
\def\@fnsymbol#1{\ensuremath{\ifcase#1\or \dagger\or \ddagger\or
   \mathsection\or \mathparagraph\or \|\or **\or \dagger\dagger
   \or \ddagger\ddagger \else\@ctrerr\fi}}
    \makeatother
    
\usepackage[textsize=tiny]{todonotes}

\usepackage{soul}
\usepackage{multirow}
\usepackage{booktabs}
\usepackage{subcaption}
\usepackage{adjustbox}
\usepackage{dsfont}
\usepackage{array}

\usepackage[title]{appendix}

\def\NN{\mathbb N}

\def\RR{\mathbb R}

\newtheorem{assumption}{Assumption}
\newtheorem{theorem}{Theorem}
\newtheorem{lemma}{Lemma}
\newtheorem{proposition}{Proposition}
\newtheorem{corollary}{Corollary}
\newtheorem{definition}{Definition}
\newtheorem{remark}{Remark}

\title{Learning to Detect Cyber Attacks: Neural Anomaly Detection for Cybersecurity with Theoretical Insights}
\author{Tian-Yi Zhou$^*$\thanks{Data Science Institute, Columbia University in the City of New York,\texttt{tz2699@columbia.edu}} , Matthew Lau$^*$\thanks{School of Cybersecurity and Privacy, Georgia Institute of Technology, \texttt{mlau40@gatech.edu}} , Jizhou Chen\thanks{School of Cybersecurity and Privacy, Georgia Institute of Technology, \texttt{jzchen@gatech.edu}} , Wenke Lee\thanks{School of Cybersecurity and Privacy, Georgia Institute of Technology, \texttt{wenke@cc.gatech.edu}} , Xiaoming Huo\thanks{H. Milton Stewart School of Industrial and Systems Engineering, Georgia Institute of Technology, \texttt{huo@gatech.edu}}}
\date{\vspace{-5ex}}

\begin{document}

\maketitle
\def\thefootnote{*}\footnotetext{These authors contributed equally to this work.}\def\thefootnote{\arabic{footnote}}
\begin{abstract}
In cybersecurity practice, new forms of cyberattacks continuously emerge, deliberately designed to evade defense systems that rely on previously observed behaviors. Motivated by this challenge, we propose a neural network–based method for anomaly detection that does not rely on (i) prior knowledge of anomaly distributions or (ii) the availability of real anomalies during training. Our proposed  method trains a neural network classifier using only normal samples, combining the supervision from synthetic anomalies, and is particularly suitable when collecting real anomaly samples is expensive or impractical.
The trained classifier is proven to attain minimax excess risk, and more importantly, it is guaranteed to learn the boundary of the normal region. Once the normal region is well estimated, the model can detect a wide range of anomalies without requiring explicit modeling of their distributions.
Extensive experiments across cybersecurity, industrial, and medical anomaly detection tasks demonstrate that our method is consistently robust and competitive compared to state-of-the-art baselines. Notably, in the context of network intrusion detection, our approach significantly enhances the detection of difficult and previously unseen cyber attacks compared to other baselines.
\end{abstract}

\section{Introduction}
Anomaly detection (AD) refers to the problem of finding patterns in data that do not conform to expected behavior.  
 The problem of detecting anomalies or outliers in data has been studied in statistics as early as the 19th century \citep{edgeworth1887xli}, and has become increasingly important in modern data-driven applications.
AD is extensively applied in various domains, such as intrusion detection for cybersecurity \citep{ambusaidi2016building, li2017intrusion, alhajjar2021adversarial}, fraud detection \citep{dorronsoro1997neural, zheng2019one}, disease detection \citep{iakovidis2018detecting, lu2018anomaly}, earthquake detection \citep{perol2018convolutional, seo2024unsupervised}, and even AI safety for large language models (LLMs) \citep{zhu2024llms,zhang2024logicode}.

In cybersecurity, intrusion detection systems (IDS) play a crucial role in identifying unauthorized access, malware, and other malicious activities. Supervised anomaly detection methods attempt to learn classification rules from labeled datasets that contain both normal and attack samples \citep{laskov2005learning, ring2019survey}. However, these methods face significant limitations because of their heavy reliance on large, high-quality labeled data. Collecting and annotating such datasets is often expensive, labor-intensive, and time-consuming \citep{chandola2009anomaly}. 

One of the most pressing challenges in modern cybersecurity is defending against ``zero-day attacks"\footnote{A zero-day attack refers to a cyber attack that exploits a software vulnerability before the security community is aware of it, and before a patch or fix is available. The name "zero-day" comes from the fact that the defenders have had zero days to prepare for it because it was previously unknown.}, which refer to new, previously unseen cyber attacks that intentionally deviate from known patterns to evade detection. These attacks are specifically designed to bypass IDS trained on historical data. Thus, as cyber attacks continually evolve, previously collected labeled data can quickly become outdated, limiting the ability of supervised models to detect previously unseen threats.

Unsupervised anomaly detection provides a promising alternative by learning models of normal behavior from only normal data and identifying deviations as potential anomalies \citep{chandola2009anomaly, ahmed2016survey}. This is particularly valuable for detecting zero-day attacks, where malicious behavior may not resemble any previously observed attack patterns \citep{kim2014novel}. Moreover, collecting normal network traffic data is generally much easier and such data are typically readily available, making the unsupervised approach both practical and scalable. In unsupervised AD, the learning process focuses on characterizing the distribution of normal data and identifying rare and unexpected deviations.

Beyond cybersecurity, unsupervised anomaly detection has proven highly valuable in domains where abnormal events are rare and difficult to label. In earthquake detection, AD methods are used to identify seismic vibration patterns that deviate from normal vibration level \citep{perol2018convolutional, seo2024unsupervised}. 
Actual earthquakes occur much less frequently and can vary significantly in size, type, and location. Because earthquakes are rare and unpredictable, it is hard to collect enough labeled examples to train supervised models. Unsupervised anomaly detection methods that train on only normal vibration data allow for real-time detection of novel or unexpected seismic events, even in regions with limited historical data or expert annotations.

More recently, with the rapid deployment of LLMs, unsupervised AD techniques are actively explored to monitor model outputs for harmful content \citep{zhu2024llms, zhang2024logicode}. Labeled datasets for harmful LLM usage often require reviewing sensitive user inputs, private conversations, or personal information, which may violate privacy regulations \citep{abadi2016deep, shokri2017membership}. Unsupervised methods learn from only normal, safe outputs to model typical behavior. They can then flag unusual or potentially unsafe outputs without needing to collect or store large amounts of sensitive, labeled data. This makes unsupervised approaches especially appealing for privacy-preserving monitoring of LLM safety. 

In the absence of labeled anomalies, a common strategy in unsupervised AD is to generate a
new set of data and assign them as anomalies, which is often referred to as “synthetic anomalies” \citep{sipple20NSNN, mayer2020privacy, schluter2022natural}. The model is then trained using normal data, augmented with supervision derived from these synthetic anomalies.
While this approach provides a flexible and practical means of introducing a supervision signal, it also presents several challenges. For example, deciding how many synthetic anomalies to generate, and how to sample them effectively, especially when dealing with complex, high-dimensional data.


As neural networks become more ubiquitous due to their strong empirical performance, there has been much work on applying neural networks for AD (e.g., DeepSVDD \citep{deep_svdd}, DAGMM \citep{zong2018deep} and references therein).
These methods leverage the representational power of neural networks to model complex data. However, despite their empirical success, the theoretical understanding of neural network-based anomaly detection remains limited, and open questions persist regarding their generalization guarantees and optimal sample complexity.
In particular, \citet{sipple20NSNN}  proposes generating synthetic anomalies and using a neural network to distinguish them from normal data. This approach is flexible and allows for explainable interpretations, and can also incorporate supervision of real anomalies, as demonstrated in DeepSAD \citep{Ruff2020DeepSAD} or in \cite{lau2024revisiting_XOR, lau2024geometric_OSR}.
However, their approach has some limitations.
First, although the neural network-based approach offers practical flexibility, it lacks theoretical guarantees regarding its anomaly detection accuracy and generalization performance.
Second, they provide no guide on the proportion of synthetic anomalies that should be generated.
The number of synthetic anomalies that \citet{sipple20NSNN} uses is 30-60 times the number of normal training data, but this is empirically tested within a small range.

Motivated by the growing prevalence of zero-day attacks in cybersecurity, we propose a neural network-based method for anomaly detection that (i) operates without prior knowledge of the anomalies it may encounter and (ii) does not rely on the availability of real anomalous data. By formulating AD as a binary classification problem through density level set estimation, our approach trains a ReLU neural network using supervision derived from synthetic anomalies.

Despite the widespread empirical success of deep learning in anomaly detection, scant theoretical results are present to corroborate the observed efficiency and effectiveness of neural networks. We address this gap by providing formal guarantees for our method: specifically, we show that the trained network achieves an optimal excess risk bound, representing a novel theoretical contribution to the AD literature. Moreover, our theory offers practical guidance on how many synthetic anomalies should be generated to achieve optimal performance.

It is important to emphasize that our method generates synthetic anomalies not with the intention of resembling any unknown anomalies, but rather to provide a contrastive signal for learning the decision boundary of normal data. In anomaly detection, unlike standard binary classification, the focus is solely on distinguishing normal from not normal. There may be infinitely many types of anomalies, but as long as the model correctly identifies the boundary of normal behavior, it can flag anomalies effectively.

Once the classifier is trained, it induces a decision set that provably converges to the true region representing normal data, thereby providing both theoretical soundness and practical robustness.

We implement our proposed method in extensive experiments on real-world datasets. While our approach is initially motivated by intrusion detection in cybersecurity, it is general and demonstrates strong performance across various domains, including network intrision detection, industrial image inspection, and anomaly detection in clinical data. The experimental results corroborate our theoretical findings. We also discuss several practical challenges encountered during implementation --- such as vanishing gradients and overfitting --- and outline strategies to effectively guide the search for an optimal neural network configuration. Empirical comparisons with existing anomaly detection methods show that our approach is competitive across various types of cyber attacks and can even achieve significantly better performance than random guessing in scenarios where other methods fail.


Our paper is organized as follows.
Section \ref{section:problem formulation} formulates AD as a density level set estimation problem, and then shows the equivalent mapping to a binary classification problem via likelihood ratio testing.
Section \ref{section:proposed method}  introduces our proposed method, which trains a ReLU network classifier using normal samples and synthetic anomalies via empirical risk minimization.
Section \ref{section:theoretical results} presents our main theorems, which proves that (i) the excess risk of the neural network classifier converges at a minimax optimal rate, (ii) such a classifier can induce a set that provably converges to the set representing normal region. 
In Section \ref{section:experiments}, we evaluate our proposed method on network intrusion detection, detailing the experimental setup, evaluation metrics, and practical considerations in implementing the neural network and its optimization process.
Section \ref{section:discussion} presents discussions on extending our framework to various types of anomaly detection problems. Section \ref{section:related} provides an extensive literature review, focusing on deep learning–based approaches to anomaly detection. Finally, we conclude the main text in Section \ref{section:conclusion}.
 Supplementary Material provides additional experimental results and discussions, detailed descriptions of the experimental datasets, further remarks on our theoretical findings, and complete technical proofs.

\section{Problem Formulation}
\label{section:problem formulation}
We first model the unsupervised AD problem as a density level set estimation problem. 
Then, we formulate a binary classification problem of ``normal'' vs. ``anomaly''.
We will show that solving the classification problem can, in turn, solve the level set estimation problem.

\subsection{Model Anomaly Detection as Density Level Set Estimation}
\label{section:AD_as_DLSE}
The estimation of the level set $\{x: h(x)>\rho\}$, where $h$ is an unknown density function of $\RR^d$ and $\rho>0$ is a given constant, has been studied by relatively few researchers and is considered a difficult problem in statistics. Most of the existing methods require strong assumptions on the level set (e.g., the boundary of the level set has a certain smoothness) that are difficult to verify in practice \citep{tsybakov1997nonparametric, dbscan, cholaquidis2022level} \footnote{A notable example is in \cite{tsybakov1997nonparametric}, where empirical excess masses were used to estimate level sets. The estimation rate deteriorates when $d>2$ \citep[Theorem 4]{tsybakov1997nonparametric}; thus is not ideal for handling problems in high-dimensional settings.}.
 
To interpret anomaly detection as a density level set estimation problem, we leverage the key property that 
anomalies typically occur less frequently than normal data. 

Let $\mathcal{X}$ be a measurable space subset of $\RR^d$, and $\mu$ be a known probability measure on $\mathcal{X}$. Let $Q$ be an unknown data-generating distribution on $\mathcal{X}$, which has an unknown density $h$ with respect to $\mu$, i.e., $dQ = h \; d\mu$.
We know that the density level set  $\{h > \rho\}:=\{x:h(x)>\rho\}$, for any $\rho>0$, describe the concentration of $Q$. 

Due to the low concentration of anomalies relative to normal data, detecting anomalies can be viewed as detecting density level sets for the data-generating density $h$. 
 Precisely, for a chosen threshold level $\rho>0$, the set $\{h \leq \rho\}$ detects anomalous data, and equivalently, the set $\{h > \rho\}$ detects normal data. 
 
Throughout this work, we assume $h$ is not $\mu$-almost surely constant, so that the data-generating distribution $Q$ is not identical to the reference measure $\mu$. Othewise, the level set  $\{h > \rho\}$ would be either empty or all of $\mathcal X$, depending on the choice of $\rho$.

We study an unsupervised AD problem where only real normal data is available. 
Throughout this work, we label normal data 
by $Y=1$ and anomalies by $Y=-1$.
Consider a set of random normal data $T= \{(X_i,1)\}_{i=1}^{n}$, where $\{X_i\}_{i=1}^n$ are assumed to be drawn i.i.d. from $Q$.
We know the level set estimation problem aims to estimate the $\rho$-level set of $h$ (i.e., the set $\{h>\rho\}$) with a chosen $\rho>0$. 
We may use the given normal data $\{X_i\}_{i=1}^n$ to construct a function $f_{T}:\mathcal{X} \to \RR$ for which the set $\{f_{T} > 0\}$
is an estimate of $\{h>\rho\}$.
For any measurable function $f: \mathcal{X} \to \RR$, the following performance measure is widely used
to measure how much the sets $\{f>0\}$ and  $\{h>\rho\}$ coincide with one another with respect to the measure $\mu$ \citep{tsybakov1997nonparametric, ben1997learning, halmos2013measure}:
\begin{equation}\label{DLDmeasure}
    S_{\mu,h,\rho}(f):= \mu\left(\{f>0\}\Delta\{h>\rho\}\right),
\end{equation}
where $\Delta$ denotes the symmetric difference. 
Intuitively, the smaller $S_{\mu,h,\rho}(f)$ is, the more $\{f>0\}$ coincide with $\{h>\rho\}$. For a function $f$, $S_{\mu,h,\rho}(f) =0$ if and only if $\{f>0\}$ is $\mu$-almost surely identical to $\{h>\rho\}$. 

Here, it is important to note that we cannot compute $S_{\mu,h,\rho}(f)$ for all $f$ because $\{h>\rho\}$ is unknown to us. Therefore, we have to estimate it. In statistics, estimating level sets have long been considered a challenging problem.  
Thus, a drawback of modeling unsupervised AD as a level set estimation problem is that there is a lack of an empirical measure of $S_{\mu,h,\rho}(\cdot)$ such that we cannot compare different estimates of $\{h>\rho\}$.

Hence, we want to find an empirical measure that effectively estimates $S_{\mu,h,\rho}(\cdot)$.
In the next subsection, we transform density level set estimation to a binary classification problem.
More precisely, we show the excess risk of the binary classification problem can be used to estimate $S_{\mu,h,\rho}(\cdot)$. 
%

\subsection{Density Level Set Estimation as Binary Classification}
We formulate a binary classification problem of ``normal'' ($Y=1$) vs. ``anomaly'' ($Y=-1$). Furthermore, we demonstrate the equivalence between solving this classification problem and performing density level set estimation.

Let $\mathcal{Y}=\{-1,1\}$. Let $P$ be the probability measure on $\mathcal{X} \times \mathcal{Y}$. 
We proceed to define $P$ based on our normal data distribution $Q$ where $dQ = h \; d\mu$.
We shall first impose an assumption on the marginal distribution on $\mathcal{X}$. 
Let $s \in (0,1)$ denote the proportion of normal data on the domain $\mathcal{X}$. 
For a pair of random variable $(X,Y) \in \mathcal{X} \times \mathcal{Y}$, we correspondingly define $P(Y=1)=s$ and $P(Y=-1)=1-s$. 

One common approach in unsupervised AD is to artificially generate a separate set of training data and assign them as anomalies, which we refer to as synthetic anomalies (see, e.g., \citep{sipple20NSNN, mayer2020privacy, schluter2022natural}). Following this approach, 
we generate a set of synthetic anomalies $\{X_i^\prime\}_{i=1}^{n^\prime}$ i.i.d. from the known measure $\mu$, and we label them with $Y=-1$ to get $T^\prime = \{(X_i^\prime,-1)\}_{i=1}^{n^\prime}$.
Then, we merge the real normal data and synthetic anomalies to form a new training data set: $$T \cup T^\prime = \{(X_i,1)\}_{i=1}^n \cup \{(X_i^\prime,-1)\}_{i=1}^{n^\prime},$$ enabling the determination of the optimal classification rule for AD.  
We assume for $Y=1$, we have $X \sim Q$ and for $Y=-1$, we have $X \sim \mu$. 
We have the marginal distribution on $\mathcal{X}$ is given by $P_\mathcal{X} = sQ + (1-s)\mu$.
The conditional class probability function $\eta(X):=P(Y=1|X)$ is given by 
\begin{equation}\label{eta}
    \eta(X)=P(Y=1|X)=\frac{s\cdot h(X)}{s\cdot h(X)+1-s}, \qquad \forall X\in \mathcal{X}.\footnote{For any function $f$ on $\mathcal{X} \times \mathcal{Y}$, 
we have $\int_{\mathcal{X}\times \mathcal{Y}} f(X,Y) dP = s\int_\mathcal{X} f(X, 1) dQ + (1-s)\int_\mathcal{X} f(X, -1) d\mu$. From this, we can deduce $\eta(X)=P(Y=1|X)$ given in \eqref{eta}.}
\end{equation}
Then, we see that 
 \begin{equation}\label{data}
     Y|X \  \sim \ \text{Bernoulli}\left(\eta(X) = P(Y=1|X) = \frac{s\cdot h(X)}{s\cdot h(X)+1-s}\right).
 \end{equation}

Now, we consider a binary classification problem where the normal class is labeled with $Y=1$ and the anomaly class is labeled with $Y=-1$.
For a classifier sign$(f)$ induced by  $f: \mathcal{X} \to \RR$, the misclassification error is given as
 \begin{equation*}
     R(f) = \mathrm{P}(\text{sign}(f(X)) \neq Y).
 \end{equation*}
 The Bayes risk, denoted by $R^*$, is the minimal achievable error defined as
 \begin{equation*}
     R^* := \inf_{f: \mathcal{X} \to \RR \text{ measurable}} R(f),
 \end{equation*}
 and is attained by the Bayes classifier
$f_c$, satisfying $R(f_c) = R^*$.

Using our training data set $T\cup T^\prime$, we are interested in finding a function $f_{T,T^\prime}$ such that $R(f_{T,T^\prime}) \to R^*$
and thus $S_{\mu,h,\rho}(f_n) \to 0$, as the number of training data  grows.

With  \eqref{data}, the  misclassification error can be expressed  explicitly as
\begin{equation}
    R(f) = s\mathrm{E}_Q[\mathbbm{1}\{1 \cdot \text{sign}(f(X))=-1\}] + (1-s) \mathrm{E}_\mu[\mathbbm{1}\{-1 \cdot \text{sign}(f(X))=-1\}]
    \end{equation}
    and the Bayes risk 
    \begin{equation}
        R_P^* = s\mathrm{E}_Q[\mathbbm{1}\{h\leq\rho\}] + (1-s)\mathrm{E}_\mu[\mathbbm{1}\{h>\rho\}]
    \end{equation}
can be attained by a Bayes classifier
\begin{equation*}
    f_c (X)= \mathbbm{1}\{h(X)>\rho\} - \mathbbm{1}\{h(X)\leq\rho\} = \begin{cases}
1, \quad h(X)>\rho,\\
-1, \quad h(X)\leq\rho.
\end{cases}
\end{equation*}
When $s = \frac{1}{1+\rho}$, we know $\eta(X) = \frac{s\cdot h(X)}{s\cdot h(X)+1-s} > 1/2$ is equivalent to $h(X) > \rho = \frac{1-s}{s}$, and thus 
\begin{eqnarray*}
     f_c (X)
     = \mathbbm{1}\{\eta(X) > 1/2\} - \mathbbm{1}\{\eta(X) \leq 1/2\} = \begin{cases}
1, \quad \eta(X) > 1/2\\
-1, \quad \eta(X) \leq 1/2.
\end{cases}   
\end{eqnarray*}

We adopt the Tsybakov noise condition \citep{tsybakov2004optimal}, a standard assumption in binary classification and is widely used to study the quantitative behaviors of classification algorithms:
\begin{assumption}\label{Tsybakov}
  For some $c_0>0$ and $q\in [0, \infty)$, $$\mathrm{P}_{\mathcal{X}}\bigl(\{X\in \mathcal{X} : \left|\eta(X) -1/2\right| \leq t\}\bigr) \leq c_0t^q,\quad \forall t>0,$$ where $q$ is often referred to as the {\bf noise exponent}.  
\end{assumption}

 It tells us that when $q$ increases, $\eta$ is more likely to move away from $1/2$. A detailed discussion and elaboration of this condition are relegated to the Supplementary Material 1.1.

 With the above noise condition with noise exponent $q \in [0,\infty)$, a comparison theorem (from \citet[Theorem 10]{steinwart2005classification}) between the excess risk (i.e., excess misclassification error)  $R(\cdot) - R^*$ and $S_{\mu,h,\rho}(\cdot)$ is established for some constant $c>0$ :
\begin{equation}
   R(f) - R^* \leq c S_{\mu,h,\rho}(f)
\end{equation}
and 
\begin{equation}\label{compare}
    S_{\mu,h,\rho}(f) \leq c \left(R(f) - R^*\right)^{\frac{q}{q+1}}
\end{equation}
for any measurable function $f: \mathcal{X} \to \RR$. 
Our goal is to find a classifier $f_{T,T^\prime}$ using our training data set $T\cup T^\prime$ such that $f \to f_c$, or equivalently $R(f)- R^* = R(f)- R(f_c) \to 0$. 
Then, by \eqref{compare}, $S_{\mu,h,\rho}(f) \to 0$ as $R(f)- R^* = R(f)- R(f_c) \to 0$, showing that the excess risk 
$R(\cdot) - R^*$ serves as a viable error metric for anomaly detection. 

Our method constructs a ReLU network classifier that converges to $f_c$. It is detailed in the next section.
\section{Proposed Method --- Empirical Risk Minimization with ReLU Neural Network}
\label{section:proposed method}
We consider merging the two labeled training sets $T$ (real normal data) and $T^\prime$ (synthetic anomalies) as in the previous section, and we use ReLU neural networks to perform the classification task. 
Our goal here is to specify a hypothesis space, which consists of functions implementable by a specific class of neural networks (i.e., neural networks with specific depths, widths, and parameter values in a specific range) such that the empirical risk minimizer (ERM) in that hypothesis space can learn the Bayes classifier well. 

\subsection{Mathematical Formulation of ReLU Neural Networks}
We consider rectified linear unit (ReLU) feed-forward neural networks that take $d$-dimensional inputs and produce scalar outputs.
To mathematically define the ReLU neural network, we first introduce some notations.

Let $\sigma(x) = \max\{0,x\}$ be the ReLU activation function. When $x$ is a vector, the ReLU activation function is applied element-wise. 
A ReLU neural network with $L\in \NN$ hidden layers and width vector $\bm {p} = (p_1, \ldots, p_L) \in \NN^L$, which indicates the width in each hidden layer, is defined in the following
compositional form:
\begin{equation} \label{relufnn}
     f(X) = f_{\bm \theta}(X) :=a \cdot \sigma \left(W^{(L)}\sigma \left(W^{(L-1)}\ldots \sigma \left(W^{(1)} X + b^{(1)}\right)\ldots + b^{(L-1)}\right)+ b^{(L)}\right),
\end{equation}
where $X\in \mathcal{X} = [0,1]^d$ is the input,  $a\in \RR^{p_L}$ is the outer weight, $W^{(i)}$ is a $p_i \times p_{i-1}$ weight matrix with $p_0=d$, and $b^{(i)} \in \RR^{p_i}$ is a bias vector, for $i=1,\ldots,L$. Denote by $\bm{W} = \left\{W^{(i)}\right\}_{i=1}^L$ the set of all weight matrices, $\bm{b} = \left\{b^{(i)}\right\}_{i=1}^L$ the set of all bias vectors, and $\bm \theta = \{\bm{W}, \bm{b},a\}$ the collection of all trainable parameters in the network. 

Let $\|W^{(i)}\|_0$ and $|b^{(i)}|_0$ denote the number of nonzero entries of $W^{(i)}$ and $b^{(i)}$ in the $i$-th hidden layer. 
Let $\|\bm \theta\|_\infty$ denote the largest absolute value of the parameters in $\bm \theta$, that is, 
$$\|\bm \theta\|_\infty = \max \left\{\max_{1\leq i \leq L} \max_{j,k}\left|W^{(i)}_{jk}\right|, \max_{1\leq i \leq L} \|b^{(i)}\|_\infty\right\}. $$
Let $\|\bm {p}\|_\infty$ denote the maximum number of nodes among all hidden layers. 
For $L,w,v,K>0$, we denote the final form of neural network we consider in this paper by
\begin{equation}
\label{eqn:hypothesis_class}
    \mathcal{F}(L, w, v, K) := \left\{f \text{ of the form of } \eqref{relufnn}:  \|\bm {p}\|_\infty \leq w, \sum_{i=1}^L \left(\|W^{(i)}\|_0 + |b^{(i)}|_0\right) \leq v, \|\bm \theta\|_\infty \leq K\right\}. 
\end{equation}
Specifically, the function class $\mathcal{F}(L , w, v, K)$ consists of ReLU networks with depth $L$, width vector $\bm {p}$ with the maximum number of nodes $w$, number of network parameters $v$, and the absolute value of all parameters is bounded by $K$. 


\subsection{Target Neural Network Function Class}

Here, we aim to specify a hypothesis space (i.e., a neural network function space) such that function(s) in this hypothesis space can learn the Bayes classifier well. 
To do so, we must impose some constraints on the density function $h$, thus imposing constraints on the conditional probability function $\eta= \frac{s\cdot h}{s\cdot h+1-s} $. 

For simplicity, we consider the domain $\mathcal{X}=[0,1]^d$; however, this can be extended to any compact subset of $\RR^d$.  We assume $h$ is $\alpha$-Hölder continuous for some $\alpha>0$ in a closed ball of radius $r>0$. Please refer to Supplementary Material 1.2 for the formal definition of the corresponding Hölder space. Note that given $h$ is $\alpha$-Hölder continuous, we can easily prove that $\eta$ is also $\alpha$-Hölder continuous.

Our goal is to construct a specific class of neural networks that learn the Bayes classifier $f_c = \text{sign}\left(\eta - 1/2\right)$ (i.e., the sign of $\eta - 1/2$). 
We employ findings from a previous paper \citep{schmidt2020nonparametric}, which tells us that a certain class of ReLU networks can approximate any Hölder continuous function well. We relegate the details of this previous result to Supplementary Material 5. 

 Now, we are in a position to define our hypothesis space. In machine learning, the hypothesis space generally refers to the set of all possible models or functions that a learning algorithm can choose from to make predictions based on the given data. In our case, the hypothesis space is a set of functions generated by a specific class of ReLU feedforward network architecture (i.e., ReLU networks with specific depths, widths, sparsity, and parameter values in a specific range). We define such a hypothesis space with the intention that its functions are well-suited to learning the Bayes classifier.
 \begin{definition}[Hypothesis Space]\label{hypothesis}
    Let $\alpha,r>0$. 
 Assume $\eta \in \mathcal{H}^{\alpha,r}([0,1]^d)$.
 For any integers $m\geq 1$ and $N \geq \max \left\{(\alpha +1)^d, (r+1)e^d\right\}$, 
    we consider ReLU networks  
    \begin{equation*}
        \widehat{\eta} \in \mathcal{F}(L^*, w^*, v^*,K^*)
    \end{equation*} with depth $$L^*= 8+ (m+5)(1+\lceil \log_2(\max\{d,\alpha\}) \rceil),$$  
    maximum number of nodes $$w^* = 6(d+ \lceil \alpha \rceil )N,$$ 
    number of nonzero parameters $$v^*= 141 (d+\alpha +1)^{3+d}N (m+6),$$ and all parameters (in absolute value) are bounded by $K^*= 1$.
    Here, we define a ``hard tanh'' function $\sigma_\tau:\RR \rightarrow [-1,1]$ for some $0<\tau\leq 1$ to be the linear combination of four scaled ReLU units given by  
\begin{align} \label{sigmatau}
    \sigma_\tau (x) &:= \sigma\left(\frac{x}{\tau}\right) - \sigma\left(\frac{x}{\tau} -1\right) - \sigma\left(-\frac{x}{\tau}\right)+\sigma\left(-\frac{x}{\tau}+1\right) 
    =
    \begin{cases} 
      1, & \text{if } x \geq \tau, \\
      \frac{x}{\tau}, & \text{if }  x\in [-\tau,\tau),\\
      -1, & \text{if } x < -\tau.
    \end{cases} 
\end{align}
    We define our hypothesis space $\mathcal{H}_\tau$ with parameter $\tau\in(0,1]$ to be functions generated by 
     \begin{eqnarray*}
         \mathcal{H}_\tau:= \text{span}\left\{\sigma_\tau \circ f: f\in \mathcal{F}(L^*,  w^*, v^*,K^*) \right\}.
     \end{eqnarray*}
 \end{definition}
 As a remark, we can see that if  $\tau$ is close to $0$, $\sigma_\tau (x)$ is close to $\text{sign}(x)$. In other words, we use the function $\sigma_\tau$ to approximate the sign function.

\subsection{Finding Classifier via Empirical Risk Minimization}
In practice, convex loss functions are commonly employed in classification tasks to make computation feasible.  
Throughout this paper, we adopt the well-known Hinge loss function  $$\phi(t):= \max \{0,1-t\}.$$
Learning a neural network classifier with Hinge loss is relatively straightforward owing to the gradient descent algorithm (see, e.g., \citep{molitor2021bias, george2023training}). 

Suppose we have a random set of normal data $T= \{(X_i,1)\}_{i=1}^{n}$, where  $\{X_i\}_{i=1}^{n}$ are drawn i.i.d. from an unknown distribution $Q$.  
 Additionally, we generate another i.i.d. training data set $\{X_i^\prime\}_{i=1}^{n^\prime}$ from a known measure $\mu$ (e.g., a Uniform distribution on $[0,1]^d$) and label each sample of it with $Y=-1$.
 
We merge $T= \{(X_i,1)\}_{i=1}^{n}$ and $T^\prime= \{(X_i^\prime,-1)\}_{i=1}^{n^\prime}$ together to form our training data set.
For any classifier $f: \mathcal{X} \to \RR$, we define its empirical risk w.r.t. the Hinge loss $\phi$ as 
\begin{equation}\label{ER}
    \varepsilon_{T,T^\prime} (f):=\frac{s}{n} \sum_{i=1}^{n} \phi\left(1 \cdot f(X_i)\right) + \frac{(1-s) }{n^\prime}\sum_{i=1}^{n^\prime}\phi(-1 \cdot f(X^\prime_i)). 
\end{equation}

Consider the hypothesis space $\mathcal{H}_\tau$ defined earlier in Definition \ref{hypothesis}. 
We aim to find the empirical risk minimizer (ERM) w.r.t. the Hinge loss $\phi$ in $\mathcal{H}_\tau$, which is given by
\begin{align}\label{ERM}
    \widehat{f}_{T, T^\prime,\phi} 
    &:= \arg \min_{f\in\mathcal{H}_\tau} \varepsilon_{T,T^\prime} (f) \nonumber\\
    &\stackrel{\mbox{\eqref{ER}}}{=} \arg \min_{f\in\mathcal{H}_\tau} \left\{ \frac{s}{n}\sum_{i=1}^{n}\phi(1 \cdot f(X_i)) + \frac{(1-s)}{n^\prime}\sum_{i=1}^{n^\prime}\phi(-1 \cdot f(X^\prime_i))\right\}. 
\end{align}

In the next section, we prove that the empirical risk minimizer $\widehat{f}_{T, T^\prime,\phi}$ of our chosen hypothesis space achieves good performance in classification-based anomaly detection. 
More specifically, we show in Theorem \ref{main1} that $R\left(\widehat{f}_{T, T^\prime,\phi}\right) - R^*$, which is equal to $R\left(\widehat{f}_{T, T^\prime,\phi}\right) - R (f_c)$, converges to $0$ with an optimal convergence rate in terms of $n$ and $n^\prime$.

\section{Theoretical Guarantee of ReLU Network Classifier}
\label{section:theoretical results}
We present our theoretical results in this section.

Consider the hypothesis space $\mathcal{H}_\tau$ defined earlier in Definition \ref{hypothesis} and the empirical risk minimizer $\Hat{f}_{T, T^\prime,\phi}$ given in \eqref{ERM}.
The following theorem establishes the convergence rate of the excess risk  $R(\text{sign}(\Hat{f}_{T, T^\prime,\phi}))-R(f_c)$.

\begin{theorem}\label{main1}
 Denote $n^* = \min\{n,n^\prime\}$.
 Let $d\in \NN, \alpha, r >0, 0 <s <1$, $n^* \geq 3$, and $\log (\max\{n,n^\prime\}) \leq \frac{n^*}{(\log(n^*))^3}$.
 Consider the hypothesis class $\mathcal{H}_\tau$ defined in Definition \ref{hypothesis} with $N =\left\lceil\left(\frac{n^*}{(\log n^*)^3\log (\max\{n,n^\prime\}) }\right)^{\frac{d}{d+ \alpha (q+2)}}\right\rceil , m= \left\lceil \left(1+ \frac{\alpha}{d}\right) \frac{\log N}{\log 2}\right\rceil,$ and $\tau = N^{-\frac{\alpha}{d}}$.  
 Let $\widehat{f}_{T, T^\prime,\phi}$ be the empirical risk minimizer w.r.t. $\phi$ in $\mathcal{H}_\tau$. Assume the Tsybakov noise condition (Assumption \ref{Tsybakov}) holds for some noise exponent $q\in [0,\infty)$ and constant $c_0>0$.
  If $N \geq \max \left\{(\alpha +1)^d, (r+1)e^d\right\}$, then for any $\delta>0 $, with probability $1-\delta$, there holds,
\begin{equation}
  R(\text{sign}(\Hat{f}_{T, T^\prime,\phi}))-R(f_c)  \leq \widetilde{C}\log \left(\frac{4}{\delta}\right) \left(\frac{(\log( n^*))^3 \log(\max\{n,n^\prime\})}{n^*}\right)^{\frac{\alpha (q+1)}{d+ \alpha (q+2)}},
\end{equation}
    where $\widetilde{C}$ is a positive constant independent of $n, n^\prime$ or $\delta$. 
\end{theorem}

The proof of Theorem \ref{main1} is given in Supplemental Material 6.
The theorem establishes that the excess risk decays at rate $\mathcal{O}\left((n^*)^{-{\frac{\alpha (q+1)}{d+ \alpha (q+2)}}}\right)$, up to logarithmic factors, where $n^* = \min\{n,n^\prime\}$ is the effective sample size determined by the smaller of the number of normal samples $n$ and synthetic anomalies $n^\prime$.
 Hence, the excess risk converges to zero as $n^* \to \infty$. The exponent $\frac{\alpha (q+1)}{d+ \alpha (q+2)}$
also reveals that smoother decision boundaries (larger $\alpha$) and higher noise level $q$ yield faster convergence, whereas higher dimensionality $d$ slows the rate. Specifically, when $q\to \infty$, the convergence rate approaches $\mathcal{O}((n^*)^{-1})$, up to logarithmic factors.

A key insight of this theorem lies in the dependence on $n^* = \min\{n,n^\prime\}$. Because the convergence rate is governed by $\min\{n,n^\prime\}$, insufficient synthetic data can bottleneck performance. In other words, for a given number of normal training samples $n$, it is beneficial to generate at least as many synthetic anomalies (i.e., $n^\prime \geq n$) to fully exploit the attainable rate of convergence. This provides, to our knowledge, the first theoretical guidance in the unsupervised AD literature on how the number of generated synthetic samples affects learning efficiency.

Later, in Section \ref{section:ablations}, we perform ablation studies to empirically examine the sensitivity of our neural network classifier to different choices of $n^\prime$ on real-world anomaly detection tasks, further validating the theoretical insight.

We note that 
it is proven in \citep{audibert2007fast, kim2021fast} that, when $\eta$ is $\alpha$-Hölder continuous, the minimax lower bound of the excess risk in binary classification is given by $\mathcal{O}\left( n^{-\frac{\alpha(q+1)}{d+\alpha(q+2)}}\right)$ in the i.i.d. case.
So, the convergence rate presented in Theorem \ref{main1} matches this optimal one, up to a logarithmic factor, because $n^*$ captures the minimum sample size across normal and anomalous training data. To our knowledge, this is the first result in the AD literature that
achieves the optimal rate.

Here, we emphasize the challenge of deriving excess risk bound for this classification problem.
The challenge arises due to the non-i.i.d. nature of the data. Our training data $T\cup T^\prime = \{X_i, 1\}_{i=1}^n \cup  \{X_i^\prime, -1\}_{i=1}^{n^\prime}$ are non-i.i.d. because $T$ and $T^\prime$ are drawn separately from different distributions. As a result, existing concentration inequalities commonly used in estimation error analysis cannot be applied. We derived novel concentration inequalities in our analysis (e.g., Lemma 6.4 and Lemma 6.7 in the Supplementary Material) to accommodate the non-i.i.d. nature of our data. For a detailed discussion of our novelty and theoretical contributions, please refer to  Supplementary Material 2.

Next, we apply the comparison theorem stated earlier in \eqref{compare} to our results in Theorem \ref{main1}. 
We recall the performance measure of density level set estimation given in \eqref{DLDmeasure}
\begin{equation}
    S_{\mu,h,\rho}(f)= \mu\left(\{f>0\}\Delta\{h>\rho\}\right),
\end{equation}
which measures how much the set $\{f>0\}$ coincides with the normal region $\{h>\rho\}$.
The following result presents a high probability upper bound for $S_{\mu,h,\rho}\left(\text{sign}\left(\Hat{f}_{T, T^\prime,\phi}\right)\right)$.
\begin{theorem}\label{main2}
Suppose the assumptions on Theorem \ref{main1} hold. Consider the density level set estimation performance measure given in \eqref{DLDmeasure}.
Let $\widehat{f}_{T, T^\prime,\phi}$ be the empirical risk minimizer w.r.t. $\phi$ in $\mathcal{H}_\tau$.
For any $\delta>0 $, with probability $1-\delta$, there holds,
\begin{equation}
    S_{\mu,h,\rho}(\text{sign}(\Hat{f}_{T, T^\prime,\phi}))  \leq c \left(\widetilde{C}\log \left(\frac{4}{\delta}\right) \right)^{\frac{q}{q+1}}\left(\frac{(\log( n^*))^3 \log(\max\{n,n^\prime\})}{n^*}\right)^{\frac{\alpha q}{d+ \alpha (q+2)}},
\end{equation}
where $c$ is some positive constant, $\widetilde{C}$ is a positive constant independent of $n$ or $\delta$. 
\end{theorem}
We can see that as $n^*$ increases, $S_{\mu,h,\rho}\left(\text{sign}\left(\Hat{f}_{T, T^\prime,\phi}\right)\right) \to 0$. This implies the trained ReLU network classifier induces a set $\left\{\text{sign}\left(\Hat{f}_{T, T^\prime,\phi}\right) >0\right\}$ that asymptotically converges to $\{h > \rho\}$, which represents the region of normal data under the underlying density $h$. In other words, although the training process involves contrasting normal samples with synthetic anomalies drawn randomly from $\mu$, the trained classifier provably learns the decision boundary of the normal region. 

Now that we have established $\left\{\text{sign}\left(\Hat{f}_{T, T^\prime,\phi}\right) >0\right\}$ which accurately estimate $\{h > \rho\}$, any data point lying outside this region is identified as anomalous.


\section{Network Intrusion Detection in Cybersecurity}
\label{section:experiments}

In this section, we conduct a series of real-data experiments to test our proposed method on real-world network intrusion datasets.
We outline the practical steps to find the ERM-network and, based on what we observed during practical implementations, share insights to further improve our model's performance in AD.
We also compare the performance of our method with other AD benchmark methods. 
Furthermore, we performed ablation studies to examine how altering the network architecture, loss function, and sampling method affects our model's performance. 

In the following, we report our choice of real-world datasets. 


\subsection{Real-World Cybersecurity Datasets for Evaluations}
\label{section:real_world_datasets}

\paragraph*{Dataset Overview}
To evaluate the performance of our method and other existing AD methods in network intrusion detection, we report results from real-world data.
We use NSL-KDD \citep{NSL_KDD_Dataset} dataset, a widely used cybersecurity benchmark for anomaly detection, and Kitsune \citep{kitsune_dataset} dataset, a recent network intrusion dataset introduced at NDSS, a top-tier security conference.
We detail our selection process in Supplementary Material 4.1.

The first set of experiments are conducted on the NSL-KDD dataset, which provides separate training and testing sets. Each sample in the dataset consists of $41$ features and a label indicating whether the network traffic is normal or belongs to one of the four types of cyber attack : “DoS” (Denial of Service), “Probe”,  “RA" (remote access), and “PE" (privilege escalatin). 
The data contains three types of features: basic features of a connection (e.g., protocol type, which is a categorical feature), content features within a connection (e.g., number of operations on access control files) and traffic features computed over a time window (e.g., number of connections to the current host in the past two seconds).
3 out of 41 features are categorical; after one-hot encoding the categorical features, the total dimension becomes 119.

The original NSL-KDD training set contains 125,973 samples, including both normal and attack samples. As per our unsupervised set-up, we remove all attack samples from the training data, leaving 67,343 normal samples. We then perform an $0.8/0.2$ split to create the training and validation sets. This results in 53,875 (real) samples for training and 13,468 (real) samples for validation. We note that the validation set contains no true anomalies.
The testing set consists of 22,544 samples, which include 9711 normal and a total of 12,833 attack instances, and is used solely for model evaluation.

The second sets of experiments are conducted on the Kitsune dataset. This dataset has a collection of nine sub-datasets, each with normal samples and a different types of cyber attack sample launched.
We select seven attacks (excluding two consistently yielding random performance by all models for brevity): SSDP flood, SSL renegotiation, SYN DoS, active wiretap, ARP MitM, fuzzing, and OS scan.
The data consists of 115 numerical features (none are categorical), with features on the bandwidth of outbound traffic, the bandwidth of outbound and inbound traffic, the packet rate of outbound traffic and the inter-packet delays of outbound traffic.
Each sample is labeled as either normal or as belonging to one of seven distinct attacks.

The Kitsune training set contains exactly 1,000,000 samples, all of which correspond to normal network traffic. Once again, we perform an $0.8/0.2$ split to obtain the training and validation sets, resulting in 800,000 (real) samples for training and 200,000 (real) samples for validation. As before, the validation set contains no true anomalies.
The test set contain a total of 923,216 attack samples and 355,473 normal samples.

\begin{table}[h]
    \centering
    \caption{Dataset details. Dimension is after one-hot encoding or feature extraction. 
        All data in the training set and validation set are normal, whereas testing data contains both normal and anomalies. Note that this table only counts the number of real data. }
    \begin{tabular}{c|ccccc}
    \toprule
    Dataset     & Data Type & Domain        & Dimension   & Num. Training/Validation & Num. Testing  \\
    \midrule
    NSL-KDD     & Tabular   & Cybersecurity & 119  & 53,875/13,468 & 22,544 \\
       Kitsune & Tabular   & Cybersecurity       & 115  & 800,000/200,000 & 1,278,689 \\
    Thyroid     & Tabular   & Medical       & 21   &  3083/342 & 3420\\
    MVTec       & Image     & Manufacturing & 1024 & 3266/363 & 1725\\
        \bottomrule
    \end{tabular}
    \label{tab:dataset_details}
\end{table} 

\paragraph*{Features}
In cybersecurity practice, experts design a set of rules or heuristics that describe normal network behavior, and they extract features (i.e., structured, measurable attributes) from network traffic data such as packet counts, connection durations, and protocol types. Both the NSL-KDD and Kitsune datasets are constructed in this way by experts: they both contains interpretable numerical features that machine learning models can process. As mentioned, the NSL-KDD dataset contains 119 features (after one-hot encoding), while the Kitsune dataset contains 115 features.

However, the extracted features are often tailored to capture known attack patterns, sometimes referred to as ``same-day attacks", which follow behaviors already observed and documented by security analysts.
In current practice, the greater challenge lies in detecting  ``zero-day attacks", which refer to new and previously unseen attacks that do not follow known behavioral patterns. Because such attacks exploit unknown vulnerabilities, their distinctive features are often absent from the previously-extracted features. For instance, some attacks depend on specific packet contents that are not included in these datasets (see \citet{features_ad_kdd_wenke, data_mining_ad_kdd_wenke, info_theoretic_ad_kdd_wenke, mcpad_one_class_ensemble_wenke} and Supplement E.1 for more information).

Nevertheless, models can sometimes identify these attacks indirectly by recognizing their side effects, such as abnormal traffic volumes or timing patterns. Attacks that cause clear and measurable disruptions, such as  DoS attacks (DoS in NSL-KDD, SSDP flood/SSL renegotiation/SYN DoS in Kitsune) and probe attacks (in NSL-KDD), tend to be more easily detected. Conversely, some attacks are more difficult to distinguish from normal traffic, as the available features in the dataset do not fully capture their distinctive and harmful characteristics (e.g., ARP MitM in Kitsune).

\subsection{Excess Risk Convergence Experiments and Synthetic Anomalies}
In this subsection, we conduct experiments using the  NSL-KDD dataset to validate our theory.
We minimize the empirical risk w.r.t. Hinge loss of a neural network model according to \eqref{ERM} by setting $s=0.5$ with real normal samples $X_i$ from $Q$ and synthetic anomalies $X_i'$ sampled i.i.d. from a uniform distribution on the domain $\mathcal{X} \subset [0,1]^d$ where $d=119$ (i.e., we consider $\mu$ to be the uniform measure on $\mathcal X$).
Here, we choose the uniform distribution as $\mu$ because it is the simplest and most unbiased choice. Since the true anomaly distribution is unknown and we have no reliable prior information about it, we deliberately avoided imposing additional assumptions or biases through alternative choices of $\mu$.

We do not have access to the density $h$ to obtain the label $Y|X$ as in \eqref{data}, but we instead use the ground truth labels provided in the dataset.

We normalize each variable to be in $[0,1]$ and convert categorical variables into one-hot vectors. 
Here, $Q$ represents the underlying distribution of normal network traffic data of NSL-KDD. 
We set $n^\prime=n$, that is, the number of synthetic anomalies in training equals the number of normal training samples. Such a choice of $n^\prime$ is motivated from Theorem \ref{main1} and \ref{main2}, which tell us that matching $n$ and $n^\prime$ would attain the best convergence of excess risk. Correspondingly, we set $s=0.5$ to be consistent with $n^\prime=n$. 
Then, we evaluate the neural network model on test data that has real benign samples (assumed to be from $Q$) and synthetic anomalies sampled i.i.d. from $\mu$.
We re-sample training anomalies and re-initialize models 5 times to obtain the mean and standard deviation of model performance.
We increase the amount of normal data $n$ (and synthetic data $n'$) from 2 to the full training dataset size to test if the accuracy of the learned model converges to an asymptote which should represent the Bayes classifier.

Below, we outline some procedures for implementing our experiments. We then present our experiment results and some discussions. 

\subsubsection{Practical Implementation}
\label{section:practical_implementation}

Previously, we defined the target neural network function class in Definition \ref{hypothesis}. As a quick recap, our target function class consists of ReLU feed-forward networks with a sparsity constraint, and the absolute value of all parameters is bounded by $1$.
In practice, searching for the empirical risk minimizer in this function class is not straightforward.
We proceed to detail some procedures to facilitate the search.

We adopt feed-forward neural networks with depth $D:=L+1=27$ and width $w=678$. 
We experimented with networks of varying depths and widths, and reported the results achieved by the best-performing configuration.
The empirical risk minimizer in \eqref{ERM} is computed by minimizing the hinge loss using Adam \citep{kingBa15} with learning rate $10^{-3}$ and mini-batch size $1024$.
We also employ weight decay with Adam to model sparsity constraints that all parameters are bounded by 1 (in absolute value) and constraints on the number of nonzero parameters. 
We train models until convergence (which, in code, corresponds to 200 epochs with early stopping on the validation loss).
For all experiments, the validation set is used to tune model hyperparameters and to apply early stopping when appropriate\footnote{Our code implementation is available on GitHub \url{https://github.com/mattlaued/Optimal-Classification-Based-Unsupervised-Anomaly-Detection}}.

\subsubsection{Results and Discussion}
\label{section:convergence_discussion}
\begin{figure}
    \centering
    \begin{subfigure}{0.45\linewidth}
    \centering
    \includegraphics[width=\linewidth]{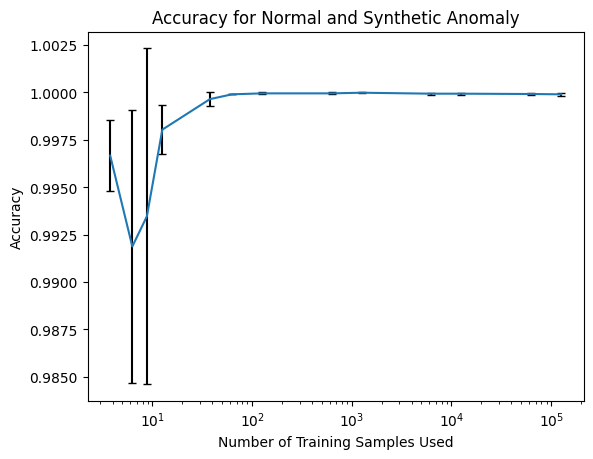}
    \caption{Real normal data and synthetic anomalies (uniformly drawn) with anomaly test ratio of $s=0.5$. Threshold is uncalibrated.}   
    \label{fig:kdd_normal_synthetic}   
    \end{subfigure}
    \hspace{5mm}
    \begin{subfigure}{0.45\linewidth}
    \centering
    \includegraphics[width=\linewidth]{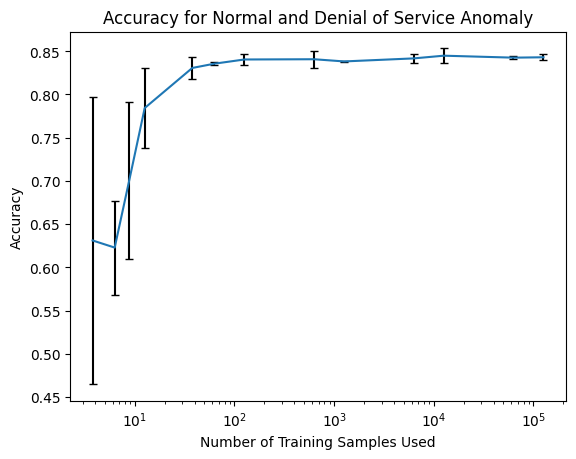}
    \caption{Real normal data and denial of service attacks (real anomalies) with calibrated threshold. Accuracy curves are similar for other attacks.}    
    \label{fig:kdd_normal_dos}    
    \end{subfigure}
    \caption{Accuracy on NSL-KDD network intrusion dataset, with various test data. Mean and standard deviation across 5 runs plotted. Accuracy converges with more training samples.}
    \label{fig:kdd_convergence_rate}
\end{figure}

We show our experiment results in Figure \ref{fig:kdd_convergence_rate}. We present the accuracy curve of our ERM-network classifier on the NSL-KDD dataset as the number of training samples increases. In short, the accuracy converges to an asymptote as the number of training samples increases, which corroborates Theorem \ref{main1}. 
We give more detailed discussions below on how we address two challenges observed in our experiments to obtain the final result. 

\paragraph*{First Insight: Size Matters for Vanishing Gradients and Overfitting}
In our initial attempts, the ERM-network classifier fails to converge across 200 training epochs during training, suggesting vanishing gradients.
Reducing the depth of the network resolves this issue, but exposes another difficulty of generalization.
For relatively deep and wide networks, the validation loss of the models can be decreased to zero without knowledge of its generalizability.
Further experiments show that this phenomenon holds even when the size of the network is much smaller than the conditions stated in Theorem \ref{main1} and \ref{main2}.
With hinge loss, this behavior is not a surprise --- the loss can decrease to zero once the margin is large enough.
There are two practical solutions we can adopt to address this overfitting.
First, we propose structural risk minimization --- reducing the complexity of the model as much as possible while maintaining high validation performance.
In addition to weight decay, we also reduce the depth to $D=3$ and width to $w=500$.
Second, we swap ReLU out for leaky ReLU activations to avoid the dead neuron problem.

Implementing these changes yields models that converge.
We plot the accuracy curve (accuracy is one minus misclassification error) in Figure \ref{fig:kdd_normal_synthetic}. 
There is high variance with few training samples $n$, but we can see that the mean accuracy converges to an asymptote as the number of training data ($n$ and $n^\prime$) increases. 
This validates our findings in Theorem \ref{main1}, which states that the excess risk goes to $0$ as $\min\{n, n^\prime\}$ increases.



However, we should note that real anomalies likely do not follow $\mu$ (the uniform distribution in our experiments).
Thus, we also evaluate our method on benign data versus each attack found in the test dataset of NSL-KDD (i.e., real anomalies). We share some findings in the following. 

\paragraph*{Second Insight: Overfitting to Synthetic Anomalies with Uncalibrated Thresholding}
Intriguingly, the accuracy of our model on real anomalies decreases with more training data (i.e., higher $n$) for 2/4 of the attacks.
In the original formulation in \eqref{DLDmeasure}, we mandated that the prediction $f(x)>0$ classifies $x$ as normal.
Although predictions of synthetic anomalies can be very negative and far away from the positively-predicted normal data, 
the prediction of real anomalies may not be so far away, with many attacks having positive predictions $f(x)>0$ and wrongly classified as anomalies.
From a feature learning perspective, it is possible that the model learns what makes data realistic, similar to what the discriminator in a generative adversarial network (GAN) learns.
Hence, since real anomalies have features that make them realistic, the model may not be able to make as accurate predictions compared to synthetic anomalies.
In this case, the model ``overfits'' to the synthetic anomalies and degrades in performance when facing covariate shift towards real anomalies.
An improvement is to choose a threshold $\kappa >0$ for the classification rule $f(x)>\kappa$ rather than the midpoint $\kappa:=0$.
In practice, it is quite common to set such a threshold to allow for $\beta$ false positives to be more ``conservative'' to detect more anomalies.
We choose $\beta=5\%$ and calibrate the threshold with the validation data.
Often times, $\kappa$ is well above 0, which reflects the conservativeness of the modeller.

Changing the threshold for classification yields the accuracy curves of all 4 attacks and the synthetic anomalies similar to the convergence rate in \eqref{main2}. 
We plot one example in Figure \ref{fig:kdd_normal_dos}.
We proceed to use this improved model in the following experiments and refer to it as our theory classifier (TC).



\subsection{Evaluation with Anomaly Detection Metrics}
In this subsection, we compare our method with other existing AD methods on the NSL-KDD dataset and the Kitsune dataset. 


\paragraph*{Metrics}
In AD, evaluations usually defer to metrics other than accuracy and area under the receiver operator characteristic curve (AUROC) because (1) negatives and positives are not necessarily equal in number and (2) predicting wrongly on each of them have different costs.
Hence, we use the area under the precision-recall curve (AUPR), also known as average precision, as our main metric (instead of accuracy, as in the previous section).
AUPR allows us to measure the separation between the positive (anomaly) and negative (normal) class without a specific threshold, which can be later set by the operator based on desired trade-offs between Type I and II errors.
We repeat experiments five times and report the mean and standard deviation.

\paragraph*{Models}
For each dataset, we compare our theory classifier with other classification-based AD methods, which includes the SVM baseline  and NS-NN \citep{sipple20NSNN}.
We adopt SVM equipped with the Radial Basis Function (RBF) kernel and trained with $\ell^2$ regularization. The model’s hyperparameters are optimized during validation using structural risk minimization.
NS-NN is a neural network-based method.
A key conceptual difference between NS-NN and our approach lies in the way synthetic anomalies are generated: NS-NN produces a substantially larger number of synthetic anomalies (it requires $n^\prime = 30n$) and samples them from the continuous space rather than from the support of the underlying data distribution. Notably, no theoretical guarantees have been established for NS-NN.
In this section, we report results for NS-NN with 2 hidden layers. Additional experiments with 3 and 4 hidden layers are provided in Supplementary Material 4.3. Empirically, the two-layer configuration generally outperforms deeper variants (3/4 layers).

We also include results from a random baseline and other unsupervised AD baselines that are not classification-based.
The random baseline is calculated in expectation for random guessing, which is equivalent to the fraction of anomalies in the test data.
The unsupervised benchmarks considered are OCSVM \citep{ocsvm}, Isolation Forest \citep{IsolationForest} and DeepSVDD \citep{deep_svdd}. Although these methods lack theoretical guarantees, they have demonstrated strong empirical performance in prior studies. 

Hyperparameters for all models are selected using validation loss when applicable, or by structural risk minimization as described in Section \ref{section:convergence_discussion}. Further implementation details are provided in Supplementary Material 4.2.

We report the AUPR (mean $\pm$ standard deviation, over five replicates) of normal data against each cyber-attack in each column of Table \ref{tab:kdd} for NSL-KDD and Table \ref{tab:kitsune} for Kitsune, where attacks are considered as anomalies.



\begin{table}[]
    \centering
    \caption{AUPR for KDD, split across attacks in each column. 
    Higher is better, best for each category in bold.
    Our theory classifier (TC) outperforms the other classification-based approaches SVM and NS-NN. Across all methods, our TC is the best in detecting probe attack, and is second best in detecting DoS, PE, and RA. 
    }
    \begin{tabular}{cl|cccc}
    \toprule
        &Model$\backslash$Attack & DoS & Probe & PE & RA \\
         \midrule
        &Random & 0.431 & 0.197 & 0.007 & 0.218 \\
        \midrule
        \parbox[t]{2mm}{\multirow{3}{*}{\rotatebox[origin=c]{90}{Classify}}} & SVM & 0.890$\pm$0.002 & 0.836$\pm$0.001 & 0.013$\pm$0.000 & 0.319$\pm$0.001 \\
        &TC (Ours) & \bf{0.907$\pm$0.009} & \bf{0.937$\pm$0.009} & \bf{0.028$\pm$0.007} & \bf{0.389$\pm$0.030} \\
        &NS-NN & 0.753$\pm$0.015 & 0.286$\pm$0.067 & 0.003$\pm$0.000 & 0.109$\pm$0.000 \\
        \midrule
        \parbox[t]{2mm}{\multirow{3}{*}{\rotatebox[origin=c]{90}{Unsup.}}}&OCSVM & 0.802$\pm$0.091 & 0.687$\pm$0.141 & 0.025$\pm$0.022 & 0.262$\pm$0.095 \\
        &IsoF & \textbf{0.942$\pm$0.006} & \textbf{0.923$\pm$0.049} & \textbf{0.045$\pm$0.014} & 0.398$\pm$0.028 \\
        &DeepSVDD & 0.860$\pm$0.003 & 0.848$\pm$0.004 & 0.019$\pm$0.000 & \textbf{0.515$\pm$0.006} \\
         \bottomrule
    \end{tabular}
    \label{tab:kdd}
\end{table}

\begin{table}[]
    \centering
    \caption{AUPR for Kitsune, split across attack types. 
    Higher is better, best for each category in bold.
    Our theory classifier (TC) is best on 3/7 attacks among all methods, achieving non-trivial performance for attacks that others struggle to do better than random (ARP MitM and fuzzing).
    }
  \begin{adjustbox}{width=\textwidth}
    \begin{tabular}{cl|ccccccc}
    \toprule
        &Model$\backslash$Attack & SSDP & SSL & SYN & Wiretap & ARP MitM & Fuzzing & OS Scan \\
         \midrule
        &Random & 0.468 & 0.077 & 0.004 & 0.217 & 0.761 & 0.348 & 0.094 \\
                \midrule
        \parbox[t]{2mm}{\multirow{3}{*}{\rotatebox[origin=c]{90}{Classify}}} & SVM & \textbf{1.000$\pm$0.000} & \textbf{0.729$\pm$0.001} & 0.146$\pm$0.007 & 0.654$\pm$0.054 & 0.682$\pm$0.014 & 0.345$\pm$0.164 & \textbf{0.744$\pm$0.001} \\
        &TC (Ours) & {{0.997$\pm$0.005}} & {0.565$\pm$0.128} & 0.198$\pm$0.093 & \textbf{0.756$\pm$0.095} & \textbf{{0.821$\pm$0.083}} & \textbf{0.582$\pm$0.106} & 0.497$\pm$0.007 \\
        &NS-NN & 0.999$\pm$0.000 & 0.649$\pm$0.002 & \bf{0.242$\pm$0.000} & 0.723$\pm$0.000 & 0.801$\pm$0.010 & 0.360$\pm$0.000 & 0.500$\pm$0.001 \\
        \midrule
        \parbox[t]{2mm}{\multirow{3}{*}{\rotatebox[origin=c]{90}{Unsup.}}} & OCSVM & 0.989$\pm$0.014 & 0.517$\pm$0.035 & 0.228$\pm$0.015 & \bf{0.695$\pm$0.011} & 0.653$\pm$0.009 & 0.252$\pm$0.008 & 0.502$\pm$0.000 \\
        &IsoF & 0.993$\pm$0.008 & 0.498$\pm$0.021 & 0.212$\pm$0.047 & 0.691$\pm$0.013 & 0.651$\pm$0.012 & \bf{0.254$\pm$0.010} & \textbf{0.502$\pm$0.001} \\
        &DeepSVDD & \bf{0.998$\pm$0.000} & \textbf{0.642$\pm$0.000} & \bf{0.291$\pm$0.003} & {0.623$\pm$0.001} & \bf{0.690$\pm$0.003} & 0.246$\pm$0.001 & {0.500$\pm$0.000} \\
         \bottomrule
    \end{tabular}
  \end{adjustbox}
    \label{tab:kitsune}
\end{table}

\paragraph*{Overall Results}
Across all methods, our theory classifier is the best in probe attack and second best in DoS, PE, and RA for NSL-KDD, and is competitive in 5/7 attacks for Kitsune (SSDP, SYN, Wiretap, ARP MitM, fuzzing).
Among the classification-based approaches, our theory classifier has the best AUPR for all attacks in NSL-KDD and 3/7 attacks in Kitsune.
%

Meanwhile, the unsupervised methods have mixed results. Isolation Forest performs the best on 3/4 attacks in NSL-KDD overall, but it performs worse than OCSVM on 5/7 attacks in Kitsune. DeepSVDD performs consistently competitive in Kitsune among the unsupervised methods. 

Overall, our theory classifier seems to consistently perform above the average.
To provide more insight, we proceed to compare our theory classifier to classification-based approaches for attacks that are easy or difficult to detect, based on the features present in the data.

\paragraph*{Easily detectable attacks}
As mentioned in Section \ref{section:real_world_datasets}, attacks in NSL-KDD and DoS in Kitsune are easier to detect, because the features present in the dataset directly correspond to the features of the attacks.
In other words, the attacks in NSL-KDD and the SSDP flood, SSL renegotiation and SYN flood in Kitsune have conditional probability $\eta(X):=P(Y=1|X)$ closer to 0 or 1 for the normal and attack data.
For these 7 attacks, we see our theory classifier having the best AUPR among classification-based detectors for 4 attacks (DoS, probe, RA, PE), and competitive on SYN, while SVM has the best AUPR for SSDP and SSL.
Our theory classifier has almost perfect AUPR for SSDP attacks, and it has somewhat unstable learning for SSL attacks evidenced by the high standard deviation.
Nevertheless, our theory classifier is the overall best for these class of attacks.

\paragraph*{Difficult attacks}
The remaining 4 attacks --- Wiretap, ARP MitM, Fuzzing, and OS Scan from the Kitsune dataset --- are considered difficult to detect, as their malicious behavior does not manifest through features directly captured in the dataset. In other words, the distinctive characteristics that make these attacks harmful are only weakly reflected in the features in the dataset.

For the Wiretap attack, all methods seem to be able to exploit side-channel information to some extent, with our theoretical classifier achieving the best performance. In contrast, most AD baselines struggle with the ARP MitM and Fuzzing attacks, where our theory classifier attains a substantially higher AUPR, well above random guessing. Overall, our method achieves the best results on 3 out of the 4 difficult-to-detect attacks, demonstrating its robustness in detecting anomalies whose patterns are not effectively captured by past observations.

\subsection{Ablations}
\label{section:ablations}

To understand how each piece of our implementation of the theory classifier contributes to its overall performance, we perform ablations on the NSL-KDD dataset on the width, depth, activation, loss function, weight decay, and number of synthetic anomalies.
To do this, we switch out each component (while the other components are the same as the original TC) to gather further insights.
Our results are in presented Table \ref{tab:kdd_ablations}.
\begin{table}[]
    \centering
    \caption{AUPR for KDD, split across attacks in each column.  The table reports the mean AUPR (
± standard deviation) over five replicates. Higher is better. We ablate our theory classifier (TC) across 
    width, depth, activation, loss, sampling proportion and method.
    Results underlined are when the performance is lower than 1 standard deviation from the original TC, while bold is for increased performance.
    Wider and shallower models are better, while changing leaky ReLU to ReLU has insignificant effect.
    Our choice of loss function and $n^\prime$ also generally improve results --- swapping them out reduces the performance of models.
    }
  \begin{adjustbox}{width=\textwidth}
    \begin{tabular}{cl|cccc}
    \toprule
        Ablation & Model & DoS & Probe & RA & PE \\
        \midrule
        Original & TC & 0.907$\pm$0.009 & 0.937$\pm$0.009 & 0.028$\pm$0.007 & 0.389$\pm$0.030 \\
        \midrule
        \midrule
\multirow{5}{*}{Width $w$}  & XL ($w=1000$) & 0.908$\pm$0.008 & 0.940$\pm$0.007 & 0.026$\pm$0.003 & 0.387$\pm$0.020 \\
 & L ($w=678$) & 0.907$\pm$0.009 & 0.937$\pm$0.009 & 0.028$\pm$0.007 & 0.389$\pm$0.030  \\
& S ($w=335$) & 0.904$\pm$0.006 & 0.937$\pm$0.007 & 0.027$\pm$0.004 & 0.390$\pm$0.027 \\
& XS ($w=119$) & \underline{0.877$\pm$0.012} & \underline{0.911$\pm$0.016} & 0.021$\pm$0.004 & 0.360$\pm$0.041 \\
& XXS ($w=59$) & \underline{0.865$\pm$0.012} & \underline{0.891$\pm$0.021} & 0.022$\pm$0.001 & 0.363$\pm$0.022 \\
\midrule
\multirow{6}{*}{Depth $D$}  & 2 layers & \underline{0.885$\pm$0.010} & \underline{0.899$\pm$0.024} & \underline{0.016$\pm$0.003} & \underline{0.343$\pm$0.018} \\
 & 4 layers & 0.902$\pm$0.010 & 0.938$\pm$0.005 & 0.029$\pm$0.007 & 0.398$\pm$0.020 \\
 & 5 layers & \underline{0.891$\pm$0.010} & \underline{0.905$\pm$0.040} & 0.021$\pm$0.007 & 0.367$\pm$0.023 \\
 & 8 layers & \underline{0.890$\pm$0.026} & \underline{0.903$\pm$0.051} & 0.021$\pm$0.004 & 0.366$\pm$0.022 \\
 & 11 layers & 0.908$\pm$0.011 & \underline{0.921$\pm$0.034} & 0.024$\pm$0.014 & \underline{0.347$\pm$0.042} \\
& 27 layers & \underline{0.431$\pm$0.000} & \underline{0.197$\pm$0.000} & \underline{0.007$\pm$0.000} & \underline{0.218$\pm$0.000} \\
\midrule
Activation & ReLU & 0.910$\pm$0.007 & 0.937$\pm$0.002 & 0.030$\pm$0.007 & 0.391$\pm$0.024 \\
\midrule
\multirow{2}{*}{Loss} & Logistic Loss & \underline{0.850$\pm$0.009} & \underline{0.735$\pm$0.021} & \underline{0.006$\pm$0.003} & \underline{0.111$\pm$0.002}\\
 & No Weight Decay & \underline{0.878$\pm$0.017} & \underline{0.882$\pm$0.014} & \underline{0.019$\pm$0.003} & \underline{0.308$\pm$0.011} \\
\midrule
\multirow{5}{*}{Sampling anomalies} & $n'=2n$ & 0.901$\pm$0.006 & 0.932$\pm$0.009 & 0.025$\pm$0.003 & 0.379$\pm$0.017 \\
& $n'=5n$ & \underline{0.889$\pm$0.007} & \underline{0.920$\pm$0.014} & \underline{0.020$\pm$0.003} & 0.362$\pm$0.019 \\
& $n'=10n$ & \underline{0.854$\pm$0.041} & \underline{0.825$\pm$0.065} & \underline{0.014$\pm$0.002} & \underline{0.332$\pm$0.050} \\
& $n'=30n$ & \underline{0.799$\pm$0.063} & \underline{0.733$\pm$0.088} & \underline{0.013$\pm$0.003} & \underline{0.280$\pm$0.079} \\
 & Continuous & \underline{0.890$\pm$0.009} & \underline{0.811$\pm$0.046} & 0.032$\pm$0.005 & \bf{0.442$\bm \pm$0.017} \\

         \bottomrule
    \end{tabular}
    \end{adjustbox}
    \label{tab:kdd_ablations}
\end{table}

Our first observation is that
\textbf{model expressiveness matters}.
We varied the model width between $w\in \{59,119,335,500,678,1000\}\approx \{0.5d, d, 3d, 4d, 5.5d, 8.5d\}$, which we nickname as XXS/XS/S/M/L/XL sizes.
The performances of the larger models with at least S width ($w\geq 335$) are comparable, but those with XS/XXS width (119/59, where $w\leq d$) lag behind.
This validates our intuition in Section \ref{section:convergence_discussion}, which are based on our theoretical results in Theorem \ref{main1} and Theorem \ref{main2}, that neural network models should be large enough for expressivity.

However, our second observation is that \textbf{gradient signal matters}.
Performance remains similar (fluctuates up and down minutely) across 2 to 11 layers, with the 27-layer model having symptoms of vanishing gradients during training and no meaningful gradients, resulting in random performance.
It appears that models that are shallow enough to avoid vanishing gradients are good, while a larger width can compensate for the expressivity lost with shallower models.
With a relatively shallow network that provides good gradient signals, swapping out ReLU with leaky ReLU does not have a significant impact on the performance.
Meanwhile, changing out the loss from hinge loss to logistic loss seems to significantly affect the gradient signal.
In fact, models trained with logistic loss have the greatest performance decline across the changes we make.
It is likely that logistic loss, an asymptotic loss, does not go exactly to zero, so our method of choosing hyperparameters with hinge loss does not work well for models trained with logistic loss.

The third observation is that \textbf{our theory provide practically helpful guidance}, because the implementation we outline in Section \ref{section:practical_implementation} mostly contributes to the empirical performance.
Recall that Theorem \ref{main1} tells us that the excess risk of the ReLU network classifier decays to $0$ at a rate $$\mathcal{O}\left((n^*)^{-{\frac{\alpha (q+1)}{d+ \alpha (q+2)}}}\right),$$ up to logarithmic factors, where $n^*= \min\{n,n^\prime\}, d\in \NN, \alpha,q>0$. Hence, the excess risk converges to zero as $n^* \to \infty$. Consequently, Theorem \ref{main2} shows that the error of AD using the ReLU classifier decays at a rate $$\mathcal{O}\left((n^*)^{-\frac{\alpha q}{d+ \alpha (q+2)}}\right).$$
Our theoretical results tell us that given $n$, it is beneficial to generate synthetic anomalies at least as many as $n$, that is, set $n^\prime = n$. Beyond matching $n$, however, further increasing 
$n^\prime$ would not improve the asymptotic rate, as the bound is governed by $\min\{n,n^\prime\}$.
Our ablation results are consistent with our theoretical findings, show that the best AUPR is achieved when $n=n^\prime$ across four types of cyber-attacks, which is adopted in our theory classifier (TC). As $n^\prime$ increases beyond $n$, the accuracy initially remains stable. However, as $n^\prime$  becomes excessively large (e.g., $ n' = 30n $), the accuracy decline across all anomaly types. This confirms that generating synthetic anomalies comparable in number to the real samples $n$ is sufficient and that excessively large $n^\prime$ offers no additional benefit.

Meanwhile, sampling anomalies continuously in the ambient space (i.e., a categorical variable encoded as a one-hot vector can now take fractional values) leads to a drop in AUPR for DoS and probe attacks but increase in AUPR for PE attacks, which does not seem to provide conclusive results on this piece yet.

Additionally, we observe that applying weight decay during neural network training yields a substantially higher AUPR compared to training without it. This finding aligns with our theoretical insights, which introduce sparsity for achieving better generalization. While our theoretical hypothesis space constrains the number of non-zero parameters (typically enforced through $L_0$ or $L_1$ regularization), in practice, we employ an $L_2$ regularization (weight decay) instead. Despite this difference, weight decay effectively promotes sparsity by driving some parameters toward zero, while also stabilizing optimization and reducing training costs.

\subsection{Additional Applications in Industrial and Medical Anomaly Detection}
In addition to the numerical results on network intrusion detection, we provide additional experimental results on industrial image inspection and medical anomaly detection in Supplementary Material Section 3. These additional experiments demonstrate that our proposed method is not specific to cybersecurity applications, but can also be applied to other unsupervised anomaly detection settings where abnormal samples are scarce, difficult to anticipate, or costly to obtain. 

\section{Discussions}\label{section:discussion}

Throughout this work, we have focused on addressing the anomaly detection problem when no real anomalies are available. Our framework trains a neural network model using only normal samples, augmented with synthetic anomalies drawn from a reference distribution, to learn the decision boundary that defines the normal region. 

As technology advances and real data collection becomes less costly, sometimes we have access to real anomalous samples. In such cases, it is natural to leverage these labeled anomalies to further guide model training. Our framework can be naturally extended to this setting, where both normal and anomalous samples are present.

In a concurrent study \citep{lau2025bridging}, our framework has been extended to compare the performance of neural network models trained under two regimes:
\begin{enumerate}
    \item using only normal samples and real anomalies (a supervised approach), and
\item using normal samples, real anomalies, and synthetic anomalies (a semi-supervised approach).
\end{enumerate}
The key finding of this study can be summarized as one sentence:
``Even in the presence of real anomalies, including synthetic anomalies during training can further improve the accuracy of neural network models in detecting real anomalies." This finding is not only supported by numerical evidences, but also theoretical proof. We see that our framework is flexible and can be extended to other kinds of AD problems.


\section{Related Works}\label{section:related}

In this work, we explored the modeling aspect of classification-based approaches for unsupervised anomaly detection.
To do this, we viewed unsupervised anomaly detection as a density level detection problem, which is common in the literature \citep{min_vol_svm_davenport, dbscan, min_vol_set_scott_jmlr}.
Other works also generate synthetic data to use a classifier for anomaly detection, such as using RIPPER (inductive decision tree learner) in \citet{kdd_artificial_anom_wenke}, random forest and neural networks in \citet{sipple20NSNN} and SVM in \citet{steinwart2005classification}.
\citet{steinwart2005classification} is the only work proving excess risk convergence.

Apart from the binary classification-based approaches, there are many other approaches to unsupervised anomaly detection.
Recent advances in deep learning have led to the emergence of three major paradigms for AD, each reflecting a distinct modeling philosophy: probabilistic modeling, reconstruction learning, and one-class classification. 
These approaches differ in how they define what is “normal” and how they infer deviations from it, but all rely on the expressive capacity of deep neural networks to model complex data distributions.

In the following, we summarize these approaches in greater detail and highlight  representative techniques within each category.

\subsection{Reconstruction-based Methods}
Reconstruction-based methods operate on the intuition that a model trained only on normal data should accurately reconstruct normal samples, but fail to do so on anomalous samples.

An autoencoder \citep{hinton2006reducing} is a neural network composed of an encoder that compresses input data into a latent representation (typically lower-dimensional) and a decoder that reconstructs the input from this representation.  These models are trained solely on normal samples, learning to reproduce their underlying patterns and structures. During inference, normal inputs are reconstructed accurately, whereas anomalous inputs yield higher reconstruction errors due to their deviation from the learned data manifold \citep{sakurada2014anomaly,zhao2017spatio,zhou2017anomaly,chen2018autoencoder}.
The  reconstruction error, measured as the difference between the input and the model's reconstructed output, serves as the anomaly score. A new sample is flagged as anomalous if its reconstruction error exceeds a predefined threshold.

The Variational Autoencoder (VAE) \citep{kingma2013auto} extends this idea by introducing a probabilistic latent space with a known prior distribution (typically Gaussian). A VAE identifies anomalies either by their high reconstruction error or low likelihood under the learned generative model \citep{an2015variational, gong2019memorizing, memarzadeh2020unsupervised}. In modern AD practice, VAEs and their numerous variants dominate this category.

A recent comparative study by \citet{nguyen2024variational} evaluates several VAE-based anomaly detectors on image datasets: a standard VAE, a VAE with a Gaussian Random Field prior (VAE-GRF) \citep{gangloff2024variational}, and a VAE incorporating a vision transformer (ViT-VAE) \citep{lee2022anovit, zhang2023exploring}. The standard VAE employs a simple isotropic Gaussian prior $\mathcal{N}(0,I)$. The VAE-GRF replaces this with a GRF prior $\mathcal{N}(0,K_\theta)$, where $K_\theta$
is a covariance matrix defined by a spatial kernel (e.g., Matérn kernel). The ViT-VAE, on the other hand, replaces the convolutional neural network-based encoder of a standard VAE with a transformer architecture that employs self-attention layers. \citet{nguyen2024variational} report that the ViT-VAE consistently outperforms the baseline VAE  even with
a limited number of training epochs, whereas the VAE-GRF underperforms unless carefully tuned, possibly due to its sensitivity to kernel parameters.

\subsection{Deep Density Estimation}
Deep density estimation approaches AD as a probabilistic modeling problem. 
This method employs neural networks to  estimate the underlying probability distribution of normal data and identify anomalies as samples that fall in low-probability regions of the learned distribution.
Unlike classical density estimators (e.g., histograms, kernel density estimators) that often struggle in high-dimensional spaces, neural network-based models such as VAEs, Generative Adversarial Networks (GANs) \citep{goodfellow2014generative}, and Normalizing Flows (NFs) \citep{kobyzev2020normalizing} can yield reliable density estimates in high-dimensional domains.
Some notable work in this category includes \citep{reed2017parallel, zong2018deep, liu2021density, liu2022unsupervised, rosenhahn2024quantum, cook2024feature}.

Among these, NFs have received particular attention for their ability to provide exact and tractable likelihoods. NFs are neural networks that learn a bijective mapping between the input data space and a latent space with a simple known distribution (usually Gaussian) \citep{dinh2017density,kingma2018glow, rosenhahn2024quantum}. In AD, such a neural network is trained only on normal samples by maximizing the likelihood of those samples. We can then compute the exact likelihood (that serve as anomaly scores) of any data point using the learned function. Samples with low likelihoods under the learned model are flagged as anomalies.

In contrast, GAN-based approaches perform AD without explicitly computing probability densities or likelihoods for data samples \citep{liu2021density, liu2022unsupervised, schlegl2019f}.
In a typical GAN framework, a generator network maps latent variables sampled from a simple prior distribution to synthetic data resembling the training (normal) data, while a discriminator is trained adversarially to distinguish real from generated samples. 
Through this process, the generator implicitly learns the manifold of normal data.
During inference, a test sample is projected onto this learned manifold. 
Samples that cannot be well reproduced by the generator (those yielding large reconstruction errors) are considered anomalous.

Overall, deep density estimation provides a principled, probabilistic foundation for AD, combining the expressive power of neural networks with the statistical rigor of likelihood-based inference.

\subsection{One-Class Classification}
One-class classification approaches are also widely used for unsupervised AD. Given only normal data points during training, these methods seek to find a region or boundary in the feature space that encloses the majority of the data. Any new sample lying outside this learned boundary is considered an anomaly.

Among the earliest and most influential one-class classification techniques is the One-Class Support Vector Machine (OCSVM) \citep{ocsvm, chen2001one, manevitz2001one},  a kernel-based approach rather than a deep learning model. 
The OCSVM maps input data into a high-dimensional feature space using a kernel function and then constructs a hyperplane that maximally separates the data from the origin. The origin represents the anomaly class. By maximizing the margin between the normal data and the origin while allowing some soft violations, the OCSVM effectively learns a decision boundary for normality.

Building on a similar intuition, the Support Vector Data Description (SVDD) \citep{SVDD} method defines a minimal hypersphere that encloses most of the normal data points in the transformed feature space, treating points outside the sphere as anomalies. More recently, Deep SVDD \citep{deep_svdd} (and its variants \citep{dohsc_ad, DevNet_AD}) extends this concept by employing deep neural networks to learn nonlinear feature transformations, enabling the model to form a compact hypersphere in a latent representation space.
By leveraging deep feature learning, Deep SVDD can capture complex, high-dimensional data structures, thus enhancing the performance of one-class classification in domains such as image data \citep{yi2020patch}

Complementary work has been done to explore the efficacy of sampling methods for classification-based approaches.
In particular, sampling around the manifold of normal data for more realistic synthetic anomalies has been explored by \citet{min_vol_svm_davenport} and \citet{kdd_artificial_anom_wenke} for tabular data and by \citet{outlier_exposure} for images.
These methods can be integrated into our approach and are not mutually exclusive to our work.

\section{Conclusion 
}\label{section:conclusion}
Motivated by the increasing occurrence of  ``zero-day attacks" in cybersecurity, this work proposed a neural network-based framework for anomaly detection that operates without prior knowledge of the anomalies it is expected to encounter.
By formulating AD as a binary classification problem through density level set estimation, our method trains ReLU neural network with supervision from synthetic anomalies.
The proposed method achieves an optimal excess risk bound, marking a novel theoretical contribution to the AD literature. 
Moreover, it provides concrete guidance on selecting an appropriate ratio between normal and synthetic anomaly samples during training.

Empirical evaluations across diverse domains --- including cybersecurity, industrial inspection, and medical decision-making --- demonstrate our method’s consistent robustness and competitiveness.
These experiments not only validate the theoretical insights but also offer practical guidance for improving empirical risk minimization of neural networks in real-world settings.
Notably, in the context of network intrusion detection, our approach significantly enhances the detection of difficult and previously unseen attack patterns compared to other existing classification-based detectors.

Ultimately, this work bridges theoretical rigor and practical utility in AD. By grounding neural anomaly detection in a statistical framework and validating its effectiveness across in various applications, our study paves the way for more principled and effective defenses against emerging threats, especially in rapidly evolving cybersecurity landscapes.

\section*{Fundings} 
This work has been supported by the National Science Foundation (NSF) AI Institute for Agent-based Cyber Threat Intelligence and Operation (ACTION) under the NSF grant number IIS-2229876. 
Huo and Zhou have also been partially sponsored by the A. Russell Chandler III Professorship at Georgia Institute of Technology. 
Zhou is also supported by the Data Science Institute postdoctoral fellowship from Columbia University.

\bibliography{ref}

\newpage

\appendix
\setcounter{section}{0}
\renewcommand{\thesection}{\arabic{section}}
\section{Remarks}
\subsection{Remarks on Tsybakov noise condition}
As a recap, Tsybakov noise condition (Assumption 2.1) asserts that for some $c_0>0$ and $q\in [0,\infty)$, there holds $$\mathrm{P}\bigl(\{X\in \mathcal{X} : |\eta(X) -1/2| \leq t\}\bigr) \leq c_0t^q,\qquad \forall t>0.$$ 

Intuitively, this noise condition describes (with parameter $q \in [0, \infty)$) how the $\eta$ function behaves around the boundary $\{x: \eta(x) =1/2\}$. 
Bigger $q$ means there is a bigger jump of $\eta$ near the boundary $\{x: \eta(x) =1/2\}$, which is favorable for classification; smaller $q$ (i.e., $\eta$ close to $1/2$) means there is a plateau behavior near the boundary, which is considered difficult for classification. For the extreme case of hard margin when $q$ is
infinity, it implies that $\eta$ is bounded away from $1/2$.

\subsection{Holder Continuity of Density Function }
We denote by 
$C^m(\mathcal{X})$ with $m\in\mathbb{N}$, the space of $m$-times differentiable functions on $\mathcal{X}$.
For any positive value $\alpha >0$, let $[\alpha]^- = \lceil \alpha-1 \rceil \in \NN \cup \{0\}$.
Let $\bm \beta = (\beta_1, \ldots, \beta_d)\in \NN_0^d$ be an index vector, where $\NN_0 = \NN \cup \{0\}$. 
We define $|\bm \beta| =\beta_1+ \ldots+ \beta_d $ and $x^{\bm \beta} = x_1^{\beta_1}\cdots x_d^{\beta_d}$ for an index vector $\bm \beta$. For a function $f:\mathcal{X} \to \RR$ and a index vector $\bm \beta \in \NN_0^d$, let the partial derivative of $f$ with $\bm \beta$ be 
\begin{equation*}
    \partial^{\bm \beta} f = \frac{\partial^{|\bm \beta|}f}{\partial x^{\bm \beta}} = \frac{\partial^{|\bm \beta|}f}{\partial x_1^{\beta_1}\cdots \partial x_d^{\beta_d}}. 
\end{equation*}

We assume the density function
\begin{equation}\label{holder}
h \in \mathcal{H}^{\alpha,r}\left([0,1]^d\right) : = \{f\in C^{[\alpha]^-} \left([0,1]^d\right): \|f\|_{\mathcal{H^{\alpha}}\left([0,1]^d\right)}\leq r\},
\end{equation}
which is the closed ball of radius $r>0$ in the Hölder space of order $\alpha >0$ w.r.t. the  Hölder norm $\|\cdot\|_{\mathcal{H^{\alpha}}([0,1]^d)}$ given by
\begin{equation*}
    \|f\|_{\mathcal{H^{\alpha}}([0,1]^d)} = \sum_{|\bm \beta|_1 \leq [\alpha]^-} \left\{\|\partial^{\bm \beta} f\|_{C([0,1]^d)} + \sup_{x \neq y \in [0,1]^d} \frac{\left|\partial^{\bm \beta} f(x) - \partial^{\bm \beta} f(y)\right|}{|x-y|^{\alpha - [\alpha]^-}} \right\}. 
\end{equation*}

\section{Discussion of Novelty and Theoretical Significance}

In this part, we provide additional discussion to clarify the novelty and theoretical contribution of our work, particularly in relation to existing methods such as support vector machines (SVMs), and to explain how our analysis advances the understanding of unsupervised anomaly detection.

Generalization analysis of kernel methods in the literature is typically studied in terms of the sample error and approximation error bounds. 
For regression problems using the least-squares loss, the approximation error $$\epsilon(f_{\mathcal H}) - \epsilon(f_\rho) =\|f_{\mathcal H} -f_\rho\|^2_{L^2_Q}$$
is typically studied under the standard assumption that the regression function $f_\rho$ lies in the range of the $r$-th power of the integral operator $L^r_K$ for some $r>0$, where the integral operator $L_K$ is induced by a Mercer kernel $K$. 
This assumption, commonly referred to as the ``source condition", serves as a natural regularity condition: this is because when $r=1/2$, it requires $f_\rho$ lies in the RKHS ${\mathcal H}_K$ induced by $K$; when $0<r<1/2$, it implies that $f_\rho$ lies in an interpolation space between ${\mathcal H}_K$ and the $L^2$ space $L^2_Q$. 

Extending this type of analysis to classification problems with SVM-type kernel methods and the hinge loss 
$\phi$ is substantially more difficult. The main challenge lies in estimating the approximation error $$\epsilon(f_{\mathcal H}) - \epsilon(f_c),$$ 
where the Bayes classifier $f_c$ is discontinuous in general and lacks regularity, and therefore fails to satisfy the source condition.
Existing analyses of classification with the hinge loss rely on restrictive assumptions, such as a hard margin condition \citep{shawe2004kernel} or a condition that the (discontinuous) Bayes rule $f_c$ lies in an interpolation space between ${\mathcal H}_K$ and $L^1_Q$ \citep{chen2004support, cucker2007learning}. Both assumptions are strong and hard to verify in practice. 

To date, minimax-optimal classification rates in the literature  have been obtained by plug-in classifiers that replaces the hinge loss $\phi$ with the least squares loss, as in the work of \citet{audibert2007fast} and \citet{smale2007learning}. 
Additionally, the only known result achieving minimax rates with the hinge loss and deep ReLU networks is due to \citet{kim2021fast}, who impose a restriction $\|f\|_\infty \leq F_n$ on functions in the hypothesis space generated by the neural network. Such restriction is impractical for empirical risk minimization of neural network functions.

Our key novelty in this work lies in Lemma \ref{lemma:approx}, which establishes a bound on the approximation error $\epsilon(f_{\mathcal H}) - \epsilon(f_c)$ in terms of a threshold parameter $\tau$ and the approximation rate of the regression function $f_\rho$ by a ReLU neural network. 
By leveraging the smoothness of $f_\rho$, our analysis effectively circumvents the lack of regularity of the Bayes classifier $f_c$. This approach enables us to achieve minimax-optimal classification rates without requiring the restrictive boundedness condition $\|f\|_\infty \leq F_n$
 imposed in prior work.

\section{Additional Applications: Industrial and Medical Anomaly Detection}\label{section:add_exp}
Unsupervised anomaly detection has important applications beyond cybersecurity, for example in identifying product defects during industrial manufacturing processes and detecting diseases among patients. In manufacturing, products can exhibit a wide variety of defects, ranging from subtle surface scratches to structural deformities. It is difficult to anticipate in advance what kinds of anomalies might occur. As a result, unsupervised approaches are particularly valuable because they do not rely on prior knowledge of specific defect types. In the medical domain, by contrast, the challenge lies in the scarcity and sensitivity of data: collecting and sharing patient diagnostic records is expensive and often constrained by privacy regulations. Consequently, unsupervised anomaly detection methods offer a promising alternative in this setting as well, allowing meaningful insights to be extracted from limited medical data.

Although our proposed method was motivated to address the challenges encountered in network intrusion detection, its design principles are not domain-specific and can be naturally generalized across domains. In the following, we demonstrate this generalization by applying our method to unsupervised AD tasks on industrial image data and medical diagnostic records.

\subsection{Industrial Visual Inspection of Products}

\subsubsection{Dataset}
In industrial manufacturing processes, it is crucial to detect and filter out defective products before they reach customers. To evaluate our method in this context, we conduct experiments on the MVTec dataset \citep{bergmann2019mvtec}, a widely used real-world industrial image dataset designed specifically for unsupervised visual anomaly detection in manufacturing and quality inspection. The MVTec dataset comprises color images from $15$ object and texture categories (e.g., bottle, capsule, leather, metal nut, wood), where each category contains both defect-free (normal) images and exhibiting defects or anomalies (e.g., scratches, dents, or contaminations). In total, the dataset includes 73 different types of anomalies, which echos our earlier discussion that anomalies in manufacturing is highly diverse.

Since the dataset was constructed specifically for unsupervised AD, the original training set ($3629$ images) containing only normal samples. We perform a $0.9/0.1$ split on this training set to create separate training and validation subsets. The test set comprises $1725$ images, including both normal and defective samples. 

For our experiments, we train our theoretically motivated classifier (TC), along with other baseline anomaly detection methods, across all 15 categories, treating all anomaly types within each category as a single unified “anomalous” class. To see examples of images of this dataset, please refer to \citep{bergmann2019mvtec}.

\subsubsection{Implementation}
To apply our method to the MVTec dataset, we first convert the image data into a tabular form using feature embeddings extracted from a pretrained DINOv2-ViT model \citep{oquab2024dinov2}. In our setup, we use the DINOv2 (ViT-B/14) model with 1024-dimensional frozen embeddings, rather than fine-tuning the feature extractor to avoid overfitting. Each image is thus converted into a  1024-dimensional feature vector, serving as the input to our neural network.

Bringing insights from our earlier experiments, we employ a feed-forward neural network with a depth of $D=3$ and Leaky ReLU activation functions in place of ReLU to mitigate the dead neuron problem. Given the high dimensionality of feature data ($d=1024$), we choose the network width to $w = 6000$ to ensure sufficient representational capacity. 

We generate synthetic anomalies by sampling from a uniform distribution. We ensure the number of synthetic samples matches the total size of the training and validation sets ($n=n^\prime$). These synthetic anomalies are combined with the normal (defect-free) samples to train our neural network classifier. 
We compute the empirical risk minimizer, $\widehat{f}_{T, T^\prime,\phi}$, defined in (12) in the main paper using  Adam with learning rate $10^{-3}$ and mini-batch size $1024$. Additionally, we employ weight decay during training to induce parameter sparsity which is motivated from our theory. We train the neural network until convergence (which, in code, corresponds to 200 epochs
with early stopping on the validation loss). Model hyperparameters are tuned on the validation set to achieve optimal performance (please refer to Supplement \ref{appendix:model_choice} for details).

We compare our theory classifier (TC) with other classification-based
AD methods: SVM baseline and NS-NN. We also compare with other unsupervised AD baselines: OCSVM, Isolation Forest, and DeepSVDD. These are the same sets of AD baselines we test in our previous experiments. We report the AUPR (mean $\pm$ standard deviation) over five replicates in Table \ref{tab:mvtec_vertical}.

\subsubsection{Results and Discussions}

\begin{table}[t]
    \centering
    \caption{AUPR (mean $\pm$ standard deviation) for MVTec, split across object (vertical layout). Results reported in bold represent the best of the category. Each object contains normal and one anomaly class. }
    \begin{adjustbox}{max width=\textwidth}
    \begin{tabular}{lcccccccc}
        \toprule
        Object & Random & \multicolumn{3}{c}{Classify} & \multicolumn{3}{c}{Unsup.} \\
        \cmidrule(lr){3-5} \cmidrule(lr){6-8}
        &  & SVM & TC (Ours) & NS-NN & OCSVM & IsoF & DeepSVDD \\
        \midrule
        Bottle & 0.683 & \bf{0.991$\pm$0.000} & 0.982$\pm$0.008 & 0.683$\pm$0.000 & 0.961$\pm$0.015 & 0.977$\pm$0.008 & \bf{0.981$\pm$0.000} \\
        Cable & 0.577 & \bf{0.875$\pm$0.002} & 0.792$\pm$0.028 & 0.577$\pm$0.000 & 0.774$\pm$0.053 & 0.799$\pm$0.021 & \bf{0.806$\pm$0.001} \\
        Capsule & 0.789 & \bf{0.959$\pm$0.003} & 0.948$\pm$0.016 & 0.789$\pm$0.000 & 0.914$\pm$0.010 & \bf{0.943$\pm$0.010} & 0.932$\pm$0.000 \\
        Carpet & 0.714 & 0.989$\pm$0.001 & \bf{0.994$\pm$0.001} & 0.714$\pm$0.000 & 0.984$\pm$0.007 & 0.985$\pm$0.006 & \bf{0.991$\pm$0.000} \\
        Grid & 0.682 & \bf{1.000$\pm$0.000} & 0.948$\pm$0.060 & 0.682$\pm$0.000 & 0.945$\pm$0.018 & 0.971$\pm$0.013 & \bf{0.972$\pm$0.000} \\
        Hazelnut & 0.565 & \bf{0.915$\pm$0.003} & 0.761$\pm$0.024 & 0.565$\pm$0.000 & 0.749$\pm$0.050 & 0.806$\pm$0.043 & \bf{0.839$\pm$0.000} \\
        Leather & 0.695 & \bf{1.000$\pm$0.000} & 0.999$\pm$0.000 & 0.756$\pm$0.000 & \bf{0.999$\pm$0.000} & \bf{0.999$\pm$0.001} & \bf{0.999$\pm$0.000} \\
        Metal Nut & 0.756 & \bf{0.961$\pm$0.001} & 0.906$\pm$0.013 & 0.817$\pm$0.000 & 0.934$\pm$0.010 & 0.940$\pm$0.004 & \bf{0.951$\pm$0.000} \\
        Pill & 0.817 & \bf{0.959$\pm$0.001} & 0.957$\pm$0.007 & 0.699$\pm$0.000 & 0.931$\pm$0.013 & 0.942$\pm$0.004 & \bf{0.944$\pm$0.000} \\
        Screw & 0.699 & \bf{0.893$\pm$0.008} & 0.716$\pm$0.070 & 0.670$\pm$0.000 & \bf{0.723$\pm$0.052} & 0.715$\pm$0.017 & 0.715$\pm$0.001 \\
        Tile & 0.670 & \bf{0.996$\pm$0.000} & 0.993$\pm$0.001 & 0.333$\pm$0.000 & 0.987$\pm$0.007 & 0.992$\pm$0.003 & \bf{0.994$\pm$0.000} \\
        Transistor & 0.333 & \bf{0.875$\pm$0.004} & 0.838$\pm$0.050 & 0.756$\pm$0.122 & 0.765$\pm$0.063 & 0.804$\pm$0.024 & \bf{0.840$\pm$0.002} \\
        Wood & 0.732 & \bf{0.987$\pm$0.001} & 0.963$\pm$0.014 & 0.732$\pm$0.000 & 0.971$\pm$0.001 & \bf{0.978$\pm$0.007} & 0.977$\pm$0.000 \\
        Zipper & 0.758 & \bf{0.998$\pm$0.000} & 0.985$\pm$0.007 & 0.758$\pm$0.000 & 0.991$\pm$0.005 & \bf{0.993$\pm$0.002} & 0.989$\pm$0.000 \\
        \bottomrule
    \end{tabular}
    \end{adjustbox}
    \label{tab:mvtec_vertical}
\end{table}

Table \ref{tab:mvtec_vertical} reports the AUPR  for normal versus anomalous samples across 15 object and texture categories. Our theory classifier (TC) performs consistently and competitively across all categories, often reaching performance close to the best result achieved by SVM, and clearly surpassing other unsupervised methods such as OCSVM, Isolation Forest, and DeepSVDD. TC attains AUPR values above 0.94 for most objects (e.g., Carpet, Metal Nut, Pill, Wood, and Zipper), demonstrating its robustness despite operating in a fully unsupervised setting. Importantly, TC maintains stable results across diverse object types, confirming that the theoretical principles underlying our model translate effectively to practical scenarios. 

Overall, our theoretically motivated classifier achieved competitive performance on the MVTec dataset compared to other established anomaly detection methods, confirming the practical effectiveness and robustness of our approach beyond the cybersecurity applications. 

\subsection{Detecting Abnormal Thyroid Conditions}
Having demonstrated the effectiveness of our method on industrial image data, we next evaluate it to a distinct domain --- medical anomaly detection. Unlike manufacturing images, medical datasets consist of structured diagnostic records that are scarce, sensitive, and costly to collect. Despite these differences, both domains share the challenge of identifying rare anomalies without explicit supervision.
\subsubsection{Dataset}
For the medical domain, we conduct experiments on the Thyroid Disease dataset from the UCI Machine Learning Repository \citep{thyroid_dataset}, a widely used benchmark for medical anomaly detection. This dataset consists of patient diagnostic records that include individuals with normal thyroid function (labeled as ``normal"),  as well as patients with over-active (``hyperfunction") and under-active (``subnormal") thyroid conditions. The latter two classes are considered anomalies. The dataset has a moderate dimensionality of $21$ features, capturing a range of physiological and clinical indicators relevant to thyroid health. Consistent with the unsupervised setting, the training set comprises $3425$ normal samples and no real anomalies. We perform a $0.9/0.1$ split to create separate training and validation subsets. The test set includes $3420$ samples containing both normal and abnormal cases.

\subsubsection{Implementation}
The implementation of our method on the Thyroid Disease dataset largely follows the same procedure described in the previous experiments. The main difference lies in adapting the network architecture to the relatively low-dimensional input space ($d = 21$): we employ a feed-forward neural network with a depth of $D = 3$ and width of $w = 200$. As before, we use Leaky ReLU activations, weight decay regularization to induce sparsity, and the Adam optimizer to minimize the empirical risk defined by our theoretical objective.
\subsubsection{Results and Discussions}
\begin{table}[h]
    \centering
    \caption{AUPR for Thyroid, split across abnormal diagnosis in each column.  Results reported in bold represent the best of the category. Suffix for NS-NN refers to the number of layers.}     
    \begin{tabular}{cl|cc}
        \toprule
        &Model$\backslash$Anom. & Hyperfunction & Subnormal \\
         \midrule
        &Random &  0.023 & 0.053\\
        \midrule
        \parbox[t]{2mm}{\multirow{5}{*}{\rotatebox[origin=c]{90}{Classify}}} & SVM & 0.241$\pm$0.023 & 0.047$\pm$0.002 \\ 
        &TC (Ours) & 0.287$\pm$0.032 & 0.056$\pm$0.005 \\
        &NS-NN2 & 0.333$\pm$0.050 & 0.061$\pm$0.006 \\
        &NS-NN3 & \bf{0.352$\pm$0.020} & \bf{0.064$\pm$0.005} \\
        \midrule
        \parbox[t]{2mm}{\multirow{3}{*}{\rotatebox[origin=c]{90}{Unsup.}}}& OCSVM & 0.144$\pm$0.015 & 0.043$\pm$0.006 \\
        &IsoF & 0.177$\pm$0.033 & \bf{0.069$\pm$0.008} \\ 
        &DeepSVDD & 
    \bf{0.335$\pm$0.031} & 0.067$\pm$0.017 \\ 
         \bottomrule
    \end{tabular}
    \label{tab:thyroid_full}
\end{table}

Table \ref{tab:thyroid_full} reports the AUPR (mean $\pm$ standard deviation) over five replicates on the Thyroid Disease dataset. The results show that our proposed theory classifier (TC) performs competitively against other unsupervised baselines, though its relative strength varies across diagnostic categories. In the hyperfunction class, TC achieves an AUPR of $0.287 \pm 0.032$, which surpasses  SVM baseline but falls slightly short of the highest-performing unsupervised methods such as NS-NN3 ($0.352 \pm 0.020$). For the subnormal class, TC attains $0.056 \pm 0.005$, again outperforming SVM and closely approaching the top unsupervised baselines. While TC does not achieve the absolute best performance in every case, it remains consistently effective across both anomaly types, showing lower variance than most baselines.

\section{Experimental Details}
\label{appendix:exp_details}

We proceed to share more details of our experiments for reproducibility and to explain our design choices.

\subsection{Dataset Selection for Cybersecurity Applications}
\label{appendix:data_details}
Here, we explain our choice of datasets.

\paragraph*{NSL-KDD}
NSL-KDD dataset is an updated version of the KDDCup99 dataset \citep{kddcup99_dataset} that is used as a common anomaly detection benchmark (e.g., \citet{deep_ad_SSL_adv_training, AD_tabular_data_internal_contrastive_learning, NeuTraL_AD, lipschitz_ae_AD}).
It contains four types of attacks as anomalies: Denial of Service (DoS), probe, remote access (RA, also known as remote-to-local (R2L)) and privilege escalation (PE, also known as user-to-root (U2R)) attacks. 
While this well-established benchmark dataset serves its purpose well for the evaluation, we note that other newer datasets may reflect the current landscape of network attacks better.
Hence, we also experiment with the Kitsune dataset, which has more up-to-date data.

\paragraph*{Kitsune}
Similar to NSL-KDD, Kitsune is a network traffic dataset but was created in 2018, by a top security conference work \citep{kitsune_dataset}. 
This dataset contains two networks: one with 8 types of network attacks and the other with just 1 attack.
For a fair comparison across attacks while maintaining diversity, we select the first network with 8 attacks.
To be precise, there are 8 datasets, each with normal data and one type of attack.
We follow their set-up to use the first 1 million samples as training and the rest for evaluation.
During evaluation, we noticed that all models tested consistently obtain random performance for the video injection attack, similar to the sub-par results obtained in the original paper, so we do not include the results from this attack in the paper for brevity.
The other seven attacks we use as anomalies can be split into 3 DoS-like attacks (Simple Service Discovery Protocol (SSDP) flood, Secure Socket Layer (SSL) renegotiation and SYN DoS), 2 reconnaissance attacks (Operating System (OS) scan and fuzzing) and 2 man-in-the-middle (MitM) attacks (active wiretap and Address Resolution Protocol (ARP) MitM).
We provide a summary of the attacks from the two datasets in Table \ref{tab:attack_summary}.

\begin{table}[]
    \centering
    \caption{Summary of attacks and examples of potentially indicative features of the attacks.}
  \begin{adjustbox}{width=\textwidth}
    \begin{tabular}{c| >{\centering\arraybackslash}m{0.07\linewidth}|m{0.4\linewidth}|m{0.35\linewidth}}
    \toprule
        Dataset & Attack & Description & Potential Features for Detection \\
        \midrule
        \multirow{4}{*}[-45pt]{NSL-KDD} & DoS & Generate a large number of request/response packets to exhaust victims' resources. & Basic TCP connection and traffic features. For example, count, protocol\_type, serror\_rate, rerror\_rate, etc.\\
        \cline{2-4}
         & Probe &Send probing packets to gather information about the target system based on the responses. &Basic TCP connection and traffic features. For example, count, service, srv\_count, diff\_srv\_rate, dst\_host\_same\_srv\_rate, etc. \\
        \cline{2-4}
         & RA &Send packets with vulnerability-exploiting payloads resulting in gaining local access on the host. &Content features. For example, is\_guest\_login, is\_hot\_login, logged\_in, num\_failed\_logins, etc.\\
        \cline{2-4}
         & PE & Send packets with vulnerability-exploiting payloads resulting in privilege escalation on the host. &Content features. For example, num\_root, root\_shell, su\_attempted, etc.\\
         \midrule
        \multirow{7}{*}[-35pt]{Kitsune} & SSDP Flood & Send a large number of spoofed Simple Service Discovery Protocol (SSDP) packets to Universal Plug and Play (UPnP) devices which generate an overwhelming amount of response packets to the victim. &Bandwidth features. For example, statistics of packet count, size, and jitter. \\
        \cline{2-4}
         & SSL Reneg. &Send a large number of SSL renegotiation packets to exhaust servers' resources. &Bandwidth features. For example, statistics of packet count, size, and jitter. \\
        \cline{2-4}
         & SYN DoS &Send a large number of SYN packets to exhaust servers' resources. &Bandwidth features. For example, statistics of packet count, size, and jitter. \\
        \cline{2-4}
         & Wiretap &Intercept loca area network (LAN) traffic via active wiretap (network bridge). & Bandwidth features (indirect). For example, statistics of packet jitter.\\
        \cline{2-4}
         & ARP MitM &Intercept LAN traffic via an ARP poisoning attack. & Bandwidth features (indirect). For example, statistics of packet jitter.\\
        \cline{2-4}
         & Fuzzing &Send random commands to web servers' Common Gateway Interface (CGI) to find vulnerabilities. & Content features (not explicitly included in the dataset). \\
        \cline{2-4}
         & OS Scan &Send selected probe packets to hosts and infer the OS based on responses' fingerprints. &Content features (not explicitly included in the dataset).  \\
         \bottomrule
    \end{tabular}
    \end{adjustbox}
    \label{tab:attack_summary}
\end{table}

\subsection{Model Choice}
\label{appendix:model_choice}
Model details can be found in our code implementation.
In this section, we outline how we chose certain hyperparameters of certain models.

For classification-based models, as mentioned in the text, it was difficult to systematically validate our choices (e.g., with cross validation) because validation performance was perfect.
We use the idea behind structural risk minimization (SRM) to decrease model complexity while validation loss converges to zero. 
For SVM, we used SVM equipped with the classical RBF kernel and adopt $\ell_2$-regularization during training. We adopt SRM on the validation dataset for selecting
these kernel parameter and the regularization parameter, which are both non-negative.

To choose theory classifier hyperparameters, we also ensure that the model is complex enough by requiring that it converges quickly.
Since we repeat experiments five times, we find the optimal model size (depth and width) that converges within a couple (1-7) of epochs for the 5 runs, rather than just 1 epoch for all 5 runs.\footnote{If it is too time-consuming, some discretion can be taken to relax this requirement. We observe that results generally do not vary too much if this condition is relaxed slightly.}
We do this by increasing the size of the neural network, and if the number of epochs to converge is within a couple of epochs, we use that neural network; otherwise, we continue increasing the size of the neural network.
We chose to start the search with a small width, but not smaller than the data dimensional, to prevent anomalies from projecting onto normal points/regions, as observed in \citet{lau2024geometric_OSR}.
For the NS-NN, the hyperparameters we could choose were the network's depth, width, and dropout probabilities.
We use their proposed configurations as hyperparameters to try.
Since they tried three configurations of depth (2/3/4 layers), we increased the width ratio (9.1/27.6/46.1 neurons per dimension per hidden layer)\footnote{We obtained the width ratio by taking the width used for the original dataset divided by the dataset dimensionality. The width we use is the width ratio multiplied by our dataset dimensionality. Using the width ratio allows us to calibrate the relative width, especially because the data \citet{sipple20NSNN} used had much lower ($<40$) dimensionality than our datasets.} and dropout (0.34/0.57/0.8) with increasing depth.
We report all 3 networks in this Supplementary Material, but we chose the network with 2 layers to report in the main text for the following reason:
the deepest had the worst test results and the other 2 had comparable results, so we adopt the SRM approach here too. 

For the unsupervised methods, we trained only with normal data.
For OCSVM, we set the contamination parameter to be $\nu\approx 0$.
For Isolation Forest, we tried 50/100/150 estimators and observed that the validation performance of normal versus synthetic anomalies were all perfect, so we chose to implement 50 estimators as per SRM.
For DeepSVDD, we use the autoencoder architecture with the lowest validation loss during pre-training.
We tried with 3 sets of encoder architectures (with the decoder mimicking the reverse): $d$-90-45-20, $d$-90-45, $d$-60-20 ($d$ is 119 for NSL-KDD, 115 for Kitsune, 1024 for MVTec, and 21 for Thyroid).

\subsection{Full Results}

We report full results for NSL-KDD in Table \ref{tab:kdd_full} (less the ablation results in Section 5.4 in the main paper) and Kitsune in Table \ref{tab:kitsune_full}.
We note that the full results for MVTec and Thyroids have already been reported in Supplement \ref{section:add_exp}.
Since the results for NSL-KDD were elaborated on in the ablations in Section 5.4 in the main paper,
we focus on Kitsune (Table \ref{tab:kitsune_full}).
For our theory classifier, we used $D=4$ layers and varied the width $w$ to control the complexity of the model (as detailed in the previous section), but report results for all models as an ablation on the width $w$.
We underline the results that were ultimately used for our theory classifier and report results in bold for the best theory classifier.
In Table \ref{tab:kitsune_full}, among all TC with varying width, we see that wider models generally work better for easier attacks (SSDP and SSL, less SYN) and narrower models generally work better for harder attacks (wiretap, ARP MitM and OS scan, less fuzzing).
More exploration can be done here on other datasets too.

\begin{table}[]
    \centering
    \caption{AUPR for KDD, split across attacks in each column. 
    Higher is better.
    Suffix for NS-NN refers to the number of layers.
    }
    \begin{tabular}{cl|cccc}
        \toprule
        &Model$\backslash$Attack & DoS & Probe & PE & RA \\
         \midrule
        &Random & 0.431 & 0.197 & 0.007 & 0.218 \\
        \midrule
        \parbox[t]{2mm}{\multirow{5}{*}{\rotatebox[origin=c]{90}{Classify}}} & SVM & 0.890$\pm$0.002 & 0.836$\pm$0.001 & 0.013$\pm$0.000 & 0.319$\pm$0.001 \\
        &TC (Ours) & \bf{0.907$\pm$0.009} & \bf{0.937$\pm$0.009} & \bf{0.028$\pm$0.007} & \bf{0.389$\pm$0.030} \\
        &NS-NN2 & 0.753$\pm$0.015 & 0.286$\pm$0.067 & 0.003$\pm$0.000 & 0.109$\pm$0.000 \\
        &NS-NN3 & 0.431$\pm$0.000 & 0.197$\pm$0.000 & 0.009$\pm$0.000 & 0.217$\pm$0.000 \\
        &NS-NN4 & 0.431$\pm$0.000 & 0.197$\pm$0.000 & 0.009$\pm$0.000 & 0.217$\pm$0.000 \\
        \midrule
        \parbox[t]{2mm}{\multirow{3}{*}{\rotatebox[origin=c]{90}{Unsup.}}}&OCSVM & 0.802$\pm$0.091 & 0.687$\pm$0.141 & 0.025$\pm$0.022 & 0.262$\pm$0.095 \\
        &IsoF & \textbf{0.942$\pm$0.006} & \textbf{0.923$\pm$0.049} & \textbf{0.045$\pm$0.014} & 0.398$\pm$0.028 \\
        &DeepSVDD & 0.860$\pm$0.003 & 0.848$\pm$0.004 & 0.019$\pm$0.000 & \bf{0.515$\pm$0.006} \\
         \bottomrule
    \end{tabular}
    \label{tab:kdd_full}
\end{table}

\begin{table}[h]
    \centering
    \caption{AUPR for Kitsune, split across attack types. Width $w$ in parentheses for our theory classifier (TC). Underlined results are the results from the models used.
    Results reported in bold represent the best of the category.
    }
  \begin{adjustbox}{width=\textwidth}
    \begin{tabular}{cl|ccccccc}
    \toprule
        &Model & SSDP & SSL & SYN & Wiretap & ARP MitM & Fuzzing & OS Scan \\
         \midrule
        &Random & 0.468 & 0.077 & 0.004 & 0.217 & 0.761 & 0.348 & 0.094 \\
        \midrule
        \parbox[t]{2mm}{\multirow{7}{*}{\rotatebox[origin=c]{90}{Classify}}} & SVM & \textbf{1.000$\pm$0.000} & \textbf{0.729$\pm$0.001} & 0.146$\pm$0.007 & 0.654$\pm$0.054 & 0.682$\pm$0.014 & 0.345$\pm$0.164 & \textbf{0.744$\pm$0.001} \\
        \cmidrule{2-9}
        &TC (115) & 0.916$\pm$0.188 & 0.560$\pm$0.064 & 0.120$\pm$0.112 & \underline{\bf{0.756$\pm$0.095}} & \bf{0.872$\pm$0.100} & 0.574$\pm$0.093 & 0.449$\pm$0.120 \\
        &TC (335) & \underline{0.997$\pm$0.005} & \underline{0.565$\pm$0.128} & \underline{0.198$\pm$0.093} & 0.710$\pm$0.117 & \underline{0.821$\pm$0.083} & \underline{0.582$\pm$0.106} & \underline{0.497$\pm$0.007} \\
        &TC (500) & 0.999$\pm$0.002 & 0.616$\pm$0.130 & 0.184$\pm$0.107 & 0.648$\pm$0.052 & 0.828$\pm$0.104 & \bf{0.631$\pm$0.130} & 0.437$\pm$0.138 \\
        &TC (678) & 0.999$\pm$0.001 & 0.689$\pm$0.140 & 0.164$\pm$0.139 & 0.641$\pm$0.056 & 0.839$\pm$0.072 & 0.610$\pm$0.119 & 0.499$\pm$0.006 \\
        \cmidrule{2-9}
        &NS-NN2 & 0.999$\pm$0.000 & 0.649$\pm$0.002 & \bf{0.242$\pm$0.000} & 0.723$\pm$0.000 & 0.801$\pm$0.010 & 0.360$\pm$0.000 & 0.500$\pm$0.001 \\
        &NS-NN3 & 0.993$\pm$0.003 & 0.563$\pm$0.009 & 0.221$\pm$0.028 & 0.722$\pm$0.000 & 0.761$\pm$0.000 & 0.360$\pm$0.000 & 0.500$\pm$0.000 \\
        &NS-NN4 & 0.994$\pm$0.002 & 0.548$\pm$0.009 & 0.111$\pm$0.078 & 0.722$\pm$0.000 & 0.761$\pm$0.000 & 0.353$\pm$0.005 & 0.501$\pm$0.000 \\
        \midrule
        \parbox[t]{2mm}{\multirow{3}{*}{\rotatebox[origin=c]{90}{Unsup.}}} & OCSVM & 0.989$\pm$0.014 & 0.517$\pm$0.035 & 0.228$\pm$0.015 & \bf{0.695$\pm$0.011} & \textbf{0.653$\pm$0.009} & \textbf{0.252$\pm$0.008} & 0.502$\pm$0.000 \\
        &IsoF & 0.993$\pm$0.008 & 0.498$\pm$0.021 & 0.212$\pm$0.047 & 0.691$\pm$0.013& 0.651$\pm$0.012 & 0.254$\pm$0.010 & \textbf{0.502$\pm$0.001} \\
        &DeepSVDD & \bf{0.998$\pm$0.000} & \textbf{0.642$\pm$0.000} & \bf{0.291$\pm$0.003} & {0.623$\pm$0.001} & 0.690$\pm$0.003 & 0.246$\pm$0.001 & {0.500$\pm$0.000} \\
         \bottomrule
        \end{tabular}
  \end{adjustbox}
    \label{tab:kitsune_full}
\end{table}

\section{Auxiliary Lemma}\label{append:lemma}
The result from \citep{schmidt2020nonparametric} demonstrates that ReLU networks can accurately approximate any H\"older continuous function.
The result is stated below.
We use this result to derive an approximation error bound (see Supplement \ref{subsec:approx}) and to define our target neural network function class (i.e., hypothesis space) $\mathcal{H}_\tau$ in Definition 3.1 in the main paper. 
\begin{lemma}[Theorem 5 in \citep{schmidt2020nonparametric}]\label{schmidt}
    Let $\alpha, r>0$. For any Hölder continuous function $\eta \in \mathcal{H}^{\alpha,r}([0,1]^d)$ and for any integers $m\geq 1$ and \\$N \geq \max \left\{(\alpha +1)^d, (r+1)e^d\right\}$, 
    there exists a ReLU neural network 
    \begin{equation*}
        \widehat{\eta} \in \mathcal{F}(L^*, w^*, v^*,K^*)
    \end{equation*} with depth $$L^*= 8+ (m+5)(1+\lceil \log_2(\max\{d,\alpha\}) \rceil),$$  
    maximum number of nodes $$w^* = 6(d+ \lceil \alpha \rceil )N,$$ 
    number of nonzero parameters $$v^*= 141 (d+\alpha +1)^{3+d}N (m+6) ,$$ and all parameters (absolute value) are bounded by $$K^*= 1$$
    such that
    \begin{equation}\label{schmidt_error}
        \left\|\widehat{\eta} -\eta\right\|_{L^\infty([0,1]^d)}\leq (2r+1)(1+d^2 + \alpha^2)6^d N 2^{-m} + r3^\alpha N^{-\frac{\alpha}{d}}. 
    \end{equation}
\end{lemma}

\section{Proof of Theorem 4.1}
\label{appendix:proof_thm_1_excess_risk}
This part presents the proof of Theorem 4.1. Specifically, we derive the convergence rate in terms of $n^* = \min\{n, n^\prime\}$ of the excess risk.

Recall that we assumed the marginal distribution on $\mathcal{X}$ is given by $P_\mathcal{X} = sQ + (1-s)\mu$ for some $s \in (0,1)$. We also assumed $dQ = hd\mu$, where $h$ is an unknown, $\alpha$-Hölder continuous density function.
Thus, we know  \begin{equation}\label{dP}
    d P_\mathcal{X} =sdQ + (1-s)d\mu = ( sh + (1-s))d\mu.
\end{equation} 
Recall also the conditional class probability function introduced in Section 2 of the main paper:
\begin{equation*}
    \eta(X)=P(Y=1|X)=\frac{s\cdot h(X)}{s\cdot h(X)+1-s}, \qquad \forall X\in \mathcal{X}.
\end{equation*}

For any function  $f: \mathcal{X} \to \RR$, its generalization error associated with the Hinge loss $\phi$ is defined by
\begin{align*}
    \varepsilon(f) &= \int_{\mathcal{X}} \int_{\mathcal{Y}} \phi(Yf(X)) dP(Y|X) dP_\mathcal{X}   \\ 
    &= \int_{\mathcal{X} } \left[\phi(f(X))\mathrm P(Y=1|X) +  \phi(-f(X))\mathrm P(Y=-1|X) \right]dP_\mathcal{X} \\
    &= \int_{\mathcal{X} } \left[\phi(f(X))\eta(X) +  \phi(-f(X))\mathrm (1-\eta(X))\right]dP_\mathcal{X}\\
    &\stackrel{\eqref{dP}}{=}\int_{\mathcal{X} }  \left[\phi(f(X))sh(X) +  \phi(-f(X))(1-s)\right]d\mu \\
     &\stackrel{\because dQ = hd\mu}{=}s \int_{\mathcal{X} } \phi(f(X))dQ + (1-s)\int_{\mathcal{X} } \phi(-f(X))d\mu.
\end{align*}
With the  training data set $T \cup T^\prime = \{(X_i,1)\}_{i=1}^{n} \cup \{(X_i^\prime,-1)\}_{i=1}^{n^\prime}$, the empirical risk w.r.t. $\phi$:
\begin{equation*}
    \varepsilon_{T,T^\prime} (f):=\frac{s}{n} \sum_{i=1}^{n} \phi\left(1 \cdot f(X_i)\right) + \frac{(1-s) }{n^\prime}\sum_{i=1}^{n^\prime}\phi(-1 \cdot f(X^\prime_i)) 
\end{equation*}
can be seen as the empirical counterpart of $\varepsilon(f)$. 

Recall that we use $f_c$ to denote the Bayes classifier.
The well-known Comparison Theorem in classification \citep{zhang2004statistical} states that, for the Hinge loss $\phi$ and any measurable function $f:\mathcal{X}\rightarrow \RR$, there holds 
\begin{equation}
    \underbrace{R(f)-R(f_c)}_{\text{excess risk}} \leq \underbrace{\varepsilon(f) - \varepsilon(f_c)}_{\text{excess generalization error}}.   
\end{equation}
In other words, we can minimize the excess generalization error (also known as the excess $\phi$-error) $\varepsilon\left(\Hat{f}_{T, T^\prime,\phi}\right) - \varepsilon(f_c)$ to, in turn, bound the excess risk  $R\left(\text{sign}\left(\Hat{f}_{T, T^\prime,\phi}\right)\right)-R(f_c)$.

We begin our analysis with a standard error decomposition (Supplement \ref{subsec:decomposition}) to decompose the excess generalization error $\varepsilon(\Hat{f}_{T, T^\prime,\phi}) - \varepsilon(f_c)$ into two estimation error terms and one approximation error term. After that, we estimate each of these error terms respectively (Supplement \ref{subsec:approx} and \ref{subsec:estimation}). Finally, we combine the error estimates and find the ideal parameters (Supplement \ref{subsec:combine}).

\subsection{Error Decomposition}\label{subsec:decomposition}
We consider the following error decomposition. Similar error decomposition can be found in \citep{zhou2023classification}.
\begin{lemma}[Decomposition of $\varepsilon(\Hat{f}_{T, T^\prime,\phi}) - \varepsilon(f_c)$]
Let $f_\mathcal{H}$ be any function in $\mathcal{H}_\tau$. There holds 
\begin{equation} \label{decomposition}
   \varepsilon(\Hat{f}_{T, T^\prime,\phi}) - \varepsilon(f_c) \leq \{\varepsilon(\Hat{f}_{T, T^\prime,\phi})- \varepsilon_{T,T^\prime}(\Hat{f}_{T, T^\prime,\phi})\}  +  \{\varepsilon_{T,T^\prime}(f_\mathcal{H}) - \varepsilon(f_\mathcal{H})\} + \{\varepsilon(f_\mathcal{H}) - \varepsilon(f_c)\}.
\end{equation}
\end{lemma}
\begin{proof} We express $\varepsilon(\Hat{f}_{T, T^\prime,\phi}) - \varepsilon(f_c)$ by inserting empirical risks as follows
\begin{eqnarray*}
    \varepsilon(\Hat{f}_{T, T^\prime,\phi}) - \varepsilon(f_c)
    &=& \{\varepsilon(\Hat{f}_{T, T^\prime,\phi})- \varepsilon_{T,T^\prime}(\Hat{f}_{T, T^\prime,\phi})\}
    + \{\varepsilon_{T,T^\prime}(\Hat{f}_{T, T^\prime,\phi}) - \varepsilon_{T,T^\prime}(f_\mathcal{H})\} \\
    &\quad +&  \{\varepsilon_{T,T^\prime}(f_\mathcal{H}) - \varepsilon(f_\mathcal{H})\} + \{\varepsilon(f_\mathcal{H}) - \varepsilon(f_c)\}.
\end{eqnarray*}
Note that both $\Hat{f}_{T, T^\prime,\phi}$ and $f_\mathcal{H}$ lie on the hypothesis space $\mathcal{H}_\tau$. From the definition of $\Hat{f}_{T, T^\prime,\phi}$ at (12) in the main paper,
$\Hat{f}_{T, T^\prime,\phi}$ minimizes the empirical risk over $\mathcal{H}_\tau$. Thus we have $\varepsilon_{T,T^\prime}(\Hat{f}_{T, T^\prime,\phi}) - \varepsilon_{T,T^\prime}(f_\mathcal{H}) \leq 0$ for all $ f_\mathcal{H} \in \mathcal{H}_\tau$. This yields the expression (\ref{decomposition}).
\end{proof}

The expression $\{\varepsilon(\Hat{f}_{T, T^\prime,\phi})- \varepsilon_{T,T^\prime}(\Hat{f}_{T, T^\prime,\phi})\}$ is the first estimation error (also known as the sample error) term, $\{\varepsilon_{T,T^\prime}(f_\mathcal{H}) - \varepsilon(f_\mathcal{H})\}$ is the second estimation error term, whereas $\{\varepsilon(f_\mathcal{H}) - \varepsilon(f_c)\}$ --- which does not depend on the training samples --- is the approximation error term induced by $f_\mathcal{H}$. To give an upper bound to the excess generalization error, we will proceed to bound these three error terms respectively. 

We will begin with estimating the upper bound of the approximation error term $\{\varepsilon(f_\mathcal{H}) - \varepsilon(f_c)\}$.

\subsection{Upper Bound of the Approximation Error}\label{subsec:approx}
Note that $\sigma_\tau(\widehat{\eta}-1/2) \in \mathcal{H}_\tau$, where $\widehat{\eta}$ is the approximation of the function $\eta$ in Lemma \ref{schmidt}. 
Recall that we use $f_\mathcal{H}$ to denote any functions in $\mathcal{H}_\tau$. In this subsection, we derive a tight upper bound for the approximation error $\varepsilon(f_\mathcal{H})- \varepsilon(f_c)$ by taking $f_\mathcal{H} = \sigma_\tau(\widehat{\eta}-1/2)$. We recall that $f_c$ is the Bayes classifier given by $f_c = \text{sign}\left(\eta - 1/2\right)$. 

\begin{lemma}\label{lemma:approx}
    Let $0<\tau\leq 1$. 
    Assume the Tsybakov noise condition (Assumption 2.1) holds for some noise 
    exponent $q\in [0,\infty)$ and constant $c_0>0$. There holds
    \begin{equation}\label{approx_eqn}    \varepsilon\left(\sigma_\tau\left(\widehat{\eta} -1/2\right)\right)- \varepsilon(f_c) \leq 
    8c_0\left(\tau +\|\widehat{\eta}-\eta\|_{L^\infty[0,1]^d}\right)^{q+1}.
    \end{equation}
\end{lemma}

\begin{proof}[Proof of Lemma \ref{lemma:approx}]
    Observe that the Hinge loss function $\phi(x) = \max \{0, 1-x\}$ is Lipschitz continuous on $\RR$ because 
\begin{equation} \label{Lip}
   |\phi(x_1)-\phi(x_2)| \leq |x_1-x_2| \qquad \forall x_1, x_2 \in \RR. 
\end{equation}

For any $f$ such that $|f(X)|\leq 1$, we have $\phi(Yf(X)) = 1-Yf(X)$.
Then,
\begin{align*}
  \varepsilon(f_\mathcal{H})- \varepsilon(f_c)  
  &= \int_\mathcal{X} \int_\mathcal{Y}   1-Yf_\mathcal{H}(X) - (1 - Yf_c(X)) \; dP(Y|X) d P_\mathcal{X}\\ 
  &= \int_\mathcal{X} \int_\mathcal{Y}  -Y(f_\mathcal{H}(X) - f_c(X)) \; dP(Y|X) d P_\mathcal{X}\\ 
  &=  \int_\mathcal{X}(f_\mathcal{H}(X) - f_c(X)) \int_\mathcal{Y}  -Y \; dP(Y|X)  d P_\mathcal{X}\\
  &=\int_\mathcal{X}(f_c(X)-f_\mathcal{H}(X) ) (2\eta(X)-1)\; d P_\mathcal{X}\\
  &= \int_\mathcal{X}  \left| f_\mathcal{H}(X) - f_c(X)\right|  |2\eta(X)-1|\;d P_\mathcal{X}.
\end{align*}
In the last equality, we have used the fact that $2\eta(X)-1>0$ if and only if $f_c(X)=1$. 

Take $f_\mathcal{H} = \sigma_\tau(\widehat{\eta}-1/2)$. We consider the cases $|\eta(X)-1/2| \leq \epsilon$ and $|\eta(X)-1/2| > \epsilon$ separately for $\epsilon :=\|\widehat{\eta}-\eta\|_{L^\infty[0,1]^d}$.

Notice that \begin{equation}\label{fH-fc}
    \left| f_\mathcal{H}(X) - f_c(X)\right| \leq |f_\mathcal{H}(X)|+ |f_c(X)| \leq 2 \qquad \forall X\in \mathcal{X}. 
\end{equation}
For the case $|\eta(X)-1/2| \leq \epsilon$, it follows from the Tsybakov noise condition (Assumption 2.1) that
\begin{eqnarray*}
 && \int_{\{X\in  \mathcal{X}:|\eta(X)-1/2| \leq \epsilon\} } \left| f_\mathcal{H}(X) - f_c(X)\right| |2\eta(X)-1| d P_\mathcal{X} \\
 &\stackrel{\eqref{fH-fc}}{\leq}& 4\epsilon \int_{\{x\in  \mathcal{X}: |\eta(X)-1/2| \leq \epsilon\} } \; d P_\mathcal{X}  \\
  &=&4\epsilon \cdot \mathrm{P}_{\mathcal{X}}(\{x\in  \mathcal{X} :|\eta(X)-1/2| \leq \epsilon\})\\
 &\stackrel{\text{Assumption 2.1}}{\leq}& 4\epsilon \cdot c_0\epsilon^q =  4c_0\epsilon^{q+1}.   
\end{eqnarray*}
 
 Next, for the case $|\eta(X)-1/2| > \epsilon$,  since  $|\widehat{\eta}(X)-\eta(X)|\leq \epsilon$, we have $\text{sign}(\widehat{\eta}(X) -1/2) = \text{sign} (\eta(X) -1/2)= f_c(X)$. 
 If $\sigma_\tau\left(\widehat{\eta}(X) -1/2\right)= f_\mathcal{H}(X)  \in \{1, -1\}$, then $f_\mathcal{H}(X) $ is exactly equal to $f_c(X)$, which implies \begin{equation}\label{equal0}
\left|\sigma_\tau\left(\widehat{\eta}(X)-1/2\right) - f_c(X)\right|=0.
 \end{equation}
 Thus, we have
 \begin{eqnarray*}
  &&\int_{\{X\in \mathcal{X}:  |\eta(X)-1/2| > \epsilon\} }  \left| f_\mathcal{H}(X) - f_c(X)\right|  |2\eta(X)-1| d P_\mathcal{X}  \\
  &\stackrel{\eqref{equal0}}{=}&\int_{\{X\in \mathcal{X}:   |\eta(X)-1/2| > \epsilon, |\sigma_\tau\left(\widehat{\eta}(X) -1/2\right)|< 1\}}  \left| \sigma_\tau\left(\widehat{\eta}(X)-1/2\right) - f_c(X)\right| |2\eta(X)-1| d P_\mathcal{X}  \\
  &\leq& 2(\tau + \epsilon)  2  \mathrm{P}_{\mathcal{X}}(X\in \mathcal{X}:|\eta(X)-1/2| > \epsilon, |\sigma_\tau\left(\widehat{\eta}(X) -1/2\right)|< 1)\\
  &\leq& 2(\tau + \epsilon) 2  \mathrm{P}_{\mathcal{X}}(X\in \mathcal{X}: |\sigma_\tau\left(\widehat{\eta}(X) -1/2\right)|< 1)\\
  &=& 2(\tau + \epsilon)2  \mathrm{P}_{\mathcal{X}}(X\in \mathcal{X}: |\widehat{\eta}(X) -1/2| < \tau) \\
  &\leq& 2(\tau + \epsilon) 2  \mathrm{P}_{\mathcal{X}}(X\in \mathcal{X}:  |\eta(X) -1/2| < \tau + \epsilon)\\
  &\stackrel{\text{Assumption 2.1}}{\leq}& 4c_0(\tau + \epsilon)^{q+1}.
\end{eqnarray*}
In the second equality, we have used the equivalence between $|\sigma_\tau\left(\widehat{\eta}(X) -1/2\right)|< 1$ and $|\widehat{\eta}(X) -1/2| < \tau$.
In the third inequality, we have used the condition $|\widehat{\eta}(X)-\eta(X)| \leq \epsilon$ for getting $|\eta(X) -1/2| < \tau + \epsilon$ when $|\widehat{\eta}(X) -1/2| < \tau$.

 Combining the above estimates, we get the desired bound and prove the lemma.
\end{proof}

\subsection{Upper Bound of the Estimation Errors}\label{subsec:estimation}
In this subsection, we derive an upper bound of the estimation errors  $\varepsilon(\Hat{f}_{T, T^\prime,\phi})- \varepsilon_{T,T^\prime}(\Hat{f}_{T, T^\prime,\phi}) + \varepsilon_{T,T^\prime}(f_\mathcal{H}) - \varepsilon(f_\mathcal{H})$. 

We first rewrite it by inserting $\varepsilon(f_c)$ and $\varepsilon_{T,T^\prime}(f_c)$:
\begin{eqnarray} 
&& \varepsilon(\Hat{f}_{T, T^\prime,\phi}) -\varepsilon_{T,T^\prime}(\Hat{f}_{T, T^\prime,\phi}) + \varepsilon_{T,T^\prime}(f_\mathcal{H}) - \varepsilon(f_\mathcal{H}) \nonumber\\
 &=&   \varepsilon(\Hat{f}_{T, T^\prime,\phi}) -\varepsilon(f_c)- (\varepsilon_{T,T^\prime}(\Hat{f}_{T, T^\prime,\phi})-\varepsilon_{T,T^\prime}(f_c)) \label{estimationerr1} \\
 &&+ \varepsilon_{T,T^\prime}(f_\mathcal{H}) - \varepsilon_{T,T^\prime}(f_c) - (\varepsilon(f_\mathcal{H})- \varepsilon(f_c)).   \label{estimationerr2}    
\end{eqnarray}
In other words, to bound $\varepsilon(\Hat{f}_{T, T^\prime,\phi})- \varepsilon_{T,T^\prime}(f_z) + (\varepsilon_{T,T^\prime}(f_\mathcal{H}) - \varepsilon(f_\mathcal{H}))$, we ought to  bound the R.H.S. of (\ref{estimationerr1}) and the R.H.S. of (\ref{estimationerr2}) respectively. 

\subsubsection{Upper bound of the first estimation error}
We will first handle the term $\varepsilon_{T, T^\prime}(f_\mathcal{H}) - \varepsilon_{T, T^\prime}(f_c) - (\varepsilon(f_\mathcal{H})- \varepsilon(f_c))$, which is the R.H.S. of (\ref{estimationerr2}).  

We define two random variables: \begin{equation}\label{xi}
    \xi(X):= \phi(f_\mathcal{H}(X)) - \phi(f_c(X)) \qquad \text{over } (\mathcal{X}, Q)
\end{equation} and 
\begin{equation}\label{kappa}
    \kappa(X):= \phi(-f_\mathcal{H}(X))- \phi(-f_c(X))\qquad \text{over } (\mathcal{X}, \mu).
\end{equation}
Then, we have
\begin{eqnarray}\label{two_variable_function}
    && \varepsilon_{T, T^\prime}(f_\mathcal{H}) - \varepsilon_{T, T^\prime}(f_c) - (\varepsilon(f_\mathcal{H})- \varepsilon(f_c)) \nonumber\\
    &=& \frac{s}{n}\sum_{i=1}^{n}\phi(f_\mathcal{H}(X_i)) + \frac{(1-s)}{n^\prime}\sum_{i=1}^{n^\prime}\phi(-f_\mathcal{H}(X^\prime_i)) \nonumber \\
    &&-  \left(\frac{s}{n}\sum_{i=1}^{n}\phi(f_c(X_i)) + \frac{(1-s)}{n^\prime}\sum_{i=1}^{n^\prime}\phi(-f_c(X^\prime_i))\right) \nonumber\\
    && - \left[s \int_{\mathcal{X} } \phi(f_\mathcal{H}(X))dQ + (1-s)\int_{\mathcal{X} } \phi(-f_\mathcal{H}(X))d\mu \right. \nonumber \\
    &&- \left.\left(s \int_{\mathcal{X} } \phi(f_c(X))dQ + (1-s)\int_{\mathcal{X} } \phi(-f_c(X))d\mu\right)\right] \nonumber\\
    &\stackrel{\eqref{xi}, \eqref{kappa}}{=}& s \left(\frac{1}{n}\sum_{i=1}^{n} \xi(X_i) - \mathrm{E}_Q[\xi(X)]\right) + (1-s) \left(\frac{1}{n^\prime}\sum_{i=1}^{n^\prime}\kappa(X^\prime_i)- \mathrm{E}_{\mu}[\kappa(X)] \right). 
\end{eqnarray}
We see that this is a function of random variables $\xi$ and $\kappa$. We can use Bernstein's inequality to estimate $\frac{1}{n}\sum_{i=1}^{n} \xi(X_i) - \mathrm{E}_Q[\xi(X)]$ and $\frac{1}{n^\prime}\sum_{i=1}^{n^\prime}\kappa(X^\prime_i)- \mathrm{E}_{\mu}[\kappa(X)] $ respectively. To apply Bernstein's inequality, we need first to establish upper bounds of the variance of $\xi$ and the variance of $\kappa$. 

Recall the Tsybakov noise condition we imposed in Assumption 2.1.
Here, we give an upper bound of the second moment and thereby the variance of $\phi(yf(X)) - \phi(yf_c(X))$ for any function $f: \mathcal{X} \to [-1,1]$ and some $y \in \{-1,1\}$ with the noise condition.
Note that $y \in \{-1,1\}$ is not random here. 
Denote $P_\mathcal{X}^y = Q$ for $y=1$, $P_\mathcal{X}^y = \mu$ for $y=-1$.

\begin{lemma} \label{secondmoment}
Let $0\leq q \leq \infty$. 
If the Tsybakov noise condition holds for some noise exponent $q$ and constant $c_0>0$, then for every function $f: \mathcal{X} \rightarrow [-1,1]$ and some $y \in \{-1,1\}$, there holds 
\begin{equation}
    \mathrm{E}_{P_\mathcal{X}^y}\left[\{\phi(yf(X))-\phi(yf_c(X))\}^2\right] \leq \frac{5}{s_y}(c_0)^{\frac{1}{q+1}} (\varepsilon(f) - \varepsilon(f_c)) ^{\frac{q}{q+1}},
\end{equation}
where \begin{equation}\label{s_y}
    s_y=s \quad \text{for } y=1 \ \text{and } s_y=1-s \quad \text{for } y=-1.
\end{equation}
\end{lemma}

\begin{proof}[Proof of Lemma \ref{secondmoment}]
Since $f(X) \in [-1,1]$, 
\begin{eqnarray*}
    \phi(yf(X)) - \phi(yf_c(X)) = 1-yf(X) - (1- yf_c(X)) = y(f_c(X) - f(X)). 
\end{eqnarray*}
It follows that 
\begin{eqnarray*}
    \mathrm{E}_{P_\mathcal{X}^y}[\{\phi(yf(X)) - \phi(yf_c(X))\}^2] 
    &=& \mathrm{E}_{P_\mathcal{X}^y}[y^2(f_c(X) - f(X))^2] \\
    &=& \mathrm{E}_{P_\mathcal{X}^y}[(f_c(X) - f(X))^2] \\
    &=& \int_{\mathcal{X}} (f_c(X) - f(X))^2 dP_{\mathcal{X}}^y. 
\end{eqnarray*}

Let $t>0$. Consider these two subsets of the domain $\mathcal{X}$: $\mathcal{X}_t^+ = \{X\in \mathcal{X}: |\eta(X)-1/2|>t\}$ and $\mathcal{X}_t^- = \{X\in \mathcal{X}: |\eta(X)-1/2|\leq t\}$. On the set  $\mathcal{X}_t^+$, we apply $|f_c(X) -f(X)| \leq 2$ and $\frac{|\eta(X)-1/2| }{t} >1$ and get
\begin{equation}\label{secmoment1}
    |f_c(X) -f(X)|^2 \leq 2 |f_c(X) -f(X)| \frac{|\eta(X)-1/2| }{t} = |f_c(X) -f(X)| \frac{|2\eta(X)-1| }{t}.
\end{equation}
On the set $\mathcal{X}_t^-$, we have \begin{equation}\label{secmoment2}
    |f_c(X) -f(X)|^2 \leq 4. 
\end{equation}
Recall from the proof of Lemma \ref{lemma:approx}, we derived the expression for  $\varepsilon(f) -\varepsilon(f_c)$ for any $|f(X)|\leq 1$ given by 
\begin{equation}\label{expression}
    \varepsilon(f) -\varepsilon(f_c)  
  = \int_\mathcal{X}  \left| f_c(X) - f(X)\right|  |2\eta(X)-1|\;d P_\mathcal{X}.
\end{equation}
It follows from the Tsybakov noise condition that 
\begin{eqnarray*}
  && \mathrm{E}_{P_\mathcal{X}^y}\left[\{\phi(yf(X))-\phi(yf_c(X))\}^2\right] \\
   &=&  \int_\mathcal{X} (f_c(X) -f(X))^2 d{P_\mathcal{X}^y}  \\
   &=& \int_{\mathcal{X}_t^-} (f_c(X) -f(X))^2 d{P_\mathcal{X}^y} + \int_{\mathcal{X}_t^+} (f_c(X) -f(X))^2 d{P_\mathcal{X}^y}\\
   &\stackrel{\eqref{secmoment1}, \eqref{secmoment2}}{\leq} & 4 {P_\mathcal{X}^y}(\{x\in \mathcal{X}: |\eta(X)-1/2| \leq t\}) + 
   \frac{1}{t}\int_{\mathcal{X}_t^+}  |f_c(X) -f(X)| |2\eta(X)-1| d{P_\mathcal{X}^y}\\
   &\stackrel{\because \mathcal{X}_t^+ \subseteq \mathcal{X}}{\leq} & 4 {P_\mathcal{X}^y}(\{x\in \mathcal{X}: |\eta(X)-1/2| \leq t\}) + 
   \frac{1}{t}\int_{\mathcal{X}}  |f_c(X) -f(X)| |2\eta(X)-1| d{P_\mathcal{X}^y}\\
   &\stackrel{\text{Assumption 2.1}, \eqref{s_y}, \eqref{expression}}{\leq} & \frac{1}{s_y}\left\{4c_0t^q + \frac{1}{t}(\varepsilon(f) - \varepsilon(f_c))\right\}.
\end{eqnarray*}
Here, in the last inequality, we have used the relations $d{P_\mathcal{X}^y} = dQ \leq \frac{1}{s}dP_\mathcal{X}$ for $y=1$ and $d{P_\mathcal{X}^y} \leq \frac{1}{1-s}dP_\mathcal{X}$ for $y=-1$. 

Now set $t= \left(\frac{\varepsilon(f) - \varepsilon(f_c)}{c_0}\right)^{1/(q+1)}$, the proof is complete.
\end{proof}

The one-side Bernstein's inequality asserts for a random variable $\xi$ on ${\mathcal X}$ with mean $\mu$ and variance $\sigma^2$ satisfying $|\xi - \mu|\le B$ almost surely that for a random i.i.d. sample $\{X_i\}_{i=1}^n$ and every
$t>0$, 
\begin{equation}\label{Bernstein}
 \hbox{Prob} \left\{\mu- \frac{1}{n}\sum_{i=1}^n \xi(X_i) >t \right\} \leq \exp \left\{-
{n t^{2} \over 2(\sigma^2 +{1 \over 3} B t)}\right\}.   
\end{equation}

We now apply Berstein's inequality to obtain a high probability upper bound of  $\varepsilon_{T,T^\prime}(f_\mathcal{H}) - \varepsilon_{T,T^\prime}(f_c) - (\varepsilon(f_\mathcal{H})- \varepsilon(f_c))$.
\begin{lemma}\label{lemma: estimationI}
   Suppose the noise condition (Assumption 2.1) holds for some $q\in [0, \infty]$ and constant $c_0>0$. For any $0 <\delta < 1$, with probability $1-\frac{\delta}{2}$, we have 
    \begin{eqnarray}\label{estimationI_eqn}
        && \varepsilon_{T,T^\prime}(f_\mathcal{H}) - \varepsilon_{T,T^\prime}(f_c) - (\varepsilon(f_\mathcal{H})- \varepsilon(f_c)) \nonumber\\
         &\leq&  C_{q} \left(\log\left(\frac{4}{\delta}\right)\right)\left(s^{\frac{1}{q+2}}\left(\frac{1}{n}\right)^{\frac{q+1}{q+2}}+(1-s)^{\frac{1}{q+2}}\left(\frac{1}{n^\prime}\right)^{\frac{q+1}{q+2}}\right) + \frac{(\varepsilon(f_\mathcal{H}) - \varepsilon(f_c))}{2},
    \end{eqnarray}
    where $C_{q}$ is a positive constant depending only on $q$ and $c_0$.
\end{lemma}
\begin{proof}[Proof of Lemma \ref{lemma: estimationI}]
Recall that from Lemma \ref{secondmoment}, we know for every function $f: \mathcal{X} \rightarrow [-1,1]$ and some $y \in \{-1,1\}$,
\begin{equation*}
\mathrm{E}_{P_{\mathcal{X}}^y}\left[\{\phi(yf(X))-\phi(yf_c(X))\}^2\right] \leq \frac{5}{s_y}(c_0)^{\frac{1}{q+1}} (\varepsilon(f) - \varepsilon(f_c)) ^{\frac{q}{q+1}}.
\end{equation*}
Then the variance $\sigma^2$ of the random variable $\xi$ for $y=1$ and $\kappa$ for $y=-1$ is bounded by \begin{equation}\label{sigma2_bound}
    \sigma^2\leq  \frac{5}{s_y}(c_0)^{\frac{1}{q+1}} (\varepsilon(f_\mathcal{H}) - \varepsilon(f_c)) ^{\frac{q}{q+1}}, 
\end{equation}
where $s_y=s$ for $y=1$ and $s_y=1-s$ for $y=-1$.

By the one-sided Bernstein’s inequality \eqref{Bernstein}, for any $\epsilon >0$, there holds, with probability at least $1- \exp\left(-\frac{n\epsilon^2}{2(\sigma^2+2\epsilon/3)}\right)$, 
\begin{equation*}
    \left(\frac{1}{n}\sum_{i=1}^{n} \xi(X_i) - \mathrm{E}_Q[\xi(X)]\right) \leq \epsilon. 
\end{equation*}
Setting this confidence bound to be $1-\frac{\delta}{4}$, we obtain a quadratic equation for $\epsilon$ as follow
\begin{equation*}
    \log\left(\frac{4}{\delta}\right) = \frac{n\epsilon^2}{2(\sigma^2+2\epsilon/3)}.
\end{equation*}
We solve this equation and get a positive solution $\epsilon^*$ given by
\begin{eqnarray*}
    \epsilon^*  &=& \frac{\frac{4}{3} \log\left(\frac{4}{\delta}\right)+ \sqrt{\frac{16}{9}\log^2\left(\frac{4}{\delta}\right)+8n\sigma^2\log\left(\frac{4}{\delta}\right) }}{2n}\\
    &\leq&  \frac{2}{3n} \log\left(\frac{4}{\delta}\right) + \frac{2}{3n} \log\left(\frac{4}{\delta}\right) + \frac{\sqrt{2\sigma^2\log\left(\frac{4}{\delta}\right)}}{\sqrt{n}}\\    &\stackrel{\eqref{sigma2_bound}}{\leq}& \frac{4}{3n} \log\left(\frac{4}{\delta}\right) + 4 \frac{\sqrt{\log\left(\frac{4}{\delta}\right)}}{\sqrt{ns}}  (c_0)^{\frac{1}{2(q+1)}} (\varepsilon(f_\mathcal{H}) - \varepsilon(f_c)) ^{\frac{q}{2(q+1)}}.
\end{eqnarray*}
Consider the classic  
 Young's Inequality for products \citep{young1912classes} given by 
 \begin{equation}\label{Young}
        a\cdot b \leq \frac{a^p}{p}+ \frac{b^{p^*}}{p^*} \qquad \text{with    } a\geq 0,b \geq 0, p>1,p^*>1 \text{ and } \frac{1}{p}+ \frac{1}{p^*} = 1.
    \end{equation}
Applying  Young's Inequality with $a= 4 \frac{\sqrt{\log\left(\frac{4}{\delta}\right)}}{\sqrt{ns}}  (c_0)^{\frac{1}{2(q+1)}} , b= (\varepsilon(f_\mathcal{H}) - \varepsilon(f_c)) ^{\frac{q}{2(q+1)}}$, $p= \frac{2(q+1)}{q+2}, p^* = \frac{2(q+1)}{q}$, we got
\begin{eqnarray}\label{epsilon*}
     \epsilon^* &\leq&  \frac{4}{3n} \log\left(\frac{4}{\delta}\right) + 4 \frac{\sqrt{\log\left(\frac{4}{\delta}\right)}}{\sqrt{ns}}  (c_0)^{\frac{1}{2(q+1)}} (\varepsilon(f_\mathcal{H}) - \varepsilon(f_c)) ^{\frac{q}{2(q+1)}} \nonumber\\
     &\leq& \frac{4}{3n} \log\left(\frac{4}{\delta}\right) + \left(\frac{q+2}{2(q+1)}\right)\left(4 \frac{\sqrt{\log\left(\frac{4}{\delta}\right)}}{\sqrt{ns}}  (c_0)^{\frac{1}{2(q+1)}}\right)^{\frac{2(q+1)}{q+2}} + 
 \frac{\varepsilon(f_\mathcal{H}) - \varepsilon(f_c)}{{\frac{2(q+1)}{q}}}.
\end{eqnarray}

Similarly, we use the same approach to obtain that, with a probability of at least $1-\frac{\delta}{4}$,
\begin{eqnarray}
\label{kappabound}
   && \left(\frac{1}{n^\prime}\sum_{i=1}^{n^\prime} \kappa(X^\prime_i) - \mathrm{E}_\mu[\kappa(X^\prime)]\right) \nonumber\\
    &\leq &\frac{4}{3n^\prime} \log\left(\frac{4}{\delta}\right) 
    + \left(\frac{q+2}{2(q+1)}\right)\left(4 \frac{\sqrt{\log\left(\frac{4}{\delta}\right)}}{\sqrt{n^\prime(1-s))}}  (c_0)^{\frac{1}{2(q+1)}}\right)^{\frac{2(q+1)}{q+2}} + 
 \frac{\varepsilon(f_\mathcal{H}) - \varepsilon(f_c)}{{\frac{2(q+1)}{q}}} . 
\end{eqnarray}
Now, combining the above error bounds, we get, with probability of at least $1-\frac{\delta}{2}$,
    \begin{eqnarray*}
         &&\varepsilon_{T, T'}(f_\mathcal{H}) - \varepsilon_{T, T'}(f_c) - (\varepsilon(f_\mathcal{H})- \varepsilon(f_c)) \\  &\stackrel{\eqref{two_variable_function}}{=}& s \left(\frac{1}{n}\sum_{i=1}^{n} \xi(X_i) - \mathrm{E}[\xi(X)]\right) + (1-s) \left(\frac{1}{n^\prime}\sum_{i=1}^{n^\prime}\kappa(X^\prime_i)- \mathrm{E}[\kappa(X^\prime)] \right)\\
          &\stackrel{\eqref{epsilon*}, \eqref{kappabound}}{\leq}& s \left(  \frac{4}{3n} \log\left(\frac{4}{\delta}\right) + \left(\frac{q+2}{2(q+1)}\right)\left(4 \frac{\sqrt{\log\left(\frac{4}{\delta}\right)}}{\sqrt{ns}}  (c_0)^{\frac{1}{2(q+1)}}\right)^{\frac{2(q+1)}{q+2}} + 
 \frac{\varepsilon(f_\mathcal{H}) - \varepsilon(f_c)}{{\frac{2(q+1)}{q}}}\right)\\
 && + (1-s)\left[\frac{4}{3n^\prime} \log\left(\frac{4}{\delta}\right) + \left(\frac{q+2}{2(q+1)}\right)\left(4 \frac{\sqrt{\log\left(\frac{4}{\delta}\right)}}{\sqrt{n^\prime(1-s))}}  (c_0)^{\frac{1}{2(q+1)}}\right)^{\frac{2(q+1)}{q+2}} \right.\\
 &&\quad\quad\quad\quad\quad  \left. + 
 \frac{\varepsilon(f_\mathcal{H}) - \varepsilon(f_c)}{{\frac{2(q+1)}{q}}}\right]\\
 &\leq& \frac{4}{3} \left(\frac{s}{n}+ \frac{1-s}{n^\prime}\right)  \log\left(\frac{4}{\delta}\right) \\
 &&+ 4^{\frac{2(q+1)}{q+2}}(c_0)^{\frac{q}{2(q+1)}}
 \left(\log\left(\frac{4}{\delta}\right)\right)^{\frac{q+1}{q+2}} \left(s^{\frac{1}{q+2}}\left(\frac{1}{n}\right)^{\frac{q+1}{q+2}}+(1-s)^{\frac{1}{q+2}}\left(\frac{1}{n^\prime}\right)^{\frac{q+1}{q+2}}\right) \\
 &&+ \left(\frac{q}{2(q+1)}\right)(\varepsilon(f_\mathcal{H}) - \varepsilon(f_c))\\
 &\leq& C_{q} \left(\log\left(\frac{4}{\delta}\right)\right)\left(s^{\frac{1}{q+2}}\left(\frac{1}{n}\right)^{\frac{q+1}{q+2}}+(1-s)^{\frac{1}{q+2}}\left(\frac{1}{n^\prime}\right)^{\frac{q+1}{q+2}}\right) + \frac{(\varepsilon(f_\mathcal{H}) - \varepsilon(f_c))}{2},
    \end{eqnarray*}
    where $C_{q}$ is a positive constant depending only on $q$ and $c_0$. 
    In the last inequality, we have used the facts that $s \leq s^{\frac{1}{q+2}}, 1-s \leq (1-s)^{\frac{1}{q+2}}$ for all $s \in (0,1]$ and $q>0$; and that $\frac{1}{n} \leq \left(\frac{1}{n}\right)^{\frac{q+1}{q+2}}, \frac{1}{n^\prime} \leq \left(\frac{1}{n^\prime}\right)^{\frac{q+1}{q+2}}$.   
    The proof is complete.
\end{proof}

\subsubsection{Upper bound of the second  estimation error}
Next, we will handle estimate the term $\varepsilon(\Hat{f}_{T, T^\prime,\phi}) -\varepsilon(f_c)- (\varepsilon_{T, T'}(\Hat{f}_{T, T^\prime,\phi})-\varepsilon_{T, T'}(f_c))$, which is the R.H.S. of \eqref{estimationerr1}.
We will derive an upper bound for this error term using a concentration inequality in terms of covering numbers.

For $\epsilon >0$, denote by $\mathcal{N}(\epsilon,\mathcal{H}):=\mathcal{N}(\epsilon,\mathcal{H},\|\cdot\|_\infty)$ the $\epsilon$-covering number of a set of functions $\mathcal{H}$ with respect to $\|\cdot\|_\infty:=  \text{ess} \sup_{x\in\mathcal{X}}|f(x)|$. More specifically, $\mathcal{N}(\epsilon,\mathcal{H})$ is the minimal $M\in \NN$ such that there exists functions $\{f_1,\ldots, f_M\}\in \mathcal{H}$ satisfying
\begin{equation}
    \min_{1\leq i \leq M} \|f-f_i\|_\infty \leq \epsilon, \qquad \forall f\in \mathcal{H}.
\end{equation}

The following Proposition gives an upper bound of the $\epsilon$-covering number of a function set generated by neural networks.
\begin{proposition}[Lemma 3 of \citep{suzuki2018adaptivity}, Lemma 5 of \citep{schmidt2020nonparametric}]\label{generalcovering}
    For any $\epsilon>0$, we have   \begin{equation}\label{logN}
        \log \mathcal{N}(\epsilon,\mathcal{F}(L, w, s, K)) \leq 2L (s+1)\log (\epsilon^{-1}(L+1)(w+1)(\max\{K,1\})).
    \end{equation}
\end{proposition}
We apply the above Proposition to obtain an estimate of the $\epsilon$-covering number of our hypothesis space $\mathcal{H}_\tau$. 
\begin{corollary}[$\epsilon$-covering number of hypothesis space $\mathcal{H}_\tau$]\label{covering}
    Recall the hypothesis space defined in Definition 3.1 in the main paper for some $0<\tau\leq 1$:
    $$\mathcal{H}_\tau:= \text{span}\left\{\sigma_\tau(f(X)): f\in \mathcal{F}(L^*, w^*, s^*, K^*) \right\}$$ with $\alpha,r >0$, integers $m \geq 1$ and $N \geq \max \left\{(\alpha +1)^d, (r+1)e^d\right\}$, $L^*= 8+ (m+5)(1+\lceil \log_2(\max\{d,\alpha\}) \rceil)$, $w^* = 6(d+ \lceil \alpha \rceil )N$,
   $s^*= 141 (d+\alpha +1)^{3+d}N (m+6)$, and $K^*= 1$. 
   
    For any $0 < \epsilon \leq 1$, we have 
    \begin{equation}\label{coveringeqn}
        \log \mathcal{N}(\epsilon,\mathcal{H}_\tau) \leq c_{\alpha, d} m^2  N \log ((\tau\epsilon)^{-1}mN),
    \end{equation}
    where $c_{\alpha, d}$ is a positive constant independent of $r, m , N, \tau$ or $\epsilon$. 
\end{corollary}

\begin{proof}[Proof of Corollary \ref{covering}]
It follows from Proposition \ref{generalcovering} that 
    \begin{eqnarray*}
    &&  \log \mathcal{N}(\epsilon,\mathcal{F}(L^*, w^*, s^*, K^*)) \\
    &\stackrel{\eqref{logN}}{\leq}& 2L^* (s^*+1)\log (\epsilon^{-1}(L^*+1)(w^*+1)(\max\{K^*,1\}))\\
         &\leq& 2 (8+ (m+5)(1+\lceil \log_2(\max\{d,\alpha\}) \rceil)) (141 (d+\alpha +1)^{3+d}N (m+6)+1)\\
         && \log (\epsilon^{-1} (8+ (m+5)(1+\lceil \log_2(\max\{d,\alpha\}) \rceil)+1)(6(d+ \lceil \alpha \rceil )N+1))   \\
         &\leq& c_{\alpha, d} m^2  N \log (\epsilon^{-1}mN), 
    \end{eqnarray*}
where $c_{\alpha, d}$ is a positive constant independent of $r, m,N , \tau$ or $\epsilon$. 

Recall the function $\sigma_\tau$  
for $0 < \tau \leq 1$:
\begin{align*}
    \sigma_\tau (x) &:= \sigma\left(\frac{x}{\tau}\right) - \sigma\left(\frac{x}{\tau} -1\right) - \sigma\left(-\frac{x}{\tau}\right)+\sigma\left(-\frac{x}{\tau}+1\right) 
    =
    \begin{cases} 
      1, & \text{if } x \geq \tau, \\
      \frac{x}{\tau}, & \text{if }  x\in [-\tau,\tau),\\
      -1, & \text{if } x < -\tau.
    \end{cases} 
\end{align*}
We observe that 
\begin{equation} 
    |\sigma_\tau(u) - \sigma_\tau(v)| \leq \frac{1}{\tau} |u-v|, \qquad \forall u,v \in \RR.
\end{equation}
It follows that for all $f,f^\prime \in \mathcal{F}(L^*, w^*, s^*, K^*)$,  $X\in \mathcal{X}$,
\begin{eqnarray*}
\left|\sigma_\tau(f(X)) - \sigma_\tau(f^\prime(X))\right| \leq \frac{1}{\tau} |f(X)-f^\prime(X)| \leq \frac{1}{\tau} \|f-f^\prime\|_\infty.
\end{eqnarray*}

 Thus, we have 
     \begin{equation*}
        \log \mathcal{N}(\epsilon,\mathcal{H}_\tau) \leq \log \mathcal{N}(\tau\epsilon,\mathcal{F}(L^*,  w^*, s^*, K^*)) = c_{\alpha, d} m^2 N \log ((\tau\epsilon)^{-1}mN).
    \end{equation*}
\end{proof}


Next, by applying the one-side Bernstein's Inequality (given in \eqref{Bernstein}), we have the following result for a fixed function $f: {\mathcal X} \to [-1, 1]$.

\begin{lemma}\label{Bernsteintau}
Suppose the Tsybakov noise condition (Assumption 2.1) holds for some $q\in [0, \infty]$ and $c_0 >0$. Let $f: {\mathcal X} \to [-1, 1]$. Then for every
$\epsilon>0$, there holds
$$\frac{\varepsilon (f) - \varepsilon (f_c) - \left(\varepsilon_{T, T'} (f) - \varepsilon_{T, T'} (f_c)\right)}{\left(\left(\varepsilon (f) - \varepsilon (f_c)\right)^{\frac{q}{q+1}} + \epsilon^{\frac{q}{q+1}}\right)^{1/2}} \leq 2 \epsilon^{1-{\frac{q}{2(q+1)}}} 
$$
with probability at least 
\begin{eqnarray*}
1-  \exp \left\{-
{n \epsilon^{2-{\frac{q}{q+1}}} \over s\left(10 (c_0)^{\frac{1}{q+1}} +2 \epsilon^{1-{\frac{q}{q+1}}}\right)}\right\} - \exp \left\{-
{n' \epsilon^{2-{\frac{q}{q+1}}} \over (1-s) \left(10 (c_0)^{\frac{1}{q+1}} +2 \epsilon^{1-{\frac{q}{q+1}}}\right)}\right\}. 
\end{eqnarray*}
\end{lemma}
\begin{proof}[Proof of Lemma \ref{Bernsteintau}]
Consider the random variable $\xi (X) = s\left(\phi(f(X)) - \phi(f_c(X))\right)$ on $({\mathcal X}, Q)$ with $f: {\mathcal X} \to [-1, 1]$ and $s\in (0, 1)$. It satisfies $|\xi| \leq 2s$ and thereby $|\xi - \mathrm{E}[\xi]|\le 4 s$ almost surely. Moreover, by Lemma \ref{secondmoment}, when the noise condition is satisfied, we have $\sigma^2 := \text{ Var}[\xi] \leq 5 s (c_0)^{\frac{1}{q+1}}\left(\varepsilon (f) - \varepsilon (f_c)\right)^{\frac{q}{q+1}}$.  
Applying the one-side Bernstein inequality \eqref{Bernstein} to this random variable and $t = \left(\left(\varepsilon (f) - \varepsilon (f_c)\right)^{\frac{q}{q+1}} + \epsilon^{\frac{q}{q+1}}\right)^{1/2} \epsilon^{1-{\frac{q}{2(q+1)}}}$ with $\epsilon >0$, we know that 
\begin{eqnarray*} 
&&s\left(\int_{{\mathcal X}} \left(\phi(f(X)) - \phi(f_c(X))\right) d Q - \frac{1}{n}\sum_{i=1}^n \left(\phi(f(X_i)) - \phi(f_c(X_i))\right)\right) \\
&\leq& \left(\left(\varepsilon (f) - \varepsilon (f_c)\right)^{\frac{q}{q+1}} + \epsilon^{\frac{q}{q+1}}\right)^{1/2} \epsilon^{1-{\frac{q}{2(q+1)}}} 
\end{eqnarray*}
with a probability of at least 
\begin{eqnarray*}
&& 1-  \exp \left\{-
{n \left(\left(\varepsilon (f) - \varepsilon (f_c)\right)^{\frac{q}{q+1}} + \epsilon^{\frac{q}{q+1}}\right) \epsilon^{2-{\frac{q}{q+1}}} \over 2\left(5 s (c_0)^{\frac{1}{q+1}}\left(\varepsilon (f) - \varepsilon (f_c)\right)^{\frac{q}{q+1}} +{4 s \over 3}\left(\left(\varepsilon (f) - \varepsilon (f_c)\right)^{\frac{q}{q+1}} + \epsilon^{\frac{q}{q+1}}\right)^{1/2} \epsilon^{1-{\frac{q}{2(q+1)}}}\right)}\right\} \\
&\geq& 1-  \exp \left\{-
{n \epsilon^{2-{\frac{q}{q+1}}} \over s\left(10 (c_0)^{\frac{1}{q+1}} +2 \epsilon^{1-{\frac{q}{q+1}}}\right)}\right\}.
\end{eqnarray*}

In the same way, applying the one-side Bernstein inequality to the random variable \\$ (1-s)\left(\phi(-f(X)) - \phi(-f_c(X))\right)$ on $({\mathcal X}, \mu)$, we find that 
\begin{eqnarray*} 
&&(1-s) \left(\int_{{\mathcal X}} \left(\phi(-f(X)) - \phi(-f_c(X)) \right) d \mu - \frac{1}{n'}\sum_{i=1}^{n'} \left(\phi(-f(X'_i)) - \phi(-f_c(X'_i))\right)\right) \\
&\leq& \left(\left(\varepsilon (f) - \varepsilon (f_c)\right)^{\frac{q}{q+1}} + \epsilon^{\frac{q}{q+1}}\right)^{1/2} \epsilon^{1-{\frac{q}{2(q+1)}}} 
\end{eqnarray*}
with probability at least 
\begin{eqnarray*}
1-  \exp \left\{-
{n' \epsilon^{2-{\frac{q}{q+1}}} \over (1-s) \left(10 (c_0)^{\frac{1}{q+1}} +2 \epsilon^{1-{\frac{q}{q+1}}}\right)}\right\}. 
\end{eqnarray*}
Combining the above two estimates proves the desired inequality. 
\end{proof}

\begin{lemma}\label{uniformBernstein}
Let ${\mathcal H}$ be a set of measurable functions from ${\mathcal X}$ to $[-1, 1]$. Suppose the noise condition (Assumption 2.1) holds for some $q\in [0, \infty]$ and $c_0 >0$. Then for every
$\epsilon>0$, 
$$\sup_{f\in {\mathcal H}} \frac{\varepsilon (f) - \varepsilon (f_c) - \left(\varepsilon_{T, T'} (f) - \varepsilon_{T, T'} (f_c)\right)}{\left(\left(\varepsilon (f) - \varepsilon (f_c)\right)^{\frac{q}{q+1}} + \epsilon^{\frac{q}{q+1}}\right)^{1/2}} \leq 5 \epsilon^{1-{\frac{q}{2(q+1)}}}
$$
with probability at least 
\begin{eqnarray*}
1-  {\mathcal N}\left(\epsilon, {\mathcal H}\right) 
\left\{\exp \left\{-
{n \epsilon^{2-{\frac{q}{q+1}}} \over s\left(10 (c_0)^{\frac{1}{q+1}} +2 \epsilon^{1-{\frac{q}{q+1}}}\right)}\right\}
 +\exp \left\{-
{n' \epsilon^{2-{\frac{q}{q+1}}} \over (1-s) \left(10 (c_0)^{\frac{1}{q+1}} +2 \epsilon^{1-{\frac{q}{q+1}}}\right)}\right\}\right\}. 
\end{eqnarray*}
\end{lemma}
\begin{proof}[Proof of Lemma \ref{uniformBernstein}]
Let $\{f_j\}_{j=1}^{\mathcal N}$ be an $\epsilon$-net of ${\mathcal H}$ with ${\mathcal N}:={\mathcal N}\left(\epsilon, {\mathcal H}\right)$. Then for each $f\in {\mathcal H}$ there exists some $j\in \{1, \ldots, {\mathcal N}\}$ such that $\|f-f_j\|_\infty \leq \epsilon$. It follows that $\|\phi(f(X)) - \phi(f_j(X))\|_\infty \leq \epsilon$ and then 
\begin{equation}\label{fjI}
  \left|\varepsilon (f) - \varepsilon (f_j)\right| \leq s \epsilon + (1-s) \epsilon =\epsilon   
\end{equation} 
and 
\begin{equation}\label{fjII}
\left|\varepsilon_{T, T'} (f) - \varepsilon_{T, T'} (f_j)\right| \leq \epsilon.   
\end{equation}  
It also implies 
$$ \left(\varepsilon (f_j) - \varepsilon (f_c)\right)^{\frac{q}{q+1}} + \epsilon^{\frac{q}{q+1}} \stackrel{\eqref{fjI}}{\leq} \left(\varepsilon (f) - \varepsilon (f_c)\right)^{\frac{q}{q+1}} + \epsilon^{\frac{q}{q+1}} +  \epsilon^{\frac{q}{q+1}} \leq 2\left(\left(\varepsilon (f) - \varepsilon (f_c)\right)^{\frac{q}{q+1}} + \epsilon^{\frac{q}{q+1}}\right)  $$
and thereby 
\begin{eqnarray*} 
&& \frac{\varepsilon (f) - \varepsilon (f_c) - \left(\varepsilon_{T, T'} (f) - \varepsilon_{T, T'} (f_c)\right)}{\left(\left(\varepsilon (f) - \varepsilon (f_c)\right)^{\frac{q}{q+1}} + \epsilon^{\frac{q}{q+1}}\right)^{1/2}} \\
&\stackrel{\eqref{fjI}, \eqref{fjII}}{\leq}& \frac{\varepsilon (f_j) - \varepsilon (f_c) - \left(\varepsilon_{T, T'} (f_j) - \varepsilon_{T, T'} (f_c)\right) + 2 \epsilon}{\left(\left(\varepsilon (f) - \varepsilon (f_c)\right)^{\frac{q}{q+1}} + \epsilon^{\frac{q}{q+1}}\right)^{1/2}} \\
&\leq& \frac{\varepsilon (f_j) - \varepsilon (f_c) - \left(\varepsilon_{T, T'} (f_j) - \varepsilon_{T, T'} (f_c)\right)}{\left(\frac{1}{2}\left(\left(\varepsilon (f_j) - \varepsilon (f_c)\right)^{\frac{q}{q+1}} + \epsilon^{\frac{q}{q+1}}\right)\right)^{1/2}}  + 2 \epsilon^{1-{\frac{q}{2(q+1)}}}. 
\end{eqnarray*}
This tells us that 
$$ \frac{\varepsilon (f) - \varepsilon (f_c) - \left(\varepsilon_{T, T'} (f) - \varepsilon_{T, T'} (f_c)\right)}{\left(\left(\varepsilon (f) - \varepsilon (f_c)\right)^{\frac{q}{q+1}} + \epsilon^{\frac{q}{q+1}}\right)^{1/2}} > 5 \epsilon^{1-{\frac{q}{2(q+1)}}} $$
implies 
$$  \frac{\varepsilon (f_j) - \varepsilon (f_c) - \left(\varepsilon_{T, T'} (f_j) - \varepsilon_{T, T'} (f_c)\right)}{\left(\left(\varepsilon (f_j) - \varepsilon (f_c)\right)^{\frac{q}{q+1}} + \epsilon^{\frac{q}{q+1}}\right)^{1/2}}  >2 \epsilon^{1-{\frac{q}{2(q+1)}}}. $$
Therefore, the event 
$$ \sup_{f\in {\mathcal H}} \frac{\varepsilon (f) - \varepsilon (f_c) - \left(\varepsilon_{T, T'} (f) - \varepsilon_{T, T'} (f_c)\right)}{\left(\left(\varepsilon (f) - \varepsilon (f_c)\right)^{\frac{q}{q+1}} + \epsilon^{\frac{q}{q+1}}\right)^{1/2}} > 5 \epsilon^{1-{\frac{q}{2(q+1)}}} $$
is contained in the event 
$$ \cup_{j=1}^{\mathcal N} \left\{\frac{\varepsilon (f_j) - \varepsilon (f_c) - \left(\varepsilon_{T, T'} (f_j) - \varepsilon_{T, T'} (f_c)\right)}{\left(\left(\varepsilon (f_j) - \varepsilon (f_c)\right)^{\frac{q}{q+1}} + \epsilon^{\frac{q}{q+1}}\right)^{1/2}}  >2 \epsilon^{1-{\frac{q}{2(q+1)}}}\right\}. $$
Then, our conclusion follows from Lemma \ref{Bernsteintau}. 
\end{proof}

We shall next apply the covering number estimate of the hypothesis space $\mathcal{H}_\tau$ (given in Corollary \ref{covering}) and Lemma \ref{uniformBernstein} to derive a high probability upper bound of the estimation
error term $\varepsilon (f) - \varepsilon (f_c) - \left(\varepsilon_{T, T'} (f) - \varepsilon_{T, T'} (f_c)\right)$, which is the R.H.S. of \eqref{estimationerr1}.

\begin{lemma}\label{lemma:estimationII}
     Let $0\leq q \leq \infty$. Let $\alpha,r >0$, and integers $m \geq 1$ and \\$N \geq \max \left\{(\alpha +1)^d, (r+1)e^d\right\}$.
    Suppose the noise condition (Assumption 2.1) holds for some $q$ and constant $c_0>0$. Suppose $\tau \geq \max\left\{\frac{1}{n}, \frac{1}{n^\prime}\right\}$. 
    For any $0 <\delta < 1$ and $n\geq 3$, with probability $1-\frac{\delta}{2}$, we have 
\begin{eqnarray*}
  &&   \varepsilon (\Hat{f}_{T, T^\prime,\phi}) - \varepsilon (f_c) - \left(\varepsilon_{T, T'} (f) - \varepsilon_{T, T'} (f_c)\right)\\
 &\leq&  C_{q,\alpha, d} \left(\log \left(\frac{4}{\delta}\right) + m^2 N (\log(nmN)+\log(n^\prime mN))\right)^{\frac{q+2}{q+1}} \\
 &&\quad\quad\cdot\left(  \max\left\{  \frac{s}{n}
\log \left(\frac{n}{s}\right), \frac{(1-s)}{n^\prime}\log \left(\frac{n^\prime}{1-s}\right) \right\} \right)^{\frac{q+2}{q+1}} 
+ \frac{\varepsilon (\Hat{f}_{T, T^\prime,\phi}) - \varepsilon (f_c)}{2},
\end{eqnarray*}
where $C_{q,\alpha, d}$ is a positive constant depending only on $q,c_0, \alpha$, and $d$. 
\end{lemma}

\begin{proof}[Proof of Lemma \ref{lemma:estimationII}]
    Lemma \ref{uniformBernstein} tells us that, for every $0<\epsilon \leq 1$, with probability at least 
    \begin{eqnarray*}
&&1-  {\mathcal N}\left(\epsilon, {\mathcal H_\tau}\right) \left\{\exp \left\{-
{n \epsilon^{2-{\frac{q}{q+1}}} \over s\left(10 (c_0)^{\frac{1}{q+1}} +2 \epsilon^{1-{\frac{q}{q+1}}}\right)}\right\}
+\exp \left\{-
{n' \epsilon^{2-{\frac{q}{q+1}}} \over (1-s) \left(10 (c_0)^{\frac{1}{q+1}} +2 \epsilon^{1-{\frac{q}{q+1}}}\right)}\right\}\right\}\\
&\geq& 1-  {\mathcal N}\left(\epsilon, {\mathcal H_\tau}\right) \left\{\exp \left\{-
{n \epsilon^{2-{\frac{q}{q+1}}} \over s\left(10 (c_0)^{\frac{1}{q+1}} +2 \right)}\right\} +\exp \left\{-
{n' \epsilon^{2-{\frac{q}{q+1}}} \over (1-s) \left(10 (c_0)^{\frac{1}{q+1}} +2 \right)}\right\}\right\}\\
&\stackrel{\eqref{coveringeqn}}{\geq}& 1- \exp \left\{c_{\alpha, d} m^2 N \log ((\tau\epsilon)^{-1}mN) -
{n \epsilon^{2-{\frac{q}{q+1}}} \over s\left(10 (c_0)^{\frac{1}{q+1}} +2 \right)}\right\}\\
&& -\exp \left\{c_{\alpha, d} m^2 N \log ((\tau\epsilon)^{-1}mN)-
{n' \epsilon^{2-{\frac{q}{q+1}}} \over (1-s) \left(10 (c_0)^{\frac{1}{q+1}} +2 \right)}\right\} ,
\end{eqnarray*} 
there holds
$$\sup_{f\in {\mathcal H}} \frac{\varepsilon (f) - \varepsilon (f_c) - \left(\varepsilon_{T, T'} (f) - \varepsilon_{T, T'} (f_c)\right)}{\left(\left(\varepsilon (f) - \varepsilon (f_c)\right)^{\frac{q}{q+1}} + \epsilon^{\frac{q}{q+1}}\right)^{1/2}} \leq 5 \epsilon^{1-{\frac{q}{2(q+1)}}},
$$
which implies 
\begin{equation}\label{5epsilon}
    \varepsilon (f) - \varepsilon (f_c) - \left(\varepsilon_{T, T'} (f) - \varepsilon_{T, T'} (f_c)\right)\leq 5 \epsilon^{1-{\frac{q}{2(q+1)}}}\left(\left(\varepsilon (f) - \varepsilon (f_c)\right)^{\frac{q}{q+1}} + \epsilon^{\frac{q}{q+1}}\right)^{1/2}, \quad \forall f\in \mathcal{H}_\tau. 
\end{equation}

We choose $\tau$ satisfying $\tau \geq \max\{\frac{1}{n}, \frac{1}{n^\prime}\}$.
Then we know $\tau^{-1} \leq n$ and $\tau^{-1} \leq n^\prime$.
We set the above confidence bound to be at least $1-\delta/2$. We need to find $\epsilon$ that satisfies the following two inequalities:
$$\exp \left\{c_{\alpha, d} m^2 N \log (\epsilon^{-1}nmN) -
{n \epsilon^{2-{\frac{q}{q+1}}} \over s\left(10 (c_0)^{\frac{1}{q+1}} +2 \right)}\right\} \leq \frac{\delta}{4}$$
and 
$$\exp \left\{c_{\alpha, d} m^2 N \log (\epsilon^{-1}n^\prime mN)-
{n' \epsilon^{2-{\frac{q}{q+1}}} \over (1-s) \left(10 (c_0)^{\frac{1}{q+1}} +2 \right)}\right\} \leq \frac{\delta}{4}.$$ 

Let $\hat \epsilon = \epsilon^{2-{\frac{q}{q+1}}}$ and let $c_q = 10 (c_0)^{\frac{1}{q+1}} +2$. We see that $c_q$ is a positive constant depending only on $q$ and $c_0$. We need to find $\hat \epsilon>0$ such that 
\begin{equation}\label{firstinq}
    \frac{c_{\alpha, d}}{2-{\frac{q}{q+1}}} m^2 N \log (\hat \epsilon^{-1}) - \frac{n\hat \epsilon }{s c_{q}}  \leq \log \left(\frac{\delta}{4}\right) -  c_{\alpha, d} m^2 N \log(nmN)    
\end{equation}
and 
\begin{equation}\label{secondinq}
    \frac{c_{\alpha, d}}{2-{\frac{q}{q+1}}} m^2 N \log (\hat \epsilon^{-1}) - \frac{n^\prime \hat \epsilon }{(1-s) c_{q}}  \leq \log \left(\frac{\delta}{4}\right) -  c_{\alpha, d} m^2 N \log(n^\prime mN).    
\end{equation}
We define a function $T:(0, \infty) \to \RR$ given by 
\begin{equation}\label{Tfunction}
     T(x) = \frac{c_{\alpha, d}}{2-{\frac{q}{q+1}}} m^2 N \log (x^{-1}) - \frac{nx }{sc_{q}}.
\end{equation}     
We notice that $T$ is a decreasing function. Then, we take
\begin{equation}\label{B}
    B:=c_{q} \left(\frac{2c_{\alpha, d}}{2-{\frac{q}{q+1}}} m^2 N + \log \left(\frac{4}{\delta}\right) + c_{\alpha, d} m^2 N\log(nmN)\right).  
\end{equation}
For $n \geq 3$ (i.e., $\log (n) >1$), there holds 
\begin{equation}\label{Bs}
    \frac{B s\log \left(\frac{n}{s}\right)}{n}
 \geq \frac{s}{n}.
\end{equation}
It follows that  
\begin{eqnarray*}
   && T\left( \frac{Bs\log \left(\frac{n}{s}\right)}{n}
\right)\\
   &\stackrel{\eqref{Tfunction}}{=}& \frac{c_{\alpha, d}}{2-{\frac{q}{q+1}}} m^2 N \log \left(\frac{n}{Bs\log (\frac{n}{s})}\right) - \frac{n}{sc_{q}}\left(\frac{Bs\log (\frac{n}{s})}{n}\right)\\
   &\stackrel{\eqref{B}, \eqref{Bs}}{\leq}& \frac{c_{\alpha, d}}{2-{\frac{q}{q+1}}} m^2 N \log \left(\frac{n}{s}\right)\\
   &&\qquad- \log \left(\frac{n}{s}\right)\left(\frac{c_{\alpha, d}}{2-{\frac{q}{q+1}}} m^2 N + \log \left(\frac{4}{\delta}\right) + c_{\alpha, d} m^2 N \log(nmN)\right)\\
   &=&  - \log \left(\frac{n}{s}\right) \left(\log \left(\frac{4}{\delta}\right) + c_{\alpha, d} m^2 N \log(nmN)\right)\\
&\stackrel{\because \log (\frac{n}{s}) > 1}{\leq}& \log \left(\frac{\delta}{4}\right) -  c_{\alpha, d} m^2  N\log(nmN). 
\end{eqnarray*}
Since $T$ is a decreasing function, we know that if
\begin{eqnarray*}
  \hat \epsilon \geq  \frac{Bs\log \left(\frac{n}{s}\right)}{n}
,
\end{eqnarray*}
then such a $\hat \epsilon$ satisfies the first inequality \eqref{firstinq}.
In the same way, we know that if
\begin{eqnarray*}
    \hat\epsilon \geq \frac{B^\prime(1-s)}{n^\prime}\log \left(\frac{n^\prime}{1-s}\right)
\end{eqnarray*}
with \begin{eqnarray*}
    B^\prime:=c_{q} \left(\frac{2c_{\alpha, d}}{2-{\frac{q}{q+1}}} m^2 N + \log \left(\frac{4}{\delta}\right) + c_{\alpha, d} m^2 N\log(n^\prime mN)\right),  
\end{eqnarray*}
then such a $\hat\epsilon$ satisfies the second inequality \eqref{secondinq}.

We choose $\hat \epsilon$ to be  
\begin{eqnarray*}
   \hat\epsilon =  \max \left\{  \frac{Bs\log \left(\frac{n}{s}\right)}{n}
, \frac{B^\prime(1-s)}{n^\prime}\log \left(\frac{n^\prime}{1-s}\right) \right\}.
\end{eqnarray*}
Thus, $\hat\epsilon $ satisfies both inequalities \eqref{firstinq} and \eqref{secondinq}. Then, we have  \begin{equation}\label{choice_of_epsilon}
    \epsilon =   \left(\max\left\{  \frac{Bs\log \left(\frac{n}{s}\right)}{n}
, \frac{B^\prime(1-s)}{n^\prime}\log \left(\frac{n^\prime}{1-s}\right) \right\}\right)^{\frac{q+1}{q+2} }
\end{equation}
because $\hat \epsilon = \epsilon^{2-{\frac{q}{q+1}}}$.

Now, we take $f = \Hat{f}_{T, T^\prime,\phi} \in \mathcal{H}_\tau$ and the above choice of $\epsilon$ \eqref{choice_of_epsilon}, we have,  with probability at least $1-\delta/2$, 
\begin{eqnarray*}
&&  \varepsilon (\Hat{f}_{T, T^\prime,\phi}) - \varepsilon (f_c) - \left(\varepsilon_{T, T'} (f) - \varepsilon_{T, T'} (f_c)\right)\\
  &\stackrel{\eqref{5epsilon}}{\leq}& 5 \epsilon^{1-{\frac{q}{2(q+1)}}}\left(\left(\varepsilon (\Hat{f}_{T, T^\prime,\phi}) - \varepsilon (f_c)\right)^{\frac{q}{q+1}} + \epsilon^{\frac{q}{q+1}}\right)^{1/2}\\
  &\stackrel{\eqref{choice_of_epsilon}}{=}& 5 \left(  \max\left\{ B \frac{s}{n}
\log \left(\frac{n}{s}\right),B^\prime \frac{(1-s)}{n^\prime}\log \left(\frac{n^\prime}{1-s}\right) \right\}\right)^{1/2}\\
&&\left(\left(\varepsilon (\Hat{f}_{T, T^\prime,\phi}) - \varepsilon (f_c)\right)^{\frac{q}{q+1}} + \left( \max\left\{ B  \frac{s}{n}
\log \left(\frac{n}{s}\right), B^\prime \frac{(1-s)}{n^\prime}\log \left(\frac{n^\prime}{1-s}\right) \right\}\right)^{\frac{q}{q+2} } \right)^{1/2} \\
&\leq& 5 \left(  \max\left\{ B \frac{s}{n}
\log \left(\frac{n}{s}\right), B^\prime \frac{(1-s)}{n^\prime}\log \left(\frac{n^\prime}{1-s}\right) \right\} \right)^{1/2} \left(\varepsilon (\Hat{f}_{T, T^\prime,\phi}) - \varepsilon (f_c)\right)^{\frac{q}{2(q+1)}} \\
&& + 5 \left(  \max\left\{ B \frac{s}{n}
\log \left(\frac{n}{s}\right), B^\prime\frac{(1-s)}{n^\prime}\log \left(\frac{n^\prime}{1-s}\right) \right\}\right)^{\frac{q+1}{q+2}}.
\end{eqnarray*}
Here, we have used $\sqrt{a+b} \leq \sqrt{a} + \sqrt{b}$ for all $a,b >0$ in the last inequality.  To further tighten this upper bound, we apply Young's Inequality \eqref{Young} with\\
$a=5 \left(  \max\left\{  B\frac{s}{n}
\log \left(\frac{n}{s}\right),  B^\prime \frac{(1-s)}{n^\prime}\log \left(\frac{n^\prime}{1-s}\right) \right\} \right)^{1/2}$,
$b= \left(\varepsilon (\Hat{f}_{T, T^\prime,\phi}) - \varepsilon (f_c)\right)^{\frac{q}{2(q+1)}}, p=\frac{2(q+1)}{q+2}$, $ p^*=\frac{2(q+1)}{q}$ and get
\begin{eqnarray}\label{apply_Young}
  &&  5 \left(  \max\left\{  B\frac{s}{n}
\log \left(\frac{n}{s}\right),  B^\prime \frac{(1-s)}{n^\prime}\log \left(\frac{n^\prime}{1-s}\right) \right\} \right)^{1/2} \left(\varepsilon (\Hat{f}_{T, T^\prime,\phi}) - \varepsilon (f_c)\right)^{\frac{q}{2(q+1)}} \nonumber\\
&\leq&   \frac{\left( 5\left(  \max\left\{ B \frac{s}{n}
\log \left(\frac{n}{s}\right),B^\prime \frac{(1-s)}{n^\prime}\log \left(\frac{n^\prime}{1-s}\right) \right\} \right)^{1/2}\right)^{\frac{2(q+1)}{q+2}}}{\frac{2(q+1)}{q+2}} + \frac{\varepsilon (\Hat{f}_{T, T^\prime,\phi}) - \varepsilon (f_c)}{\frac{2(q+1)}{q}} \nonumber\\
&\leq&5^{\frac{2(q+1)}{q+2}} \left(  \max\left\{  B\frac{s}{n}
\log \left(\frac{n}{s}\right), B^\prime \frac{(1-s)}{n^\prime}\log \left(\frac{n^\prime}{1-s}\right) \right\} \right)^{\frac{q+1}{q+2}} + \frac{\varepsilon (\Hat{f}_{T, T^\prime,\phi}) - \varepsilon (f_c)}{2}.
\end{eqnarray}
Furthermore, we have 
\begin{eqnarray}\label{Bbound}
    B&\stackrel{\eqref{B}}{=}&c_{q} \left(\frac{2c_{\alpha, d}}{2-{\frac{q}{q+1}}} m^2 N + \log \left(\frac{4}{\delta}\right) + c_{\alpha, d} m^2 N\log(nmN)\right) \nonumber\\
    &\leq& c_{q} \left(\frac{2c_{\alpha, d}}{2-{\frac{q}{q+1}}} + c_{\alpha, d}\right)\left(m^2 N + \log \left(\frac{4}{\delta}\right) + m^2 N\log(nmN)\right)\nonumber\\
     &\leq&  c_{q,\alpha,d}\left(\log \left(\frac{4}{\delta}\right) + m^2 N (1+\log(nmN))\right),
\end{eqnarray}
where $c_{q,\alpha, d} = c_{q} \left(\frac{2c_{\alpha, d}}{2-{\frac{q}{q+1}}} + c_{\alpha, d}\right) \geq 1$ is a positive constant depending only on $q, c_0 , \alpha$, and $d$. 
Similarly, we know 
\begin{equation}\label{Bprimebound}
    B^\prime \leq  c_{q,\alpha,d}\left(\log \left(\frac{4}{\delta}\right) + m^2 N (1+\log(n^\prime mN))\right). 
\end{equation}
 
Finally, we get with probability at least $1-\delta/2$,
\begin{eqnarray*}
    &&  \varepsilon (\Hat{f}_{T, T^\prime,\phi}) - \varepsilon (f_c) - \left(\varepsilon_{T, T'} (f) - \varepsilon_{T, T'} (f_c)\right)\\ 
    &\stackrel{\eqref{apply_Young}}{\leq}&  5^{\frac{2(q+1)}{q+2}}  \left(  \max\left\{ B \frac{s}{n}
    \log \left(\frac{n}{s}\right), B^\prime \frac{(1-s)}{n^\prime}\log \left(\frac{n^\prime}{1-s}\right) \right\} \right)^{\frac{q+1}{q+2}} + \frac{\varepsilon (\Hat{f}_{T, T^\prime,\phi}) - \varepsilon (f_c)}{2} \\
 &\leq&  5^{\frac{2(q+1)}{q+2}}  \left(\max\{B, B^\prime\}\right)^{\frac{q+1}{q+2}}\left(  \max\left\{  \frac{s}{n}
\log \left(\frac{n}{s}\right), \frac{(1-s)}{n^\prime}\log \left(\frac{n^\prime}{1-s}\right) \right\} \right)^{\frac{q+1}{q+2}}\\
&&\qquad\qquad+ \frac{\varepsilon (\Hat{f}_{T, T^\prime,\phi}) - \varepsilon (f_c)}{2} \\
&\stackrel{\eqref{Bbound}, \eqref{Bprimebound}}{\leq}& C_{q,\alpha, d} \left(\log \left(\frac{4}{\delta}\right) + m^2 N (\log(nmN)+\log(n^\prime mN))\right)^{\frac{q+1}{q+2}} \\
&& \cdot \left(  \max\left\{  \frac{s}{n}
\log \left(\frac{n}{s}\right), \frac{(1-s)}{n^\prime}\log \left(\frac{n^\prime}{1-s}\right) \right\} \right)^{\frac{q+1}{q+2}} 
 + \frac{\varepsilon (\Hat{f}_{T, T^\prime,\phi}) - \varepsilon (f_c)}{2}, 
\end{eqnarray*}
where $C_{q,\alpha, d}$ is a positive constant depending only on $q,c_0, \alpha$, and $d$. 

The proof is complete.
\end{proof}

\subsection{Combining error bounds together}\label{subsec:combine}
 Now that we have derived the upper bounds of the estimation errors and the approximation error, we can combine them together to prove Theorem 4.1. 

 \begin{proof}[Proof of Theorem 4.1]
 We first plug in the upper bounds of the estimation error terms given in Lemma \ref{lemma: estimationI} and Lemma \ref{lemma:estimationII}, respectively. 
Recall $n^* = \min\{n,n^\prime\} \geq 3$. Since the function $x \mapsto x \log(1/x)$ is increasing on $(0, 1/3]$, we find
\begin{eqnarray*}
    \max\left\{\frac{s}{n} \log\left(\frac{n}{s}\right), \frac{1-s}{n^\prime} \log\left(\frac{n^\prime}{1-s}\right)\right\} \leq \frac{1}{n^*} \log(n^*).
\end{eqnarray*}
Then, when $N \geq \max \left\{(\alpha +1)^d, (r+1)e^d\right\}$, 
  with probability at least $1-\delta$, we have 
  \begin{eqnarray*}
  &&\varepsilon(\Hat{f}_{T, T^\prime,\phi}) - \varepsilon(f_c)\\
       &\stackrel{\eqref{decomposition}}{\leq}&   \{\varepsilon(\Hat{f}_{T, T^\prime,\phi})- \varepsilon_{T,T^\prime}(\Hat{f}_{T, T^\prime,\phi})\}  +  \{\varepsilon_{T,T^\prime}(f_\mathcal{H}) - \varepsilon(f_\mathcal{H})\} + \{\varepsilon(f_\mathcal{H}) - \varepsilon(f_c)\}\\
       &\leq & C_{q,\alpha, d} \left(\log \left(\frac{4}{\delta}\right) + 2m^2 N \log(nmN)\right)^{\frac{q+1}{q+2}} \left(   \frac{s}{n}
\log \left(\frac{n}{s}\right)\right)^{\frac{q+1}{q+2}} 
+ \frac{\varepsilon (\Hat{f}_{T, T^\prime,\phi}) - \varepsilon (f_c)}{2}\\
&& +  C_{q} \left(\log\left(\frac{4}{\delta}\right)\right)\left(s^{\frac{1}{q+2}}\left(\frac{1}{n}\right)^{\frac{q+1}{q+2}}+(1-s)^{\frac{1}{q+2}}\left(\frac{1}{n^\prime}\right)^{\frac{q+1}{q+2}}\right) \\
&\leq& C_{q,\alpha, d} \left(\log \left(\frac{4}{\delta}\right) + 2m^2 N \log(\max\{n, n^\prime\}mN)\right)^{\frac{q+1}{q+2}} \left(   \frac{\log(n^*)}{n^*}
\right)^{\frac{q+1}{q+2}} 
+ \frac{\varepsilon (\Hat{f}_{T, T^\prime,\phi}) - \varepsilon (f_c)}{2}\\
&& +  C_{q} \left(\log\left(\frac{4}{\delta}\right)\right)\left(s^{\frac{1}{q+2}}\left(\frac{1}{n}\right)^{\frac{q+1}{q+2}}+(1-s)^{\frac{1}{q+2}}\left(\frac{1}{n^\prime}\right)^{\frac{q+1}{q+2}}\right) \\
&&+ \frac{(\varepsilon(f_\mathcal{H}) - \varepsilon(f_c))}{2} + (\varepsilon(f_\mathcal{H}) - \varepsilon(f_c)),
\end{eqnarray*}
which implies
\begin{eqnarray}\label{final_decom}
&&  \varepsilon(\Hat{f}_{T, T^\prime,\phi}) - \varepsilon(f_c)\nonumber\\
&\leq& 2C_{q,\alpha, d} \left(\log \left(\frac{4}{\delta}\right) + 2m^2 N \log(\max\{n, n^\prime\}mN)\right)^{\frac{q+1}{q+2}} \left(   \frac{\log(n^*)}{n^*}\right)^{\frac{q+1}{q+2}} \nonumber\\
&&+ 4C_{q} \left(\log\left(\frac{4}{\delta}\right)\right)\left(\frac{1}{n^*}\right)^{\frac{q+1}{q+2}} + 3(\varepsilon(f_\mathcal{H}) - \varepsilon(f_c)). 
  \end{eqnarray}
  
  For the last term $\varepsilon(f_\mathcal{H}) - \varepsilon(f_c)$ (i.e., the approximation error term), we use the upper bound derived in Lemma \ref{lemma:approx} in \eqref{approx_eqn}, which is given by 
  \begin{equation*}
  \varepsilon(f_\mathcal{H}) - \varepsilon(f_c) \leq 8c_0\left(\tau +\|\widehat{\eta}-\eta\|_{L^\infty[0,1]^d}\right)^{q+1}.
  \end{equation*}
  We can take $\widehat{\eta} \in \mathcal{H}_\tau$ such that
  $$\|\widehat{\eta}-\eta\|_{L^\infty[0,1]^d}\leq (2r+1)(1+d^2 + \alpha^2)6^d N 2^{-m} + r3^\alpha N^{-\frac{\alpha}{d}}$$
by \eqref{schmidt_error} in Lemma \ref{schmidt}. 

We choose the smallest $m\in \NN$ such that 
$N 2^{-m} \leq N^{-\frac{\alpha}{d}}$, that is \begin{equation}\label{choice_of_m}
    m =\left \lceil \left(1+ \frac{\alpha}{d}\right) \frac{\log N}{\log 2}\right \rceil.
\end{equation} 
Then 
\begin{eqnarray}\label{large_N}
   N 2^{-m}  +  N^{-\frac{\alpha}{d}} \leq 2 N^{-\frac{\alpha}{d}}. 
\end{eqnarray}

Take $\tau = N^{-\frac{\alpha}{d}}$.  By choosing $c_{r,\alpha,d} = 1+ (2r+1)(1+d^2 + \alpha^2)(6^d+3^\alpha)$, we see that 
    \begin{eqnarray}\label{approx_update}
  \varepsilon(f_\mathcal{H}) - \varepsilon(f_c)&\leq&  
  8c_0\left(\tau +\|\widehat{\eta}-\eta\|_{L^\infty[0,1]^d}\right)^{q+1} \nonumber\\
 &\leq& 8c_0 \left(c_{r,\alpha,d}\right)^{q+1} N^{-\frac{\alpha(q+1)}{d}} .
  \end{eqnarray}
  Next, we plug in the above upper bound of $\varepsilon(f_\mathcal{H}) - \varepsilon(f_c)$ in \eqref{approx_update} to \eqref{final_decom} and get
  \begin{eqnarray*}
      \varepsilon(\Hat{f}_{T, T^\prime,\phi}) - \varepsilon(f_c)
      &\stackrel{\eqref{approx_update}}{\leq}& 2C_{q,\alpha, d} \left(\log \left(\frac{4}{\delta}\right) + 2m^2 N \log(\max\{n, n^\prime\}mN)\right)^{\frac{q+1}{q+2}} \left(\frac{\log(n^*)}{n^*}\right)^{\frac{q+1}{q+2}} \\
&&+ 4C_{q} \left(\log\left(\frac{4}{\delta}\right)\right)\left(\frac{1}{n^*}\right)^{\frac{q+1}{q+2}} + 24 c_0 \left(
 c_{r,\alpha,d}\right)^{q+1}  N^{-\frac{\alpha(q+1)}{d}}. 
  \end{eqnarray*}

  Our choice of $N$ is the smallest integer such that 
  \begin{eqnarray}\label{choice_of_N_II}
   && \left(N\frac{(\log n^*)^3 \log(\max\{n, n^\prime\})}{n^*}\right)^{1/(q+2)} \geq N^{-\frac{\alpha}{d}}.  
  \end{eqnarray}
  That is, $$N =\left\lceil\left(\frac{n^*}{(\log n^*)^3\log (\max\{n,n^\prime\}) }\right)^{\frac{d}{d+ \alpha (q+2)}}\right\rceil.$$
  With our choices of $m$ and $N$, we have 
  \begin{eqnarray*}
      mN \leq 2\left(1+ \frac{\alpha}{d}\right)N^2 \leq 2\left(1+ \frac{\alpha}{d}\right) (2n^*)^2,
  \end{eqnarray*}
  and thereby
  \begin{eqnarray*}
      m^2 N \log(\max\{n, n^\prime\}mN)\leq 4\left(1+ \frac{\alpha}{d}\right)^2 N (\log(n^*))^2 \left(1+\log\left(8\left(1+ \frac{\alpha}{d}\right)\right)\right)\log(\max\{n, n^\prime\}).
  \end{eqnarray*}
  Hence, 
  \begin{eqnarray*}
     && \left(\log \left(\frac{4}{\delta}\right) + 2m^2 N \log(\max\{n, n^\prime\}mN)\right) \left(\frac{\log(n^*)}{n^*}\right)\\
     &\leq& \left(\log \left(\frac{4}{\delta}\right) + 8\left(1+ \frac{\alpha}{d}\right)^2 N (\log(n^*))^2 \left(1+\log\left(8\left(1+ \frac{\alpha}{d}\right)\right)\right)\log(\max\{n, n^\prime\})\right) \left(\frac{\log(n^*)}{n^*}\right)\\
     &\leq&  \left(\log \left(\frac{4}{\delta}\right) + 8\left(1+ \frac{\alpha}{d}\right)^2  \left(1+\log\left(8\left(1+ \frac{\alpha}{d}\right)\right)\right)\right) \left(\frac{(\log(n^*))^3\log(\max\{n, n^\prime\}) }{n^*}\right)N\\
     &\leq&  \left(\log \left(\frac{4}{\delta}\right) + 8\left(1+ \frac{\alpha}{d}\right)^2  \left(1+\log\left(8\left(1+ \frac{\alpha}{d}\right)\right)\right)\right) N^{-\frac{\alpha(q+2)}{d}}.
  \end{eqnarray*}

  Finally, we have, with probability at least $1-\delta$,
  \begin{eqnarray*}
       \varepsilon(\Hat{f}_{T, T^\prime,\phi}) - \varepsilon(f_c) \stackrel{\eqref{choice_of_N_II}}{\leq}
      C_{q, r,\alpha,d}\log \left(\frac{4}{\delta}\right) \left(\frac{n^*}{(\log n^*)^3 \log(\max\{n, n^\prime\})}\right)^{-\frac{\alpha (q+1)}{d+ \alpha (q+2)}},
  \end{eqnarray*}
  where $C_{q, r,\alpha,d}$ is a positive constant depending only on $q,c_0, r,\alpha$, and $d$. 
  
  The proof is complete. 

 \end{proof}


\end{document}